\documentclass{article} 

\usepackage{times}

\usepackage{booktabs}
\usepackage{graphicx,caption}
\usepackage{authblk}

\usepackage{amsmath,amsthm,amsfonts,amssymb,amscd,pifont}
\usepackage{verbatim}
\usepackage{url}
\usepackage{natbib}

\usepackage{makecell}

\usepackage{mathrsfs}
\usepackage[ansinew]{inputenc}
\usepackage{graphicx}
\usepackage{a4wide}
\usepackage{subfigure}
\usepackage{graphics}

\usepackage{tabularx}
\usepackage{booktabs}

\usepackage{graphicx}
\usepackage{hyperref}
\usepackage{xfrac}
\usepackage{array}
\usepackage[all]{xy}
\usepackage{epic}
\usepackage{latexsym}
\usepackage{enumerate}
\usepackage{appendix}

\newtheorem*{Theorem*}{Theorem}

\newtheorem{maintheorem}{Theorem}

\newcommand{\cmt}{\begin{maintheorem}}
\newcommand{\fmt}{\end{maintheorem}}

\newtheorem{maincorollary}[maintheorem]{Corollary}

\newcommand{\cmc}{\begin{maincorollary}}
\newcommand{\fmc}{\end{maincorollary}}

\newtheorem{T}{Theorem}[section]
\newcommand{\ct}{\begin{T}}
\newcommand{\ft}{\end{T}}
\newtheorem{D}{Definition}[section]
\newcommand{\cd}{\begin{D}}
\newcommand{\fd}{\end{D}}

\newtheorem{Corollary}[T]{Corollary}
\newcommand{\cco}{\begin{Corollary}}
\newcommand{\fco}{\end{Corollary}}

\newtheorem{Proposition}[T]{Proposition}
\newcommand{\cpr}{\begin{Proposition}}
\newcommand{\fpr}{\end{Proposition}}

\newtheorem{Lemma}[T]{Lemma}
\newcommand{\cle}{\begin{Lemma}}
\newcommand{\fle}{\end{Lemma}}

\newtheorem{Sublemma}[T]{Sublemma}
\newcommand{\csle}{\begin{Sublemma}}
\newcommand{\fsle}{\end{Sublemma}}

\usepackage{graphicx}
\usepackage{xcolor}
\usepackage{mathtools}
\usepackage{mathrsfs}

\newcommand{\be}{\begin{eqnarray}}
\newcommand{\ee}{\end{eqnarray}}

\usepackage{caption}  
\usepackage{textcomp} 
\usepackage{multirow}  
\usepackage{amsmath}

\usepackage{tikz}
\usetikzlibrary{positioning, shapes.geometric}

\usepackage{pgfplots}
\usepackage{pgfplotstable}
\usepackage{tikz}
\usetikzlibrary{shapes.geometric, arrows.meta, positioning}

\usepackage{pgfplots}
\usepgfplotslibrary{polar}
\pgfplotsset{compat=1.18}

\usepackage{hyperref}
\usepackage{url}

\title{ Model Comparisons: XNet Outperforms KAN}

\author[1,2]{Xin Li\thanks{Email: \texttt{xinli2023@u.northwestern.edu}}}
\author[3,4]{Zhihong Xia\thanks{Corresponding author: \texttt{xia@math.northwestern.edu}}}
\author[5]{Xiaotao Zheng\thanks{Email: \texttt{20234013002@stu.suda.edu.cn}}}

\affil[1]{Department of Computer Science, Northwestern University, Evanston, IL, USA}

\affil[2]{Mathematical Modelling and Data Analytics Center, Oxford Suzhou Centre for Advanced Research,Suzhou, China}

\affil[3]{School of Natural Science, Great Bay University, Guangdong, China}
\affil[4]{Department of Mathematics, Northwestern University, Evanston, IL, USA}
\affil[5]{Department of Mathematics,Soochow University,Suzhou, China}

\begin{document}

\maketitle

\begin{abstract}
In the fields of computational mathematics and artificial
  intelligence, the need for precise data modeling is crucial,
  especially for predictive machine learning tasks. This paper
  explores further XNet, a novel algorithm that employs the complex-valued
  Cauchy integral formula, offering a superior network architecture
  that surpasses traditional Multi-Layer Perceptrons (MLPs) and
  Kolmogorov-Arnold Networks (KANs). XNet significant
  improves speed and accuracy across various tasks in both low
  and high-dimensional spaces, redefining the scope of data-driven
  model development and providing substantial improvements over
  established time series models like LSTMs.

\end{abstract}

\section{Introduction}

We initially proposed a novel method for constructing real networks
from the complex domain using the Cauchy integral formula in
\cite{LXZ24,ZLX24}, utilizing Cauchy kernels as basis functions. This
work comprehensively compares these networks with KANs, which use
B-spline as basis functions in \cite{liu2024kan}, and MLPs to
highlight our significant improvements.

Multi-layer perceptrons (MLPs) (\cite{Haykin94, Cybenko89, Hornik89}),
recognized as fundamental building blocks in deep learning, have their
limitations despite their wide use, particularly in its accuracy, and
large number of parameters needed in structures such as in transformers
(\cite{Vaswani17}), and lack interpretability without post-analysis
tools (\cite{Cunningham23}). The Kolmogorov-Arnold Networks (KANs) were
introduced as a potential alternative, drawing on the Kolmogorov-Arnold
representation theorem (\cite{Kolmogorov56, Braun09}), and demonstrate
their efficiency and accuracy in computational tasks, especially in
solving PDEs and function approximation (\cite{Sprecher02,Koppen02,Lin93,Lai21,Leni13,Fakhoury22}).

In the swiftly advancing domain of deep learning, the continuous
search for novel neural network designs that deliver superior accuracy
and efficiency is pivotal. While traditional activation functions such
as the Rectified Linear Unit (ReLU) (\cite{Hinton2010}) have been widely
adopted due to their straightforwardness and efficacy in diverse
applications, their shortcomings become evident as the complexity of
challenges escalates. This is particularly true in areas that demand
meticulous data fitting and the solutions of intricate partial
differential equations (PDEs). These limitations have paved the way
for architectures that merge neural network techniques with PDEs,
significantly enhancing function approximation capabilities in
high-dimensional settings (\cite{Sirignano2018, Raissi2019, Jin2021,
  Transolver2024, PINNsFormer2023}).
  
Time series forecasting is critical in various sectors including
finance, healthcare, and environmental science. While LSTM models are
well-regarded for their ability to capture temporal dependencies
(\cite{Yu2019,Zhao2017}), KAN models have also shown promise in managing
time series predictions (\cite{Hochreiter1997LongSM,
  staudemeyer2019understanding, xu2024kolmogorov}). Our study compares
these models, providing insights into their applications and
theoretical foundations. We also examine the performance of
transformers and our novel XNet model in time series forecasting in
the appendix, highlighting their capabilities in managing sequential
data (\cite{Vaswani17, wen2023transformers}).

Inspired by the mathematical precision of the Cauchy integral theorem,
\cite{LXZ24} introduced the XNet architecture, a novel neural network
model that incorporates a uniquely designed Cauchy activation
function. This function is mathematically expressed as:

$$\phi_a(x) = \frac{\lambda_1 * x}{x^2+d^2}+
\frac{\lambda_2}{x^2+d^2},$$ where $\lambda_1$, $\lambda_2$, and $d$
are parameters optimized during training. This design is not only a
theoretical advancement but also empirical advantageous, offering a
promising alternative to traditional models for many applications. By
integrating Cauchy activation functions, XNet demonstrates superior
performance in function approximation tasks and in solving
low-dimensional PDEs compared to its contemporaries, namely Multilayer
Perceptrons (MLPs) and Kolmogorov-Arnold Networks (KANs). This paper
will systematically compare these architectures, highlighting XNet's
advantages in terms of accuracy, convergence speed, and computational
demands.

Furthermore, empirical evaluations reveal that the Cauchy activation
function possesses a localized response with decay at both ends,
significantly benefiting the approximation of localized data
segments. This capability allows XNet to fine-tune responses to
specific data characteristics, a critical advantage over the globally
responding functions like ReLU.

The implications of this research are significant. It has been
demonstrated that the XNet can serve as an effective foundation for
general AI applications, our findings in this paper indicate that it
can even outperform meticulously designed networks tailored for
specific purposes.

Principal Contributions

Our study elucidates several critical advancements in the domain of
neural network architectures and their applications:

\begin{enumerate}[(i)]
\item Enhanced Function Approximation Capabilities: We conduct a
comparative analysis between XNet and KAN within the context of
function approximation, demonstratting the superior performance of XNet,
particularly in handling the Heaviside step function and complex
high-dimensional scenarios. Detailed examinations are presented in
Sections \ref{sec:1d} through \ref{sec:nd}, showcasing empirical
validations that underscore XNet's robust adaptability across varying
dimensions.

\item Superiority in Physics-Informed Neural Networks: Utilizing the
Poisson equation as a benchmark, we demonstrate XNet's enhanced
efficacy within the Physics-Informed Neural Network (PINN)
framework. Our results indicate that XNet significantly outstrips the
performance metrics of both Multi-Layer Perceptron (MLP) and KAN, as
detailed in Section \ref{sec:LSTM}. This investigation not only highlights
XNet's prowess but also sets a new benchmark for subsequent
applications in the field.

\item Innovation in Time Series Forecasting--By innovatively
substituting the conventional feedforward neural network (FNN) with
XNet in the LSTM architecture, we introduce the XLSTM model. In a
series of time series forecasting experiments, XLSTM consistently
surpasses traditional LSTM models in accuracy and reliability,
establishing a new frontier in predictive analytics. 
\end{enumerate}

\begin{figure}[h!]
    \centering
    \includegraphics[width=0.5\textwidth]{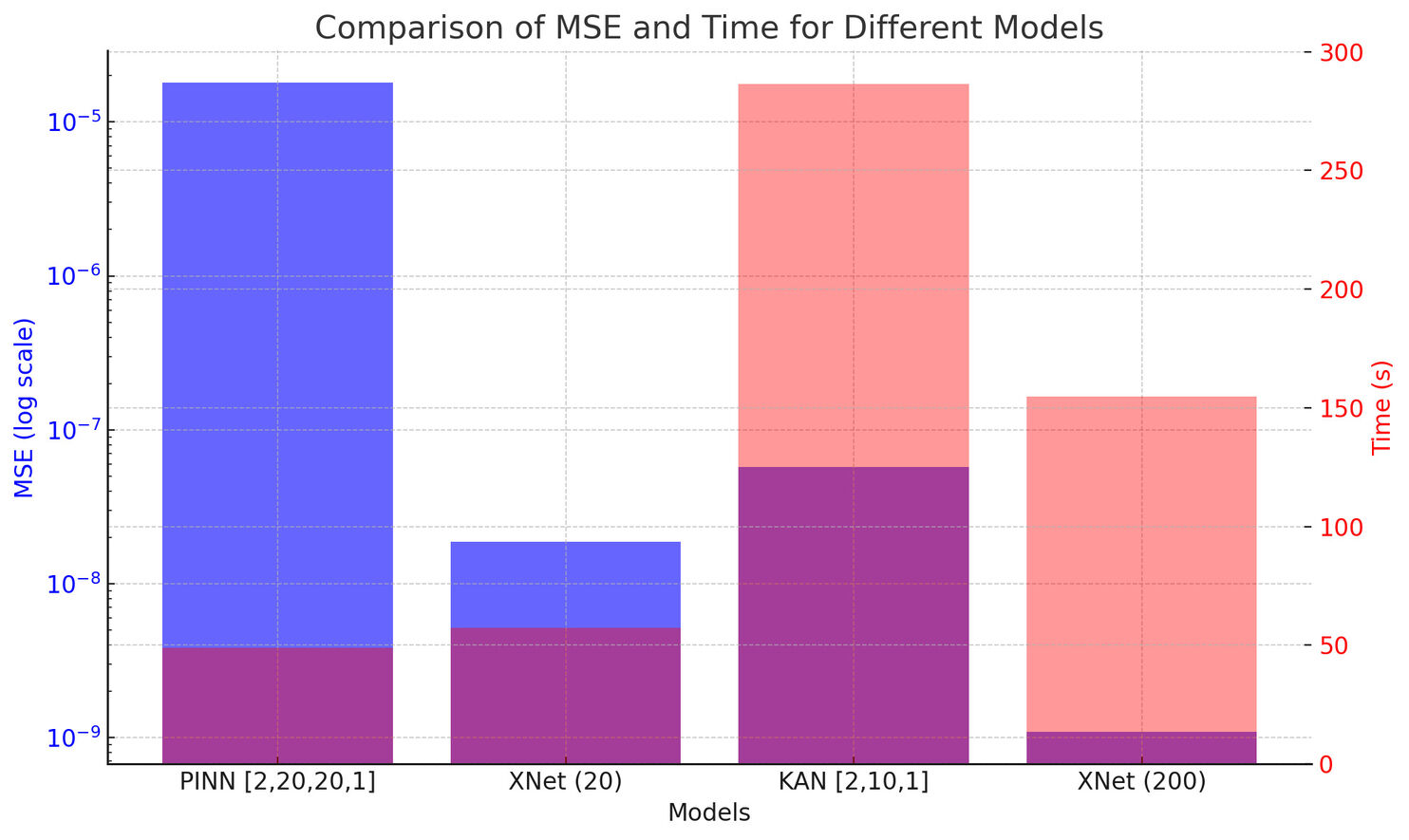}
    \caption{Comparing the MSE and training time for: PINN, XNet(20), KAN, and XNet(200). The MSE values are displayed on a logarithmic scale to better visualize the differences among the models.}
    \label{fig:compare4}
\end{figure}

We summarize our results with a representative graph (fig
\ref{fig:compare4}), which compares the performance of various models
in solving partial differential equations (PDEs). The parameterization
of Kolmogorov-Arnold Networks (KANs) is fundamentally different from
that of Multi-layer Perceptrons (MLPs); thus, even though KANs
sometimes require fewer parameters and fewer training iterations, the
training time can be substantially longer. In the context of solving
PDEs, XNets with 200 basis functions typically operate at a pace that
is 3-4 times slower than Physics-Informed Neural Networks (PINNs), 2
times faster than KANs, yet they achieve significantly higher
precision-10000 times more precise than PINNs, to be exact.

\section{Experimental Setup}

Our research is designed to rigorously evaluate the capabilities of
KAN and XNet across three fundamental domains: function approximation,
solving partial differential equations (PDEs), and time series
prediction. This structured evaluation allows us to systematically
assess the performance and applicability of each model in varied
computational tasks.

\textbf{Function Approximation:} We divide the function approximation
experiments based on the dimensionality and complexity of the
functions:
\begin{itemize}
\item \textbf{Low-Dimensional Functions:} Both irregular and regular
  functions are tested to evaluate the models' ability to handle
  variations in functional behavior and data distribution
  irregularities.
\item \textbf{High-Dimensional Functions:} Smooth functions that
  simulate complex real-world phenomena are used to examine the
  models' generalization in higher-dimensional spaces.
\end{itemize}
Evaluation metrics for accuracy, computational efficiency, and convergence are applied to each functional type.

\begin{table}[htbp]
\caption{Low-dimensional and High-dimensional Functions Examples}
\centering
\begin{tabular}{ccc}  
\toprule
\multicolumn{3}{c}{\textbf{Several Types of Functions and Their Examples}} \\
\midrule
\begin{minipage}[b]{0.3\textwidth}  
\centering
$f(x) = \begin{cases} 
1, & x > 0 \\
0, & \text{otherwise}
\end{cases}$
\end{minipage} & 
\begin{minipage}[b]{0.3\textwidth} 
\centering
$f(x,y) = \exp(\sin(\pi x) + y^2)$
\end{minipage} &jpg
\begin{minipage}[b]{0.3\textwidth} 
\centering
$f(x,y) = xy$
\end{minipage} \\
\begin{minipage}[b]{0.3\textwidth}
\centering
\includegraphics[width=0.7\textwidth]{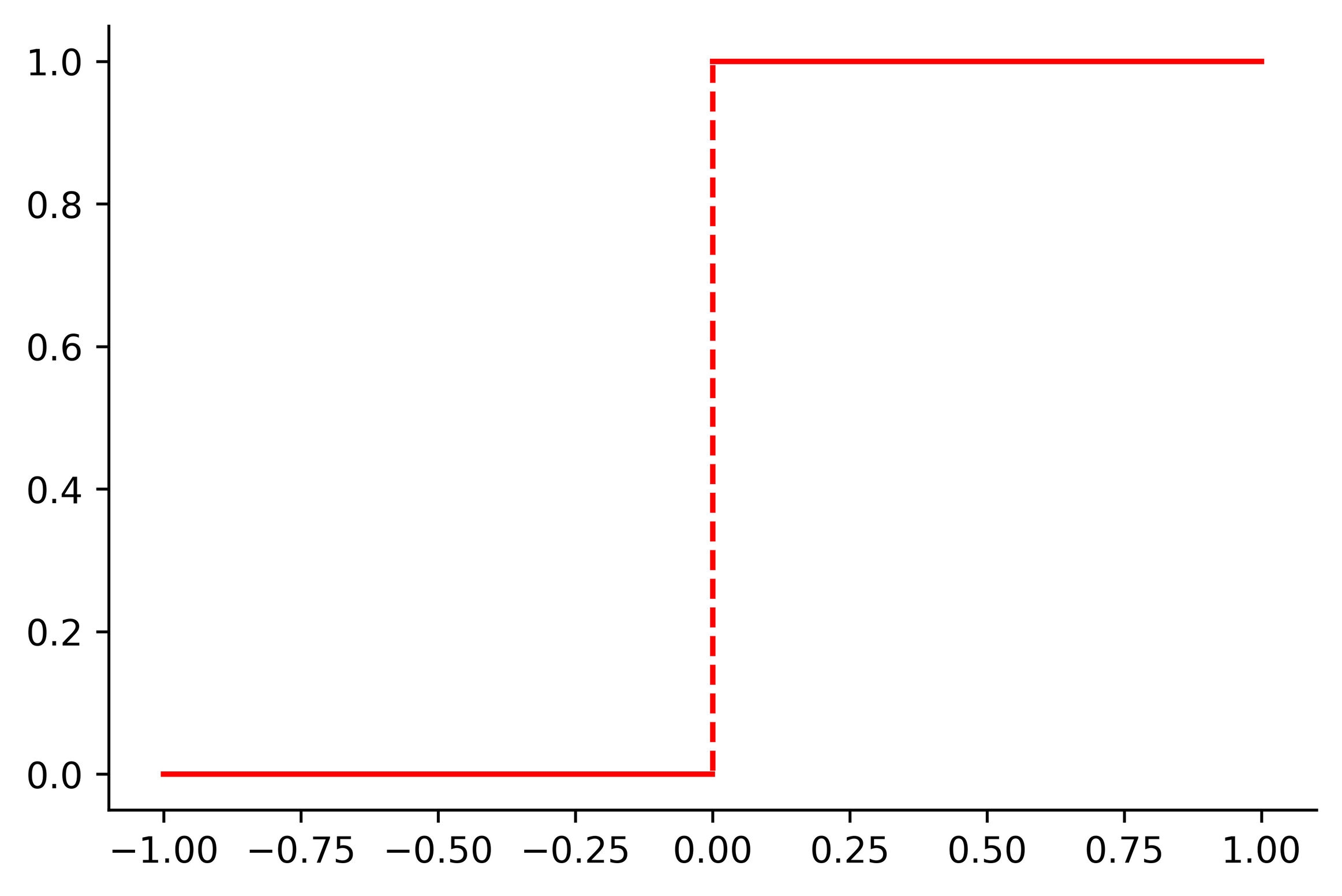}  
\end{minipage} & 
\begin{minipage}[b]{0.3\textwidth}
\centering
\includegraphics[width=0.7\textwidth]{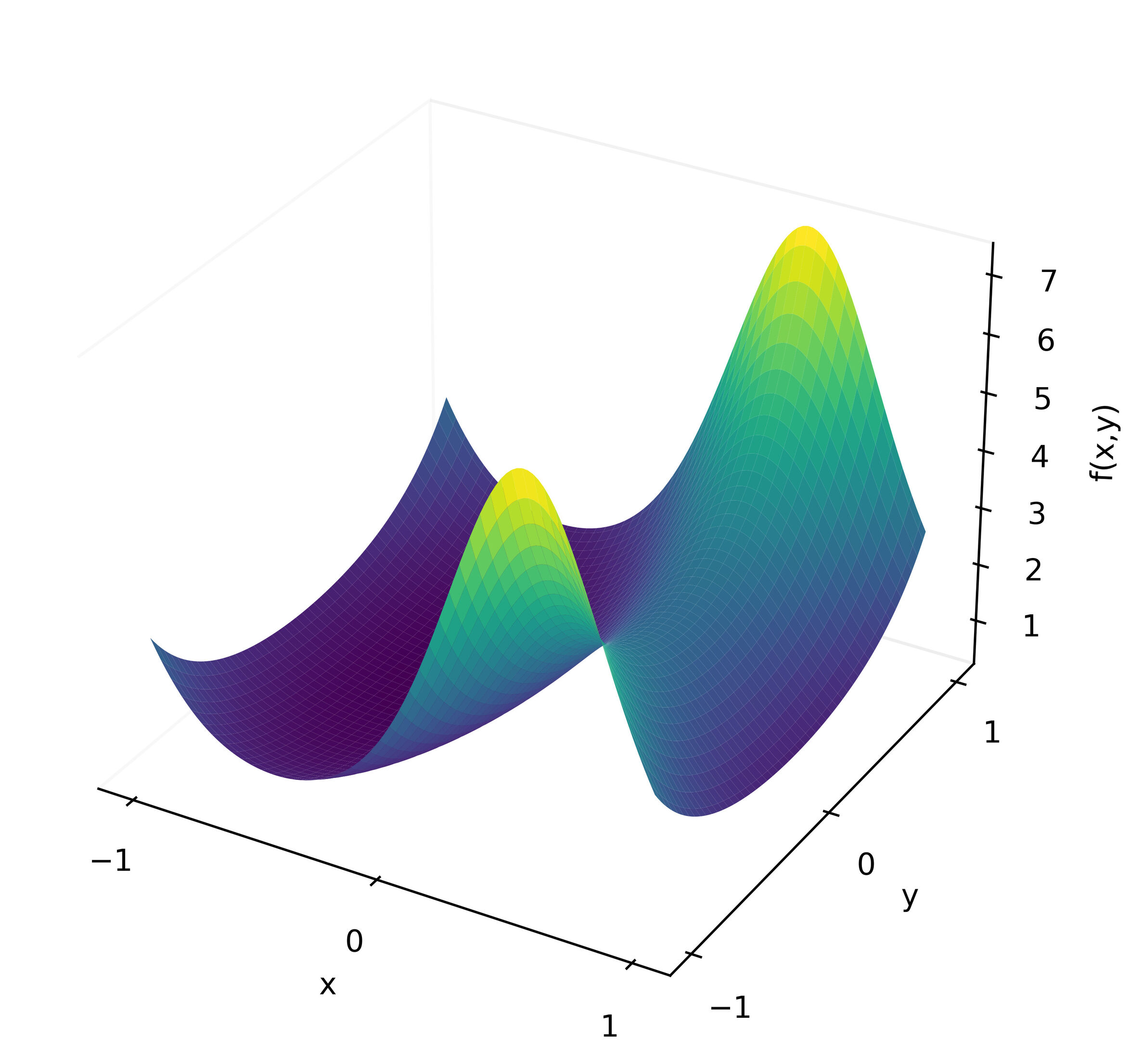}  
\end{minipage} &
\begin{minipage}[b]{0.3\textwidth}
\centering
\includegraphics[width=0.7\textwidth]{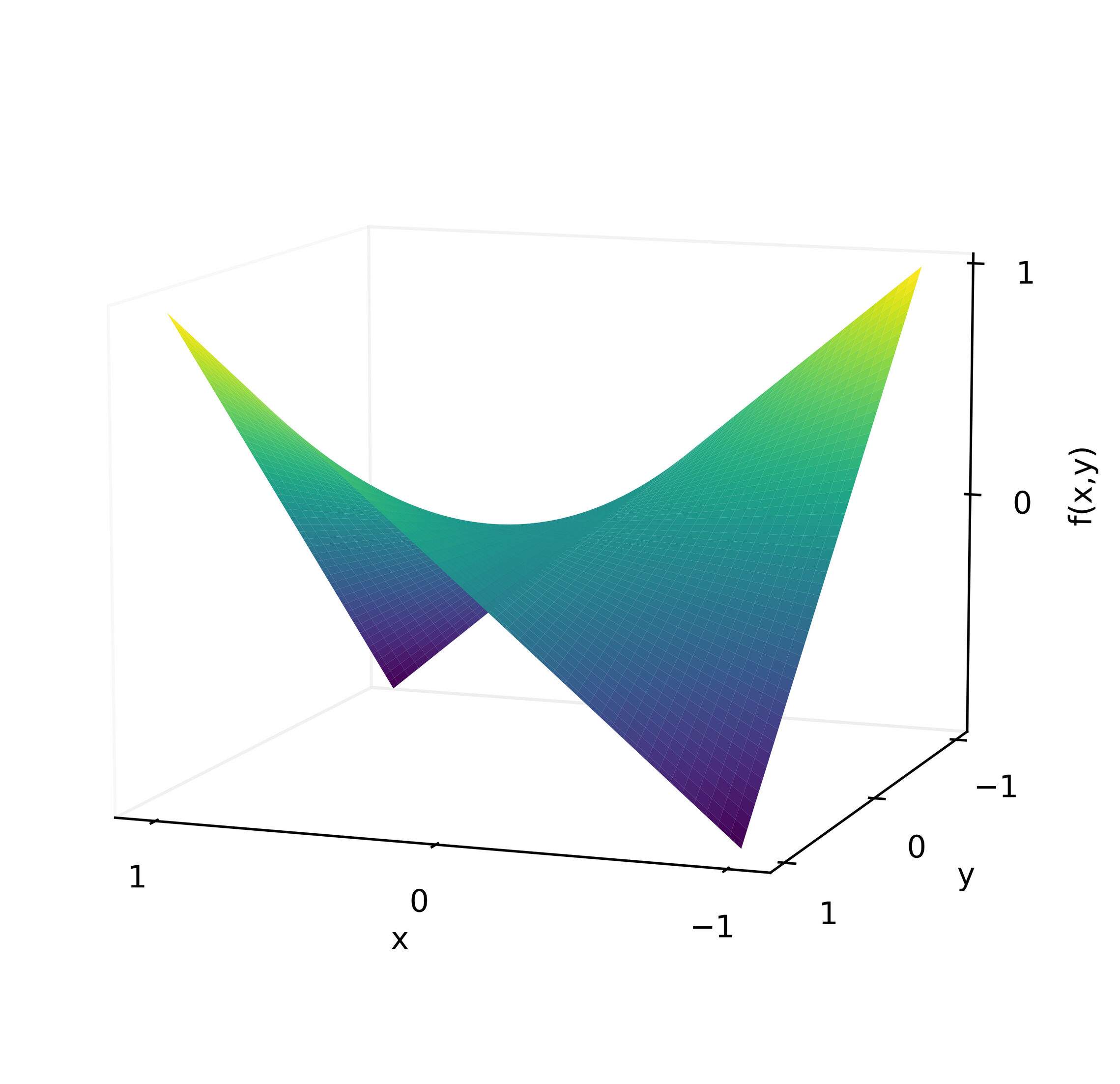}  
\end{minipage} \\
\midrule
\multicolumn{3}{c}{\textbf{High-dimensional Functions}} \\
\midrule
\multicolumn{3}{c}{
\begin{minipage}[c]{0.8\textwidth}
\centering
\begin{equation*}
f(x_1,x_2,x_3,x_4) = \exp\left(\frac{1}{2}\left(\sin\left(\pi(x_{1}^{2}+x_{2}^{2})\right) + x_{3}x_{4}\right)\right)
\end{equation*}
\begin{equation*}
f(x_1, \dots, x_{100}) = \exp\left(\frac{1}{100}\sum_{i=1}^{100}\sin^2\left(\frac{\pi x_i}{2}\right)\right)
\end{equation*}
\end{minipage}
} \\
\bottomrule
\end{tabular}
\end{table}

\textbf{Solving Partial Differential Equations:} We utilize a series of well-known differential equations from physics and engineering to test the efficacy of KAN and XNet. These include:
\begin{itemize}
    \item Both linear and non-linear systems to provide a comprehensive assessment reflective of common scientific computing scenarios.
\end{itemize}

\begin{figure}[h!]
    \centering
    \begin{minipage}[t]{0.2\textwidth}
        \centering
        \includegraphics[width=\textwidth]{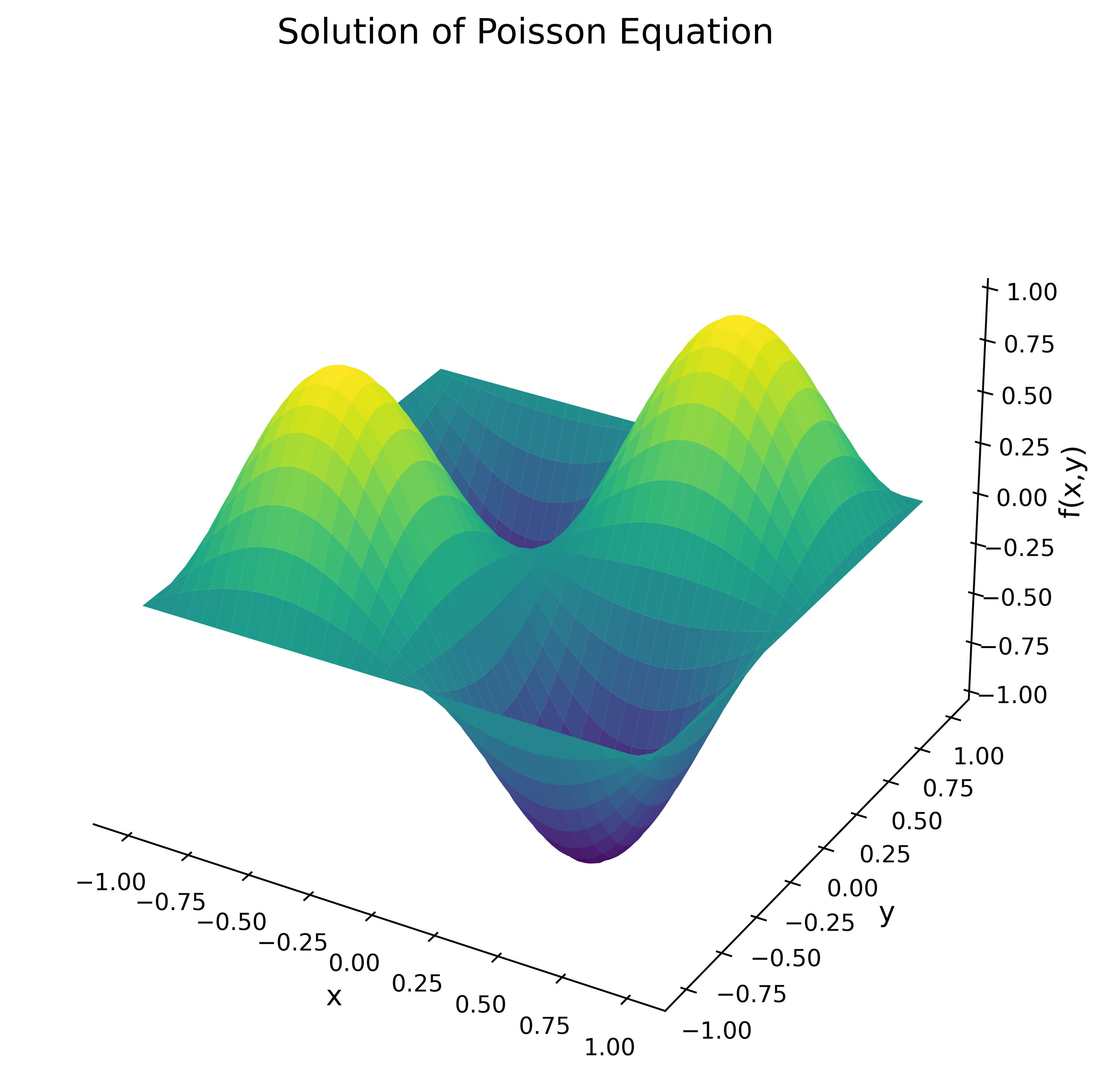}
        \caption{Solution of the Poisson equation}
        \label{fig:poisson}
    \end{minipage}
    \hspace{15pt}
    \begin{minipage}[t]{0.3\textwidth}
        \centering
        \includegraphics[width=\textwidth]{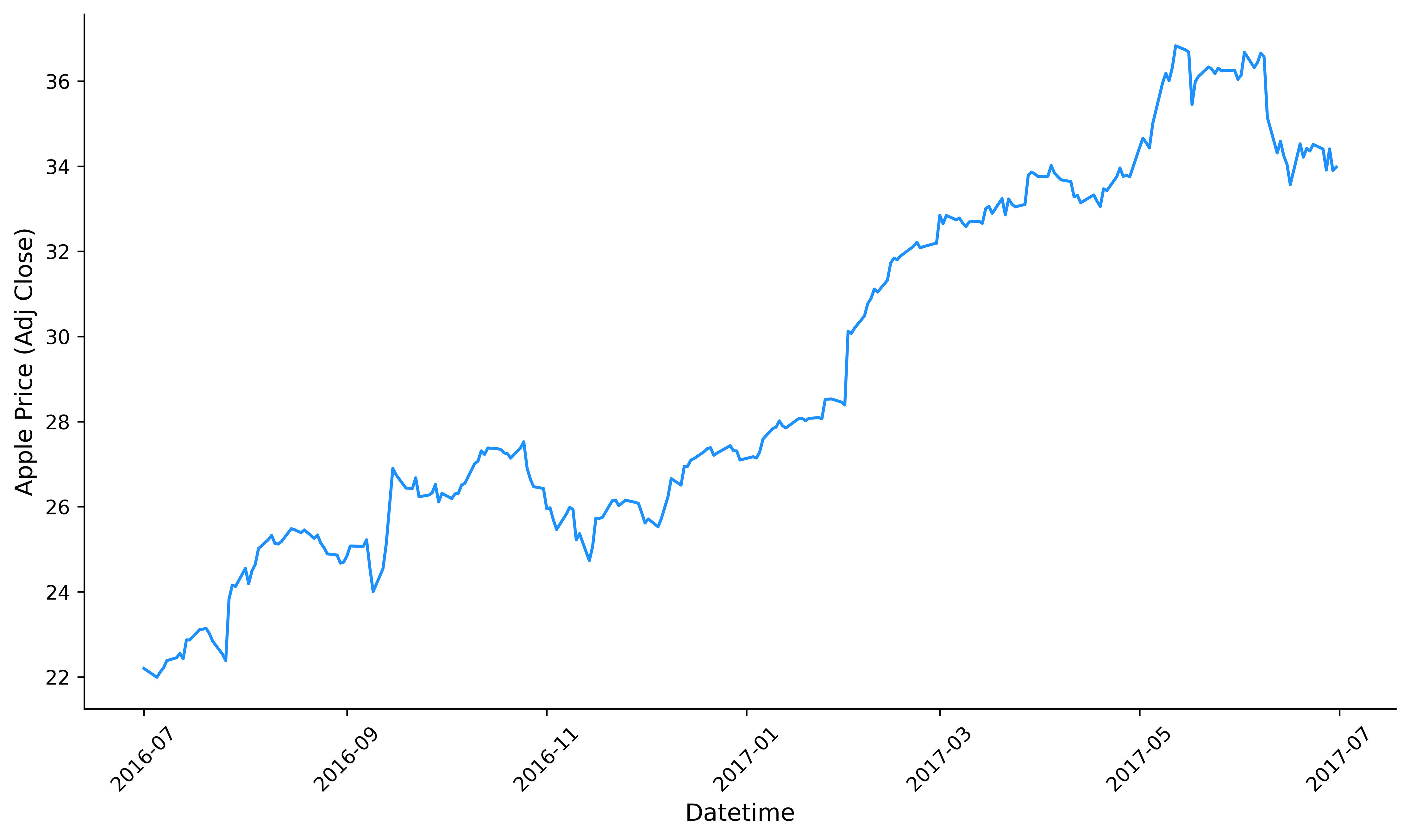}
        \caption{\small{Apple's stock price: 7/1/2016 - 7/1/2017}}
        \label{fig:apple_price}
    \end{minipage}
\end{figure}

We consider the Poisson equation:
$$\nabla^2 v(x, y) = f(x, y), \quad f(x, y) = -2\pi^2 \sin(\pi x) \sin(\pi y),$$
with the boundary conditions,$v(-1, y) = v(1, y) = v(x, -1) = v(x, 1) = 0.$ The PDE has the explict solution, $v(x,y)={\rm sin}(\pi x){\rm sin}(\pi y)$, as shown in the figure \ref{fig:poisson}.
In the subsection, we aim to compare the performance of three neural network architectures: PINN, KAN, and XNet.

\textbf{Time Series Prediction:} The proficiency of the models in capturing temporal dynamics and dependencies is explored through:
\begin{itemize}
    \item The use of both synthetic and real-world time series datasets, which range from financial market data to weather forecasting, focusing on predictive accuracy, response time, and robustness at various temporal scales.
\end{itemize}

we also conducted time series forecasting experiments in different scenarios. 
One scenario is driven by mathematical and physical models.
The example we provide is Apple's stock close price (adj) from the U.S. market, with the test period spanning from July 1, 2016 to July 1, 2017, as shown in the figure \ref{fig:apple_price}.

\textbf{Data Sets and Implementation Details:} Detailed descriptions
of the datasets is provided in Section 3.7. Additionally,
implementation specifics such as hyperparameter settings, training
procedures, and computational resources used are documented to ensure
the experiments' reproducibility and transparency.

\section{RESULTS}\label{sec:cauchy}
In Section \ref{sec:1d}, we perform the heaviside function approximation tasks using KAN and XNet. In Section \ref{sec:2d}, we conduct 2D smooth function approximation tasks using KAN and XNet. Section \ref{sec:nd} evaluates the approximation of high-dimensional functions. In Section \ref{sec:application}, we employ PINN, KAN, and XNet to construct physics-informed machine learning models for solving the 2D Poisson equation.
In Section \ref{sec:LSTM}, we apply XNet to improve the performance of LSTM across various scenarios, then compare with KAN.

\subsection{Heaviside step function apprxiamtion}\label{sec:1d}
The experimental comparison between XNet, B-spline, and KAN demonstrates XNet's superior approximation ability. Except for the first example, all other examples are from the referenced article, with KAN settings matching those from the original experiments. This ensures a fair comparison, fully proving that XNet has stronger approximation capabilities in various benchmarks.

\begin{table}[h!]
\centering
\small
\setlength{\tabcolsep}{8pt}
\begin{tabular}{cccc} 
\toprule
\textbf{Metric} & \textbf{MSE} & \textbf{RMSE} & \textbf{MAE} \\
\midrule
\textbf{XNet with 64 basis functions} & 8.99e-08 & 3.00e-04 & 1.91e-04 \\
\textbf{[1,1]KAN with 200 grids} & 5.98e-04 & 2.45e-02 & 3.03e-03 \\
\bottomrule
\end{tabular}
\caption{Performance comparison between XNet and KAN.}
\label{tab:xnet_kan_performance_1d}
\end{table}

\begin{figure}[h!]
    \centering
    \begin{minipage}[b]{0.45\textwidth}
        \centering
        \includegraphics[width=\textwidth]{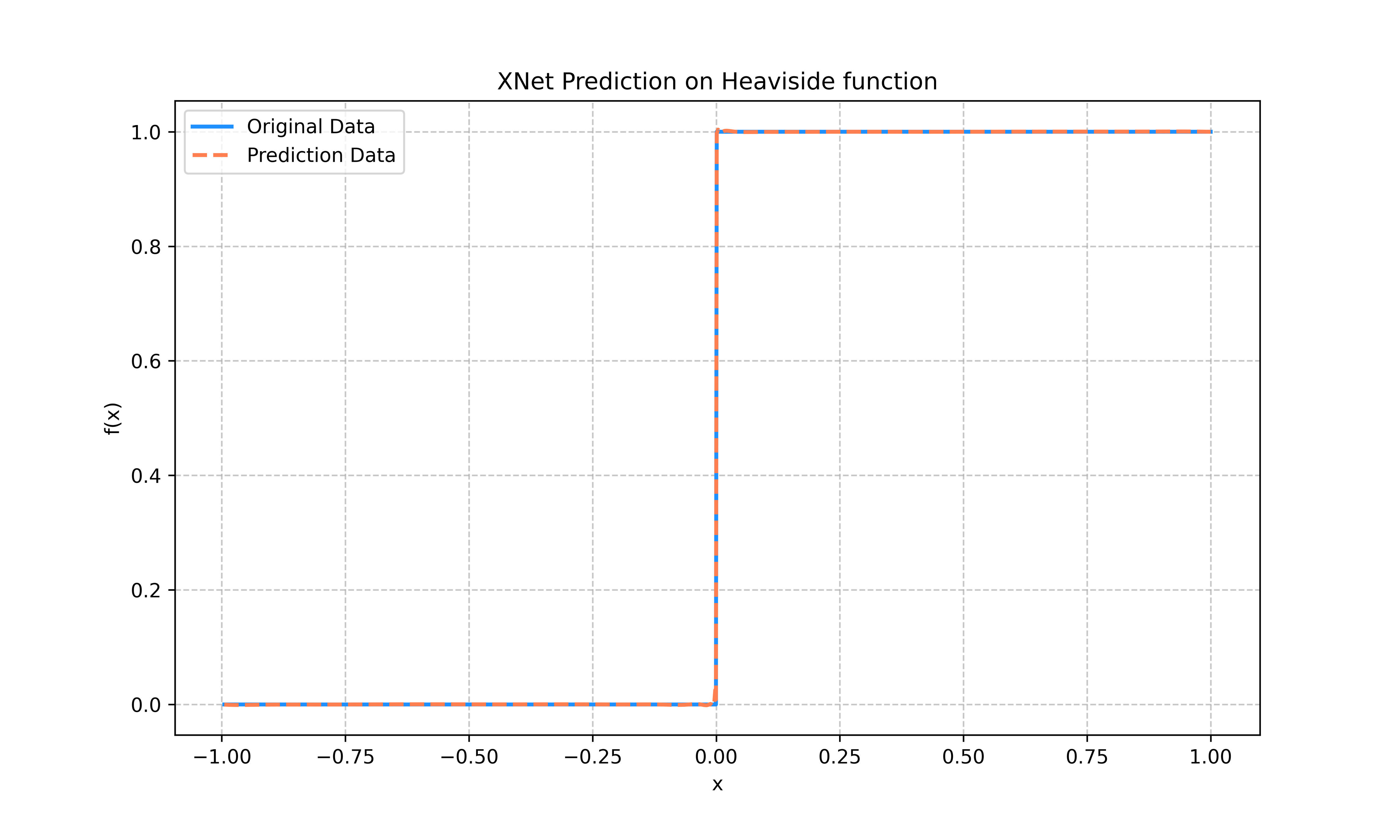}
	\caption{XNet approximation, with 64 basis functions}
    \end{minipage}
    \hfill
    \begin{minipage}[b]{0.45\textwidth}
        \centering
        \includegraphics[width=\textwidth]{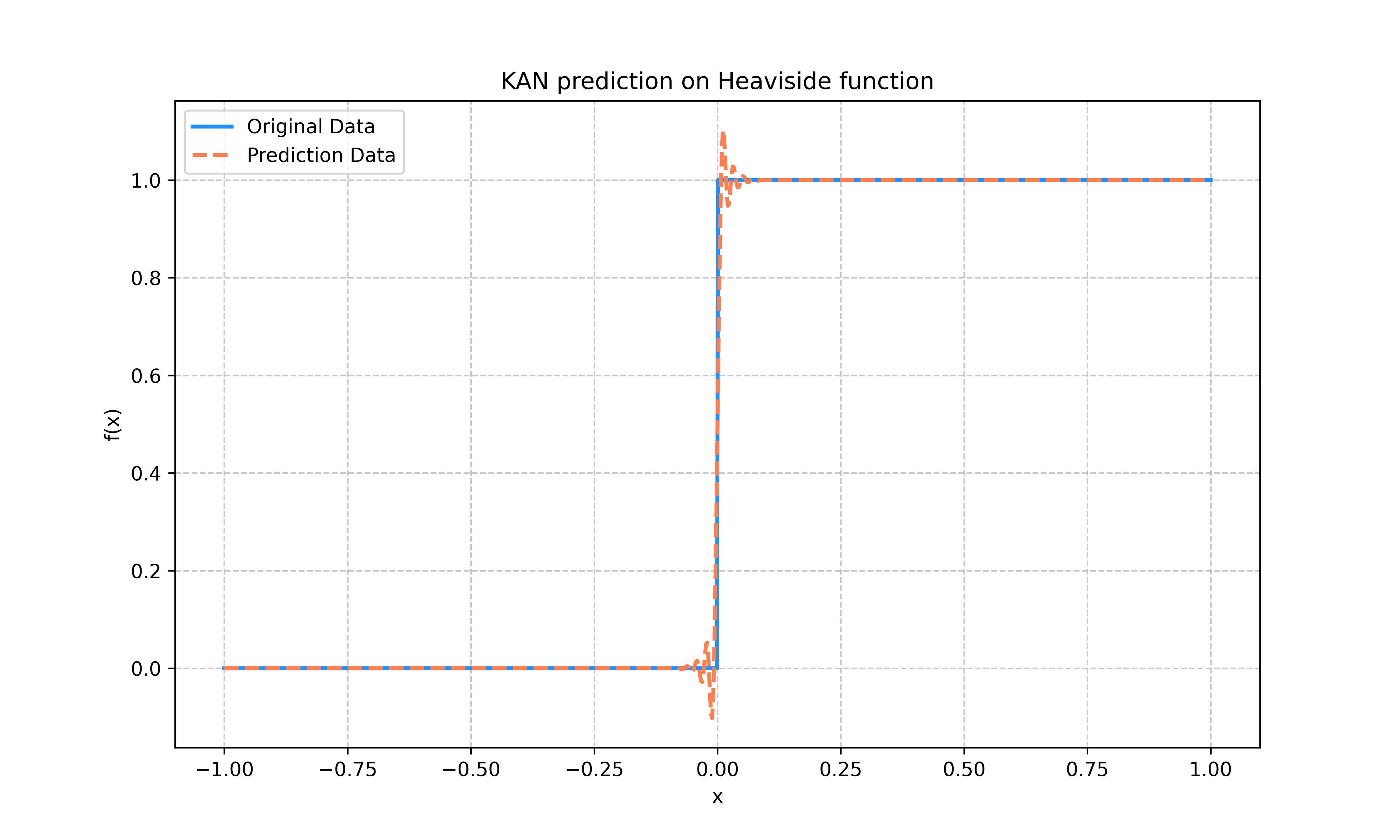}
	\caption{[1,1] KAN approximation, with k=3, grid =200 }
    \end{minipage}
\end{figure}

\begin{figure}[h!]
    \centering
    \begin{minipage}[b]{0.45\textwidth}
        \centering
        \includegraphics[width=\textwidth]{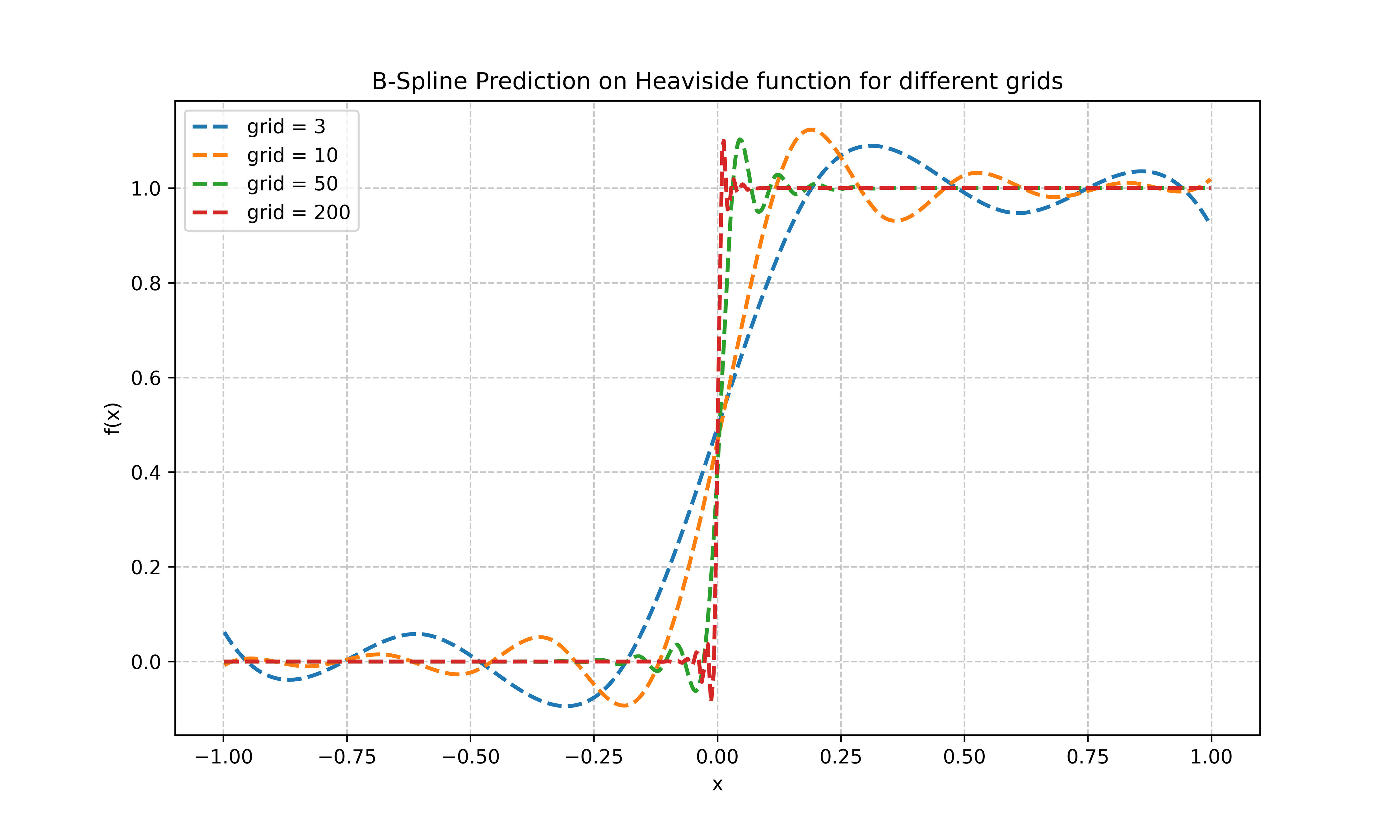}
        \caption{B-Spline comparision, with k=3}
	\label{B_Spline1}
    \end{minipage}
    \hfill
    \begin{minipage}[b]{0.45\textwidth}
        \centering
        \includegraphics[width=\textwidth]{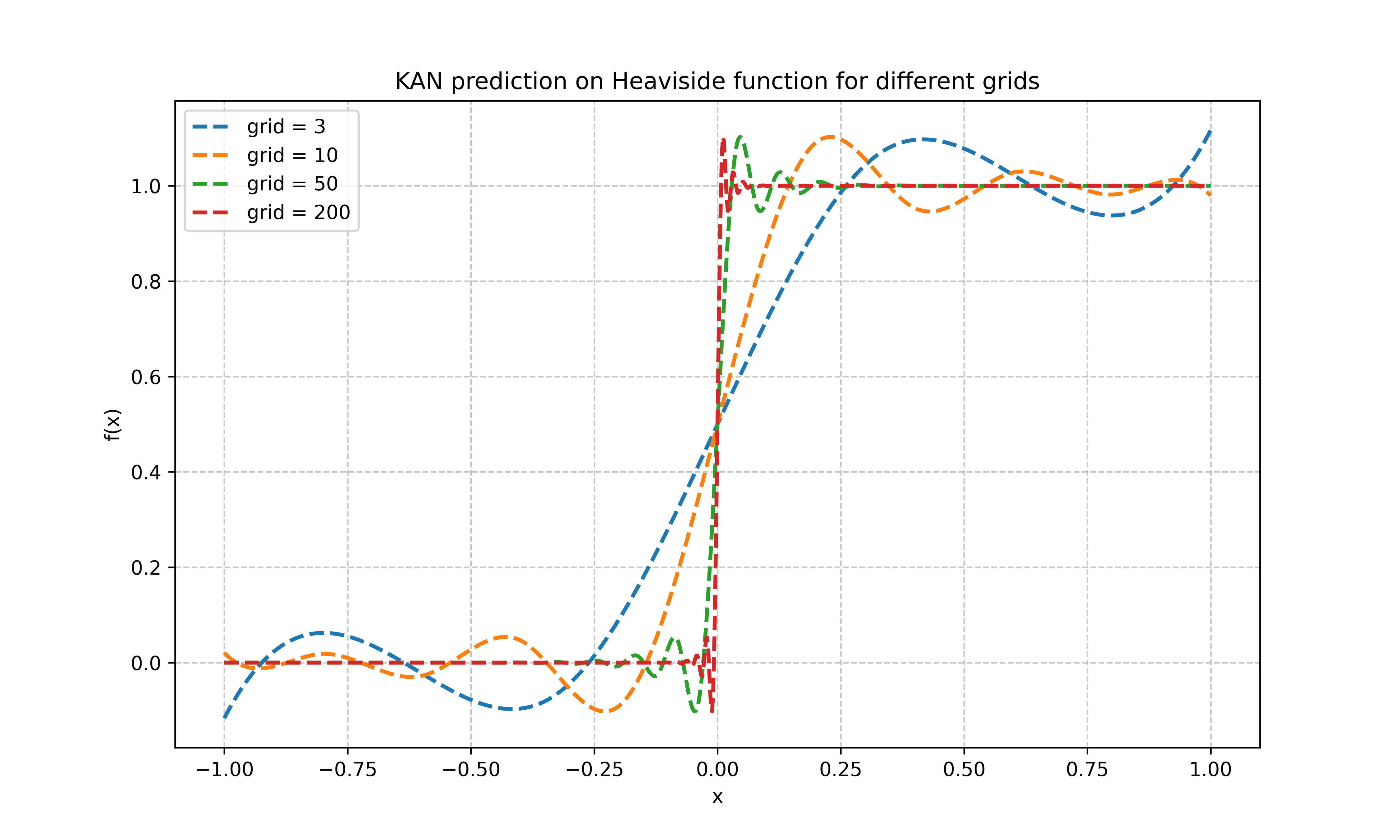}
        \caption{[1,1] KAN comparision, with k=3}
	\label{KAN_1d}
    \end{minipage}
\end{figure}

As shown in Figure \ref{B_Spline1} and \ref{KAN_1d}, both B-Spline and KAN exhibit "overshoot," leading to local oscillations at discontinuities. We speculate that this is due to the fact that a portion of KAN's output is represented by B-Splines. 
While adjusting the grid can alleviate this phenomenon, it introduces complexity in tuning parameters (see Table \ref{table:kan_1d} in appendix A.1). In contrast, XNet demonstrates superior performance, providing smooth transitions at discontinuities. Notably, in terms of fitting accuracy in these regions, XNet's MSE is 1,000-fold times smaller than that of KAN.

\subsection{Function Approximation with \(\exp(\sin(\pi x) + y^2)\) and \(xy\) }\label{sec:2d}
The function used is \( f(x, y) = \exp(\sin(\pi x) + y^2) \). Following the procedure described in the article, 1,000 points were used for training and another 1,000 points for testing. 
After sufficient training, the model's predictions were evaluated on a \(100 \times 100\) grid. The KAN structure consists of a two hidden layer with configuration [2, 1, 1], We compare its computational efficiency with the XNet model using two examples: \(\exp(\sin(\pi x) + y^2)\) and \(xy\) .

\begin{figure}[h!]
    \centering
    \begin{minipage}[b]{0.48\textwidth}
        \centering
        \includegraphics[width=0.48\textwidth]{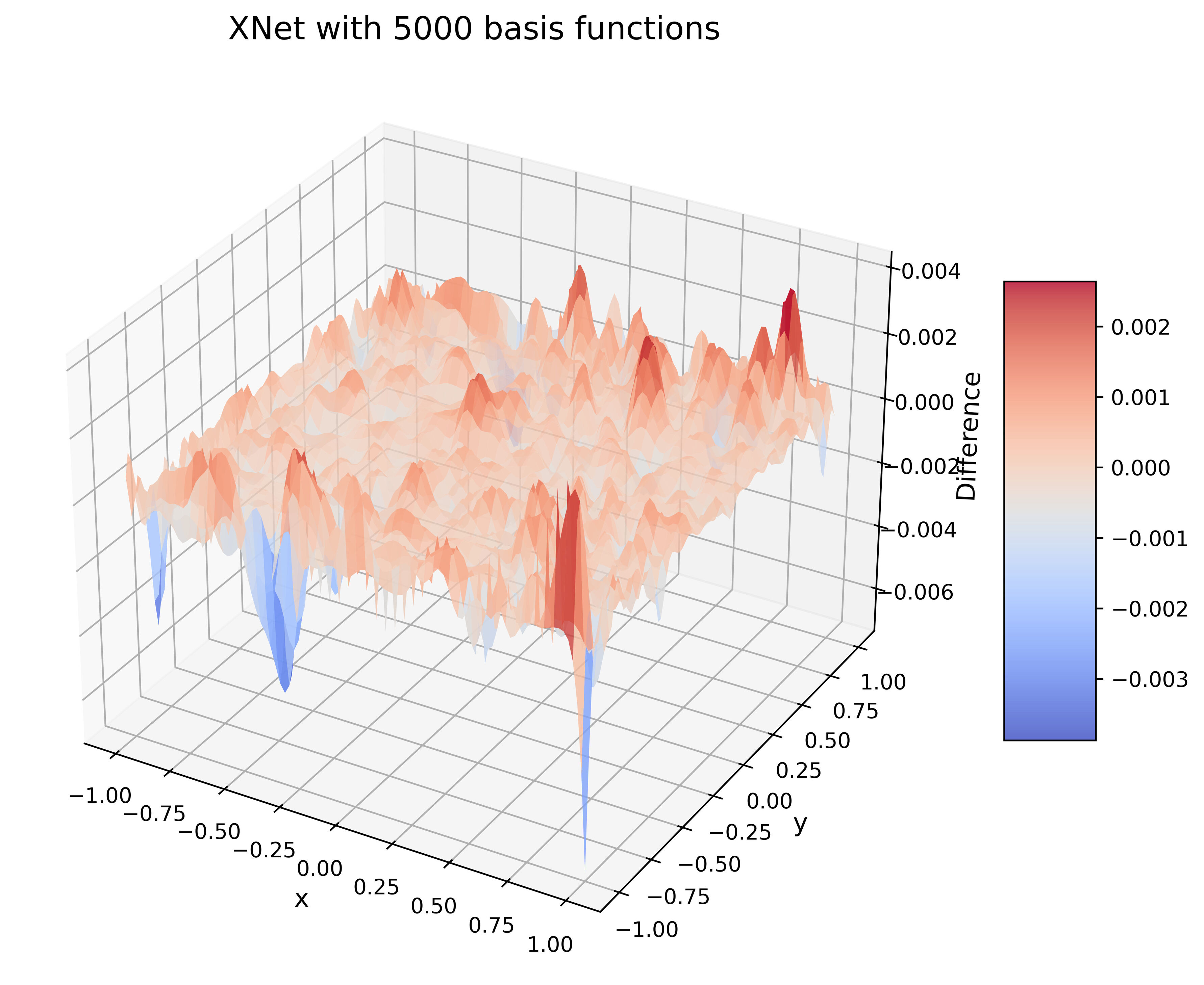}
        \hfill
        \includegraphics[width=0.48\textwidth]{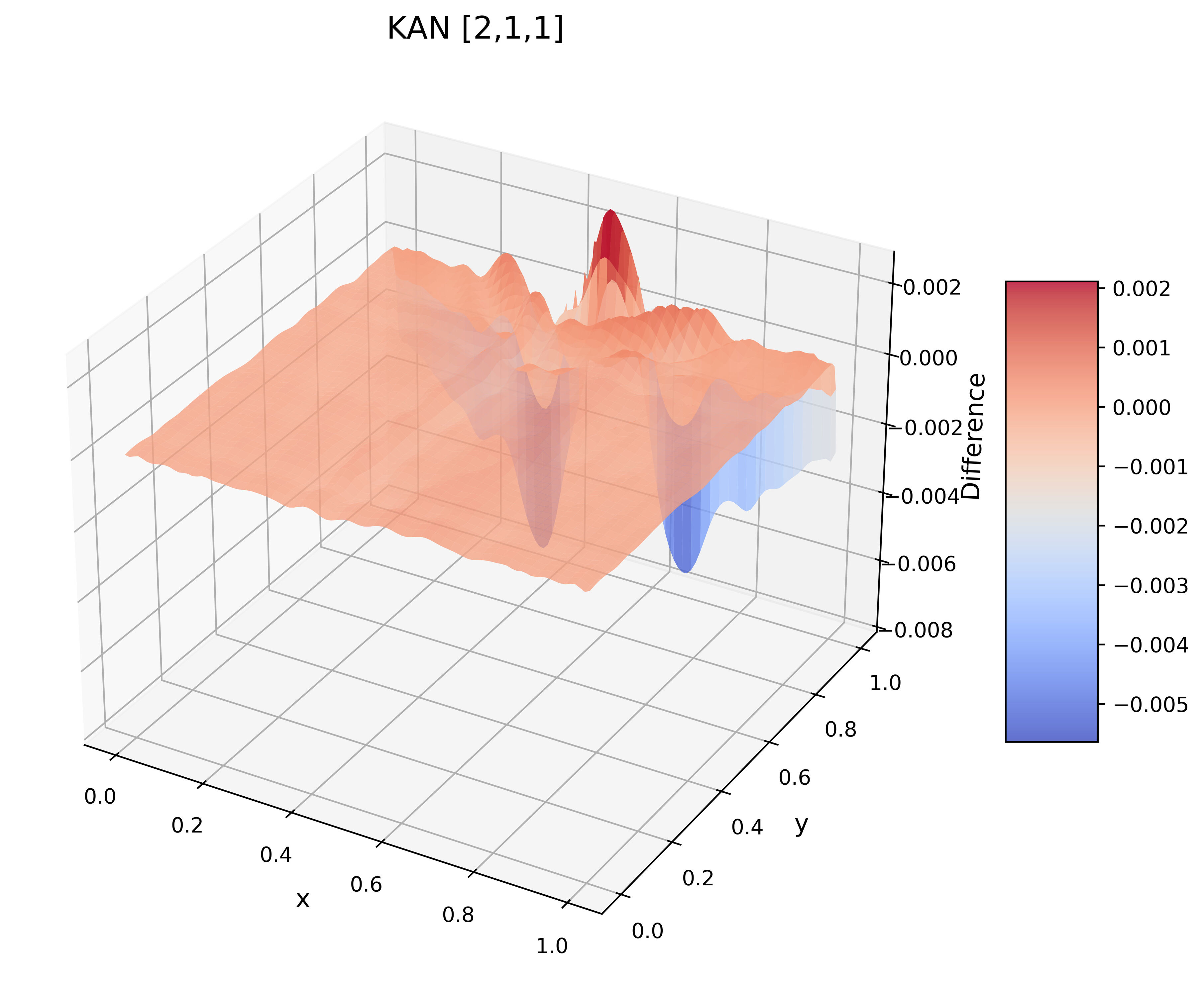}
        \captionof{figure}{Difference on \(\exp(\sin(\pi x) + y^2)\)}
    \end{minipage}
    \hfill
    \begin{minipage}[b]{0.48\textwidth}
        \centering
        \includegraphics[width=0.48\textwidth]{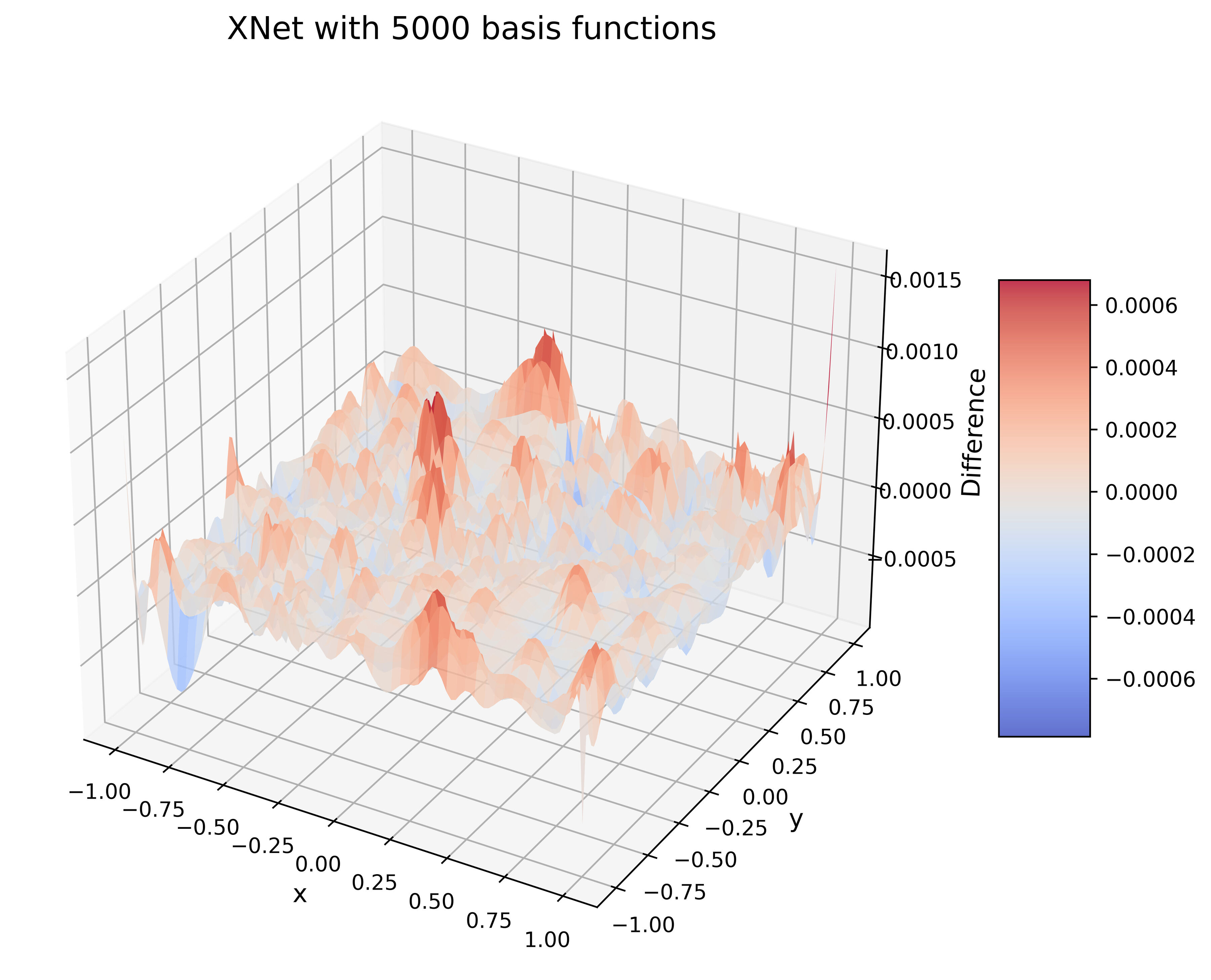}
        \hfill
        \includegraphics[width=0.48\textwidth]{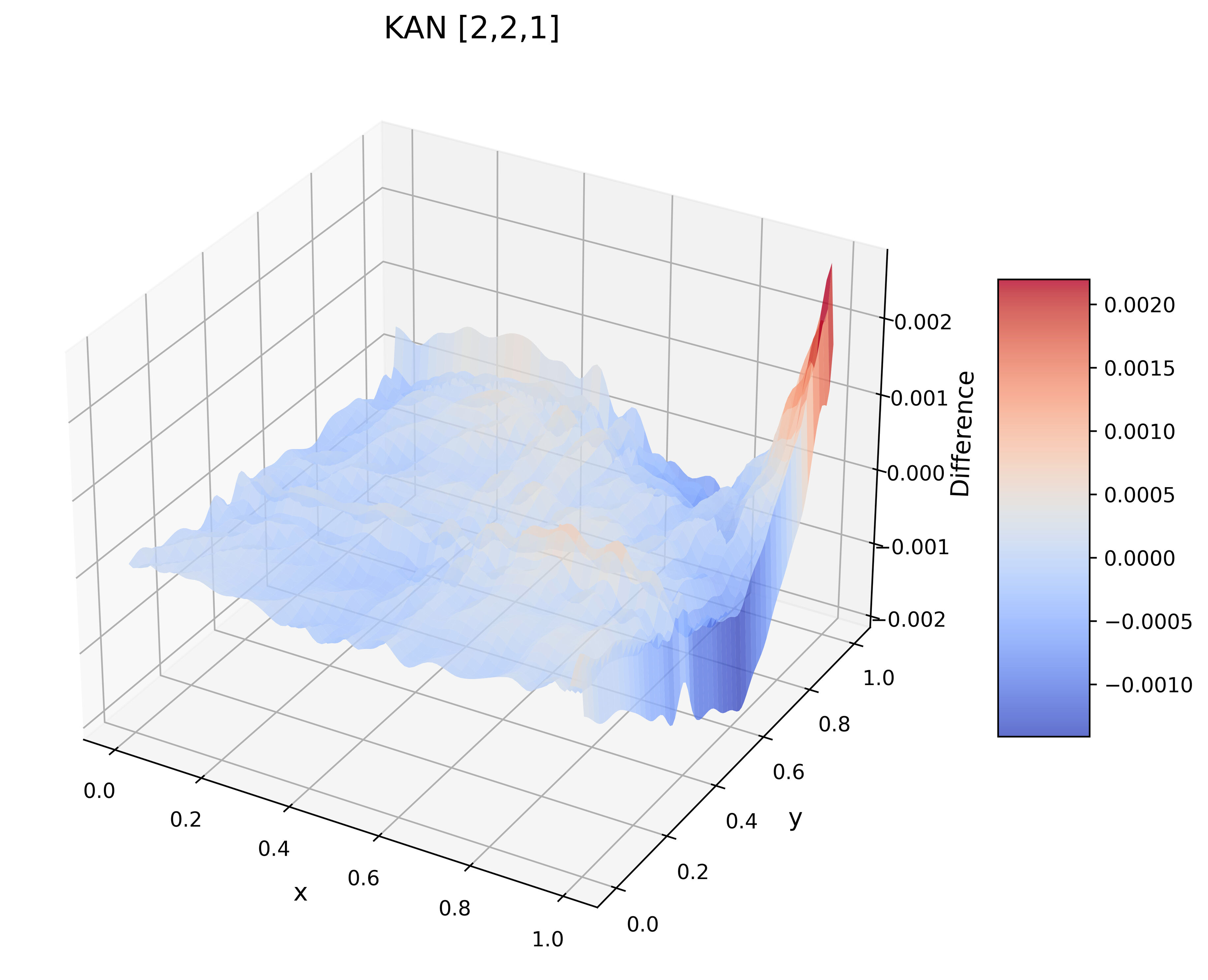}
        \captionof{figure}{Difference on \(xy\)}
	\label{fig:2d_difference}
    \end{minipage}
\end{figure}

Following the official model configurations,
XNet with 5,000 basis functions is trained with adam, while KAN is initialized to have G = 3, trained with LBFGS, with increasing number of grid points every 200 steps to cover G = {3, 5, 10, 20, 50}.
Overall, both networks performed similarly on these two-dimensional examples (see Table \ref{table:2d_1} and \ref{table:2d_2}). However, XNet produced a more uniform fit, with no significant local oscillations (see Figure \ref{fig:2d_difference}). In contrast, KAN exhibited sharp variations in certain regions, consistent with the behavior observed in the heaviside step function (see Section \ref{sec:1d}).

\begin{table}[!ht]
\centering
\small 
\caption{Comparison of XNet and KAN on \(\exp(\sin(\pi x) + y^2)\).}
\label{table:2d_1}
\begin{tabular}{ccccc}
\toprule
\textbf{Metric} & \textbf{MSE} & \textbf{RMSE} & \textbf{MAE} & \textbf{Time (s)} \\
\midrule
\textbf{XNet (5000)} & 3.9767e-07 & 6.3061e-04 & 4.0538e-04 & 61.0 \\ 
\textbf{KAN[2,1,1]} & 3.0227e-07 & 5.4979e-04 & 1.6344e-04 & 56.1 \\ 
\bottomrule
\end{tabular}
\end{table}

\begin{table}[!ht]
\centering
\small 
\caption{Comparison of XNet and KAN on \(xy\).}
\label{table:2d_2}
\begin{tabular}{ccccc}
\toprule
\textbf{Metric} & \textbf{MSE} & \textbf{RMSE} & \textbf{MAE} & \textbf{Time (s)} \\
\midrule
\textbf{XNet (5000)} & 2.1544e-08 & 1.4678e-04 & 1.0439e-04 & 61.8 \\ 
\textbf{KAN[2,2,1]} & 4.9306e-08 & 2.2205e-04 & 1.4963e-04 & 62.4 \\ 
\bottomrule
\end{tabular}
\end{table}

\subsection{Approximation with high-dimensional functions}\label{sec:nd}
We continue to compare the approximation capabilities of KAN and XNet in solving high-dimensional functions. Following the procedure described in the article, 8000 points were used for training and another 1000 points for testing. XNet is trained with adam, while KAN is initialized to have G = 3, trained with LBFGS, with increasing number of grid points every 200 steps to cover G = {3, 5, 10, 20, 50}.

First, we consider the four-dimensional function
\( \exp\left(\frac{1}{2}\left(\sin\left(\pi(x_{1}^{2}+x_{2}^{2})\right) + x_{3}x_{4}\right)\right) \). For this case, the KAN structure is configured as [4,4,2,1], while XNet is equipped with 5,000 basis functions. Under the same number of iterations, XNet achieves higher accuracy in less time (see Table \ref{table:4d_1}), the MSE is 1,000-fold smaller than that of KAN.

\begin{table}[!ht]
\centering
\small 
\caption{Comparison of XNet and KAN on \(  \exp\left(\frac{1}{2}\left(\sin\left(\pi(x_{1}^{2}+x_{2}^{2})\right) + x_{3}x_{4}\right)\right) \).}
\label{table:4d_1}
\begin{tabular}{ccccc} 
\toprule
\textbf{Metric}  & \textbf{MSE} & \textbf{RMSE} & \textbf{MAE} & \textbf{Time (s)} \\
\midrule
\textbf{XNet (5,000)} & 2.3079e-06 & 1.5192e-03 & 8.3852e-04 & 78.18 \\ 
\textbf{KAN [4,2,2,1]} & 2.6151e-03 & 5.1138e-02 & 3.6300e-02 & 143.1 \\ 
\bottomrule
\end{tabular}
\end{table}

Next, we consider the 100-dimensional function \( \exp(\frac{1}{100}\sum_{i=1}^{100}\sin^2(\frac{\pi x_i}{2})) \). For this case, the KAN structure is configured as [100,1,1], while XNet has 5,000 basis functions. Under the same number of iterations, XNet achieved higher accuracy in less time compared to KAN (see Table \ref{table:100d_1}).

\begin{table}[!ht]
\centering
\small 
\caption{Comparison of XNet and KAN on \(  \exp\left(\frac{1}{100}\sum_{i=1}^{100}\sin^2\left(\frac{\pi x_i}{2}\right)\right) \).}
\label{table:100d_1}
\begin{tabular}{ccccc} 
\toprule
\textbf{Metric}  & \textbf{MSE} & \textbf{RMSE} & \textbf{MAE} & \textbf{Time (s)} \\
\midrule
\textbf{XNet (5,000)} & 6.8492e-04 & 2.6171e-02 & 2.0889e-02 & 158.69 \\ 
\textbf{KAN [100,1,1]} & 6.5868e-03 & 8.1159e-02 & 6.4611e-02 & 556.5 \\ 
\bottomrule
\end{tabular}
\end{table}

As dimensionality increases, the computational efficiency of KAN decreases significantly, while XNet shows an advantage in this regard. The approximation accuracy of both networks declines with increasing dimensions, which we hypothesize is related to the sampling method and the number of samples used.

\begin{figure}[h!]
    \centering
    \begin{minipage}[b]{0.23\textwidth}
        \centering
        \includegraphics[width=\textwidth]{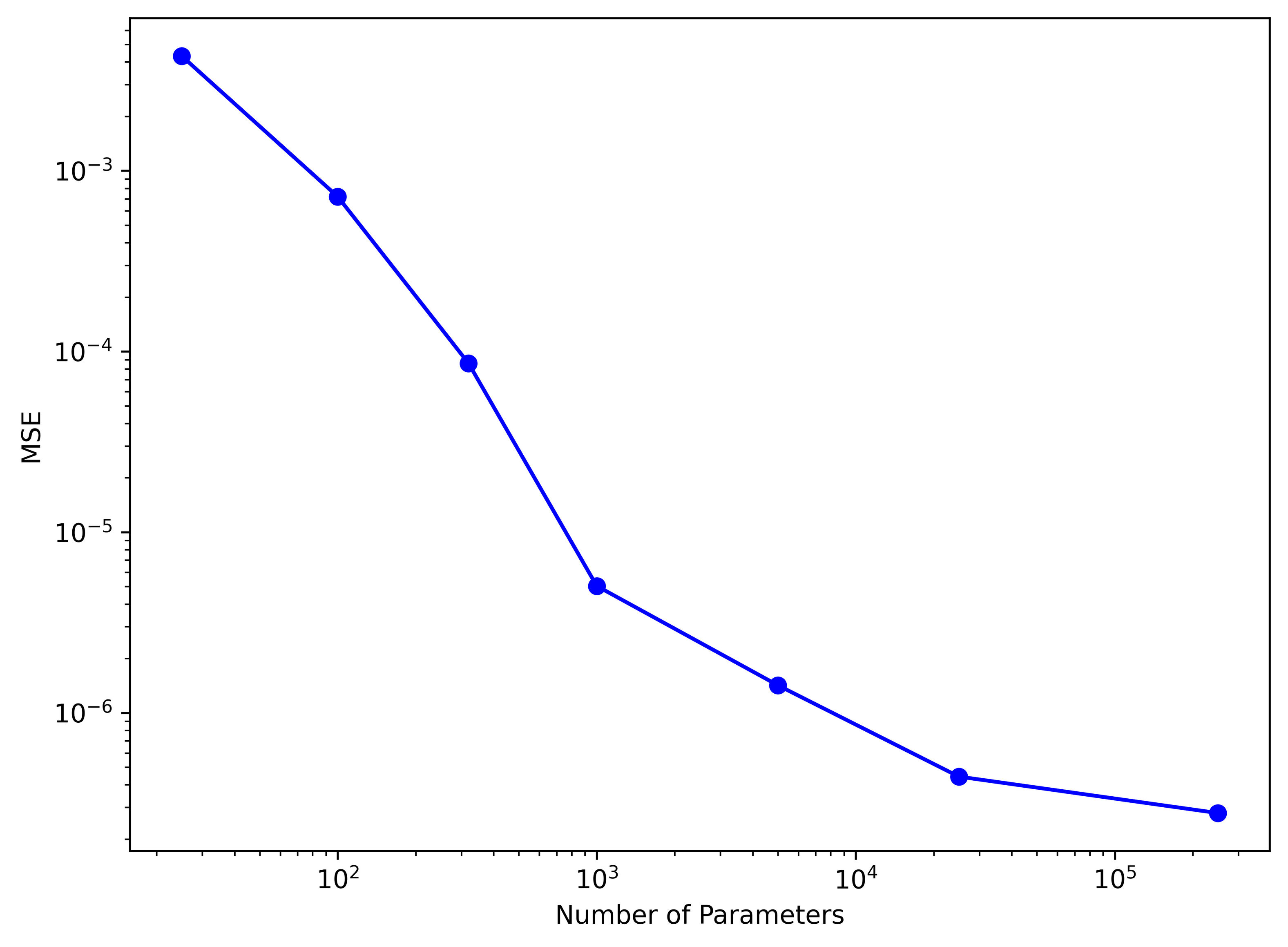}
        \caption*{\tiny \(\exp(\sin(\pi x) + y^2)\)}
    \end{minipage}
    \hfill
    \begin{minipage}[b]{0.23\textwidth}
        \centering
        \includegraphics[width=\textwidth]{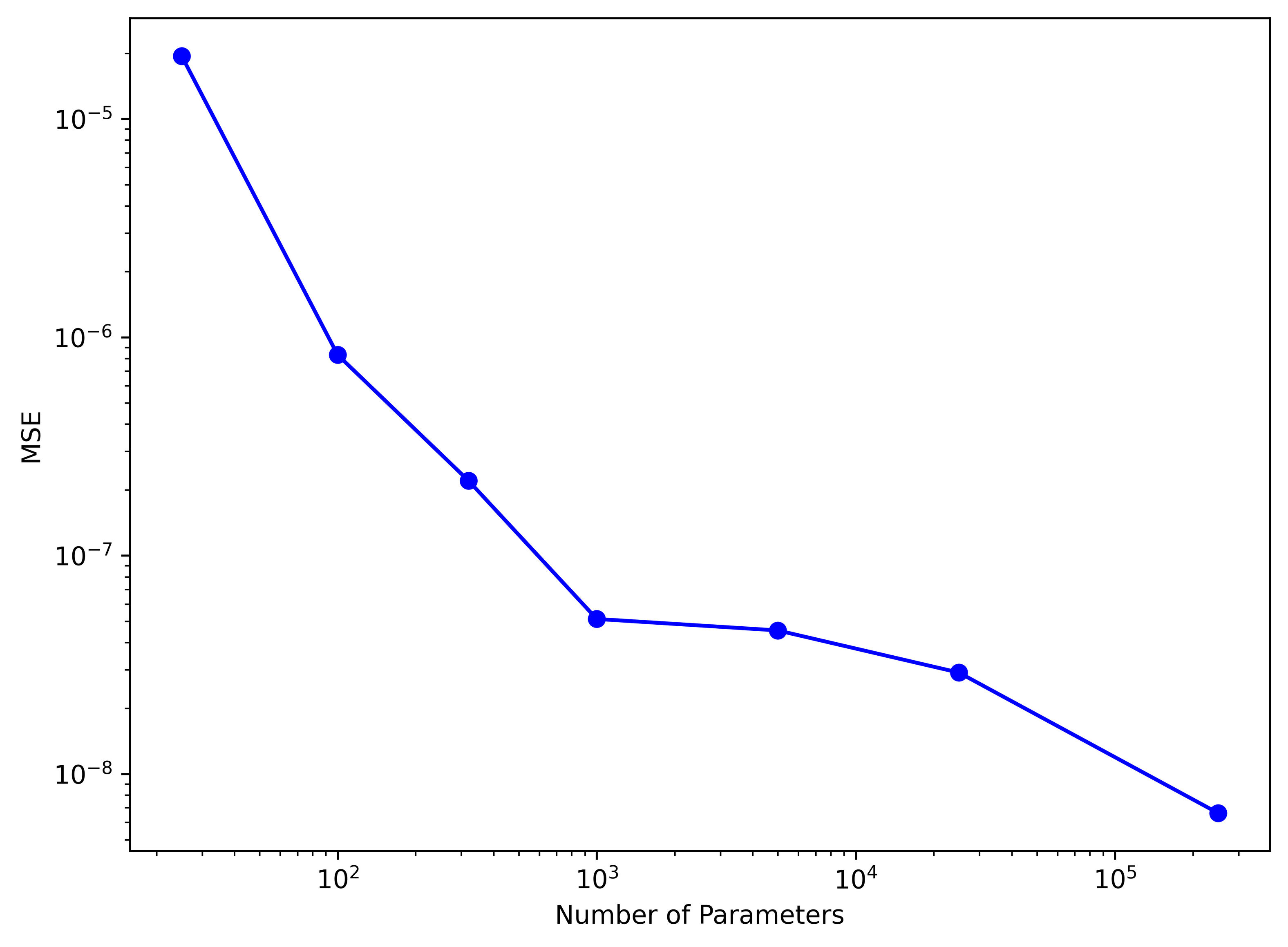}
        \caption*{\tiny \(xy\)}
    \end{minipage}
    \hfill
    \begin{minipage}[b]{0.23\textwidth}
        \centering
        \includegraphics[width=\textwidth]{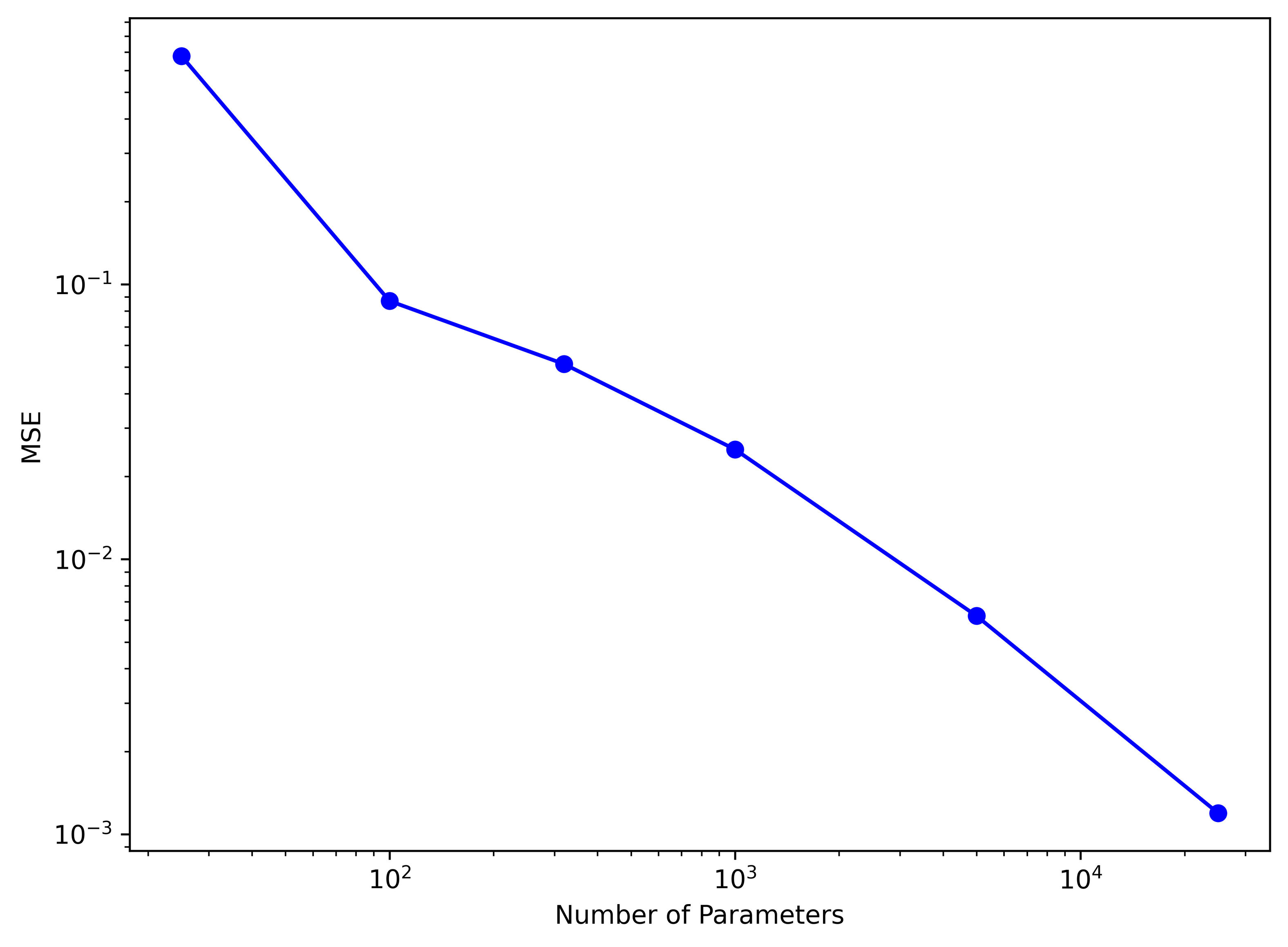}
        \caption*{\tiny \(\exp\left(\frac{1}{2}\left(\sin\left(\pi(x_{1}^{2}+x_{2}^{2})\right) + x_{3}x_{4}\right)\right)\)}
    \end{minipage}
    \hfill
    \begin{minipage}[b]{0.23\textwidth}
        \centering
        \includegraphics[width=\textwidth]{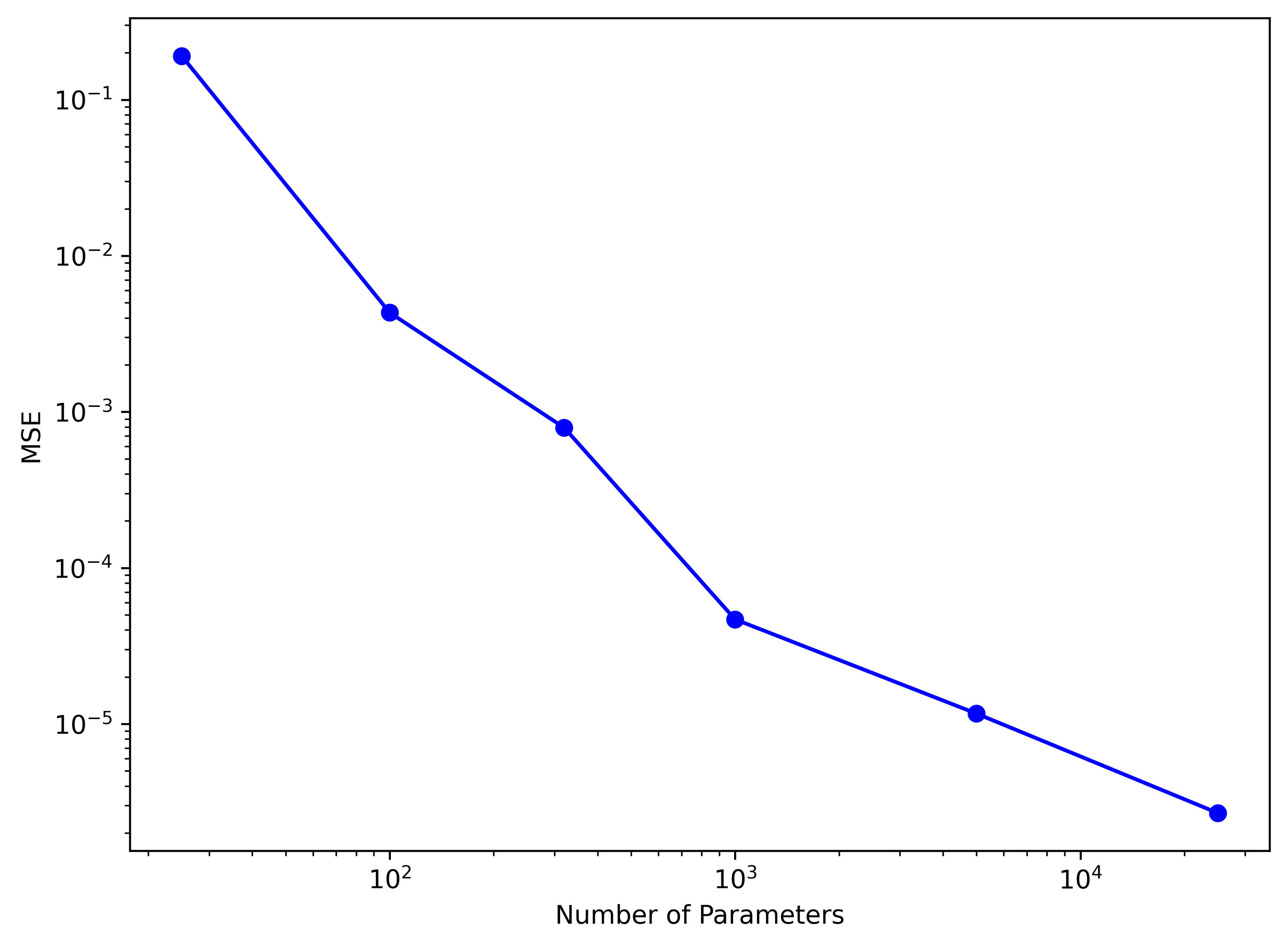}
        \caption*{\tiny \(\exp\left(\frac{1}{100}\sum_{i=1}^{100}\sin^2\left(\frac{\pi x_i}{2}\right)\right)\)}
    \end{minipage}
    \caption{XNet Performance with Number of Parameters}
\label{XNet_params1}
\end{figure}

As shown in Figure \ref{XNet_params1}, XNet achieves high accuracy with relatively few network parameters. Moreover, as the number of parameters increases, XNet can further enhance its accuracy. Given its performance in function approximation tasks, both in terms of computational efficiency and accuracy, we conclude that XNet is a highly efficient neural network with strong approximation capabilities.
Building on this, in the following subsection, we apply PINN, KAN, and XNet to approximate the value function of the Poisson equation.

\subsection{Possion function} \label{sec:application}
We aim to solve a 2D poisson equation $\nabla^2 v(x,y) = f(x,y)$, $f(x,y)=-2\pi^2{\rm sin}(\pi x){\rm sin}(\pi y)$, with boundary condition $v(-1,y)=v(1,y)=v(x,-1)=v(x,1)=0$. The ground truth solution is $v(x,y)={\rm sin}(\pi x){\rm sin}(\pi y)$. 
We use the framework of physics-informed neural networks (PINNs) to solve this PDE, with the loss function given by
$$\mathrm{loss}_{\mathrm{pde}}=\alpha\mathrm{loss}_i+\mathrm{loss}_b:=\alpha\frac{1}{n_i}\sum_{i=1}^{n_i}|v_{xx}(z_i)+v_{yy}(z_i)-f(z_i)|^2+\frac{1}{n_b}\sum_{i=1}^{n_b}v^2\:,$$

where we use loss$_i$to denote the interior loss, discretized and evaluated by a uniform sampling of $n_i$ points $z_i=(x_i,y_i)$ inside the domain, and similarly we use loss$_b$ to denote the boundary loss, discretized and evaluated by a uniform sampling of $n_b$ points on the boundary. $\alpha$ is the hyperparameter balancing the effect of the two terms.

\begin{figure}[h!]
    \centering
    \begin{minipage}[b]{0.23\textwidth}
        \centering
        \includegraphics[width=\textwidth]{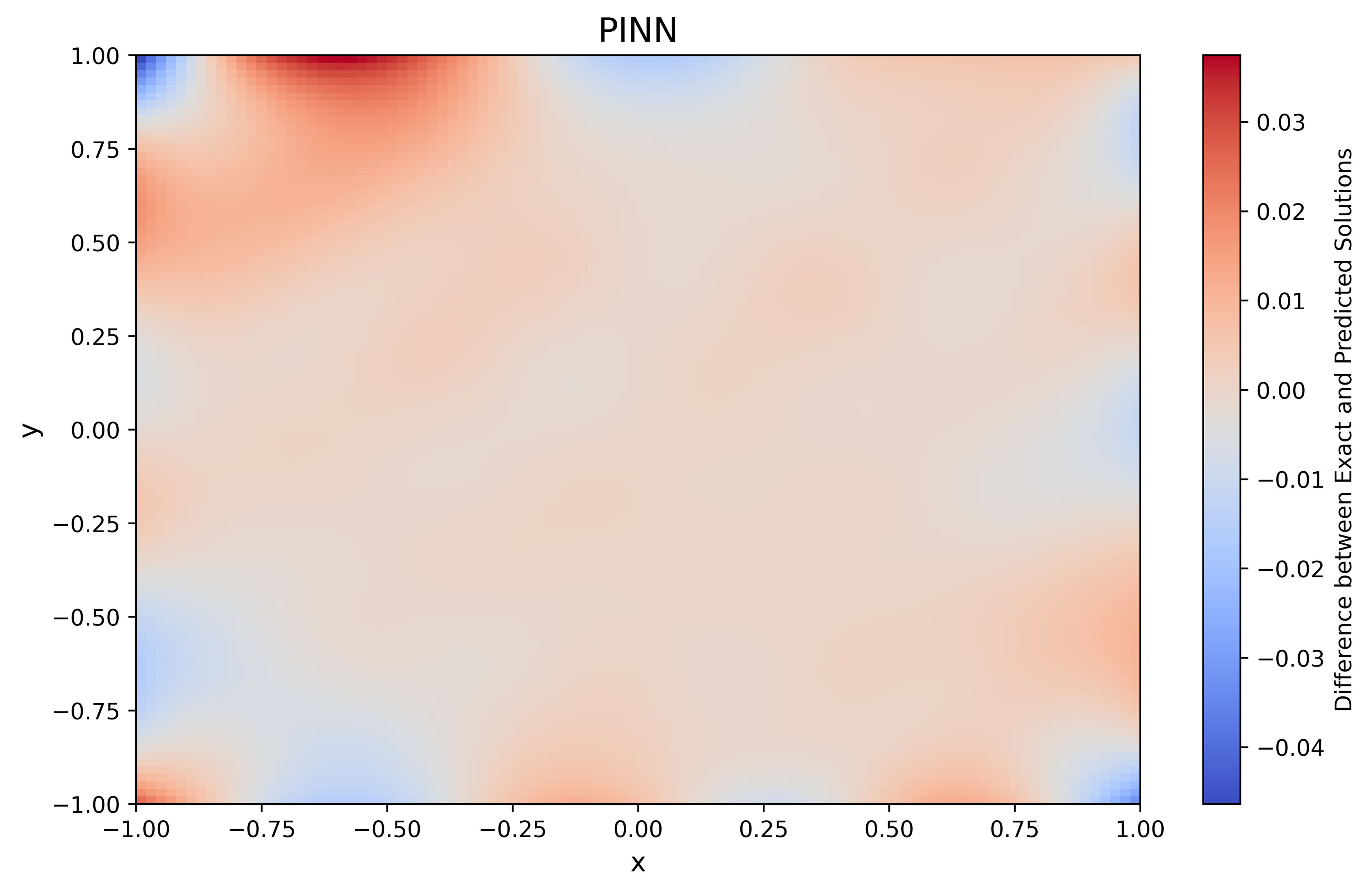}
    \end{minipage}
    \hfill
    \begin{minipage}[b]{0.23\textwidth}
        \centering
        \includegraphics[width=\textwidth]{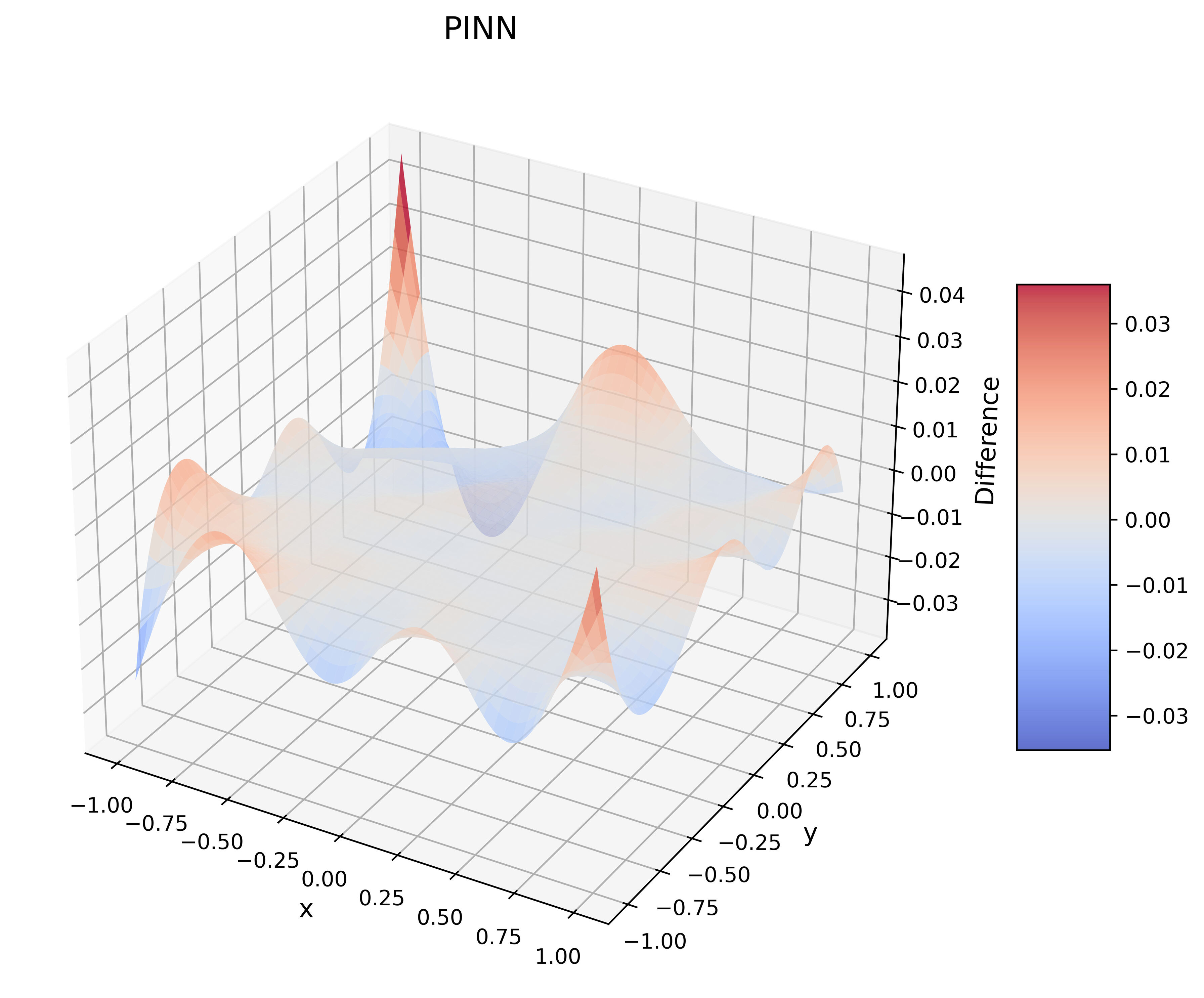}
    \end{minipage}
    \hfill
    \begin{minipage}[b]{0.23\textwidth}
        \centering
        \includegraphics[width=\textwidth]{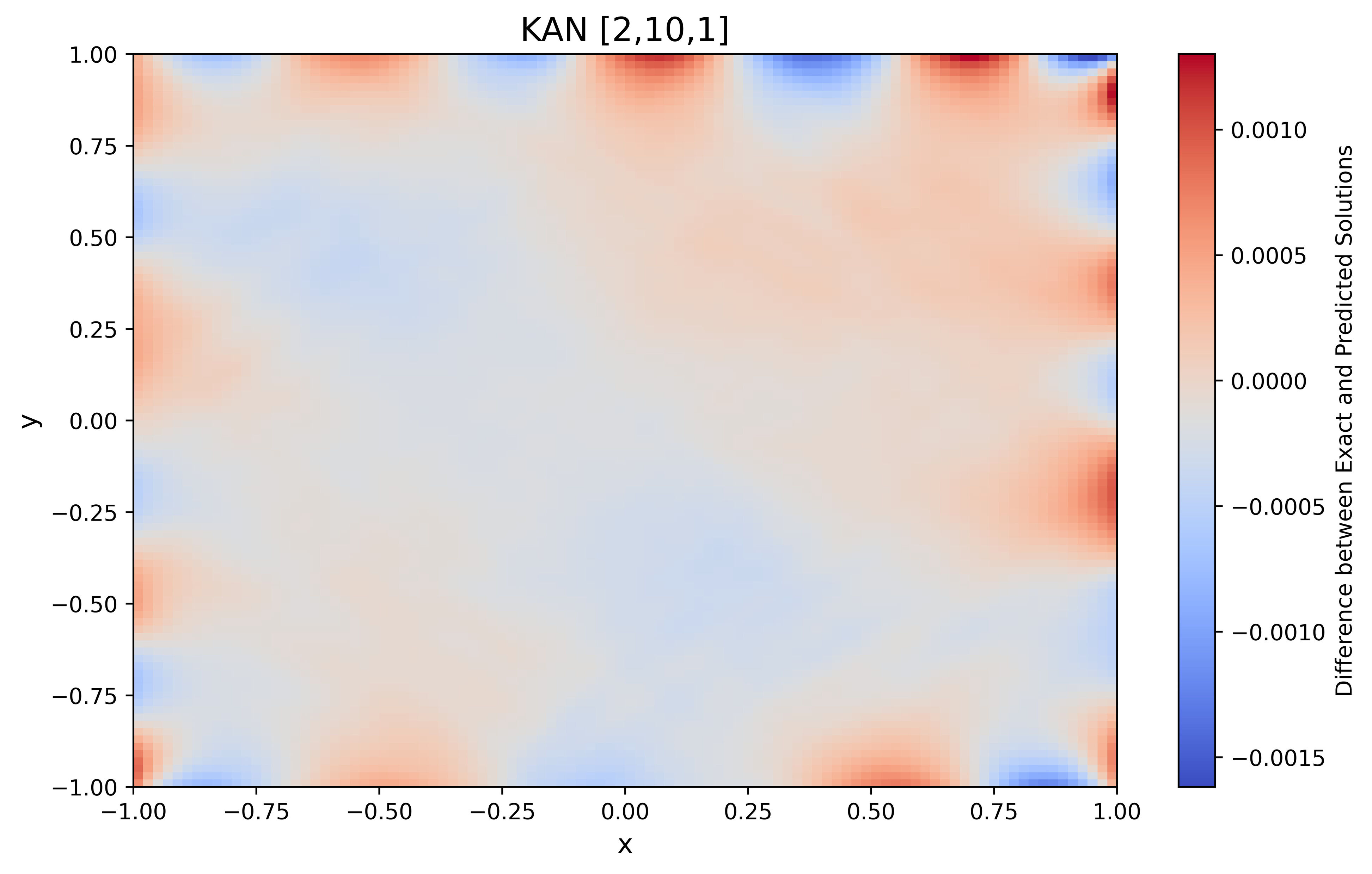}
    \end{minipage}
    \hfill
    \begin{minipage}[b]{0.23\textwidth}
        \centering
        \includegraphics[width=\textwidth]{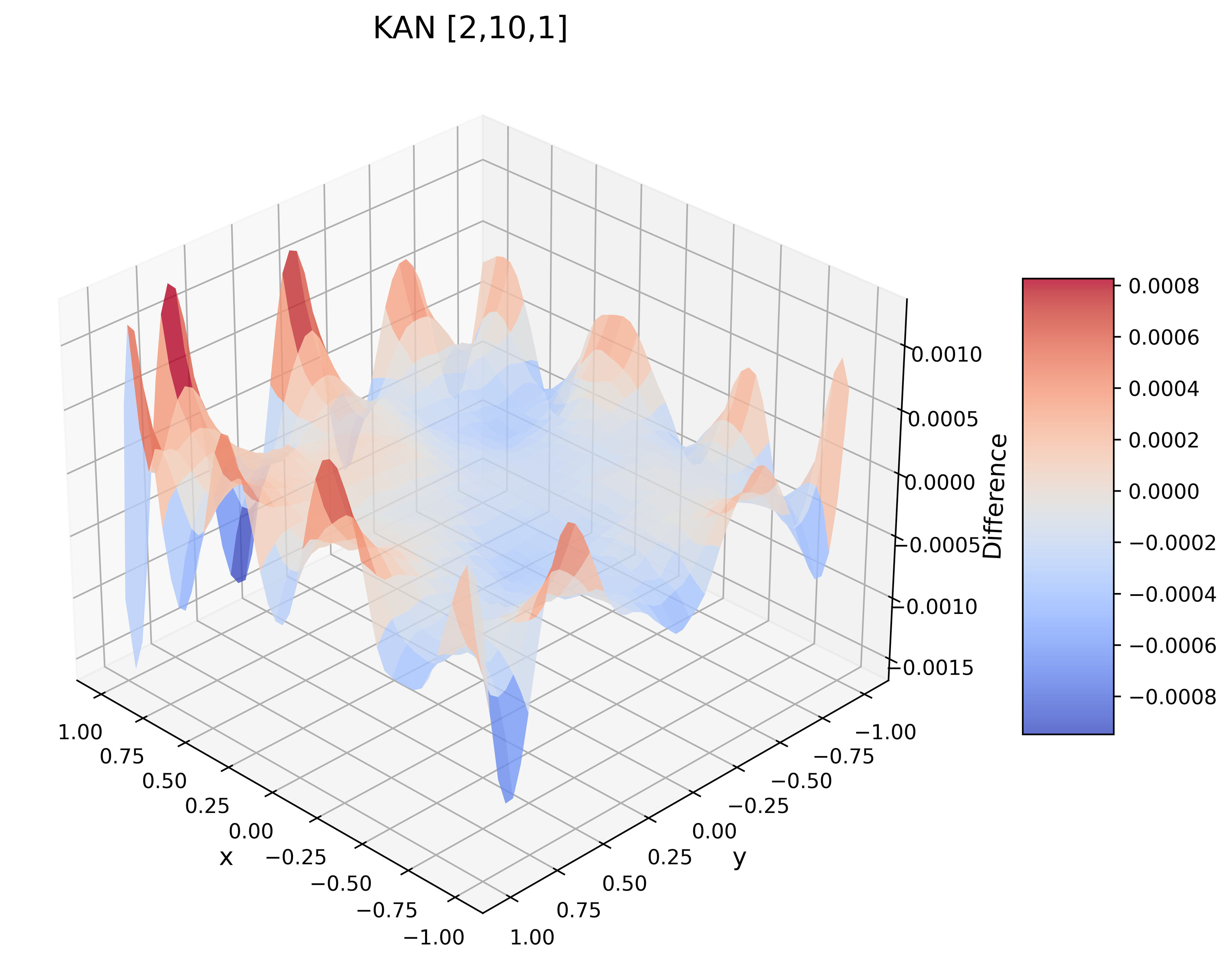}
    \end{minipage}
    \caption{PINN and KAN Performance}
\end{figure}

\begin{figure}[h!]
    \centering
    \begin{minipage}[b]{0.23\textwidth}
        \centering
        \includegraphics[width=\textwidth]{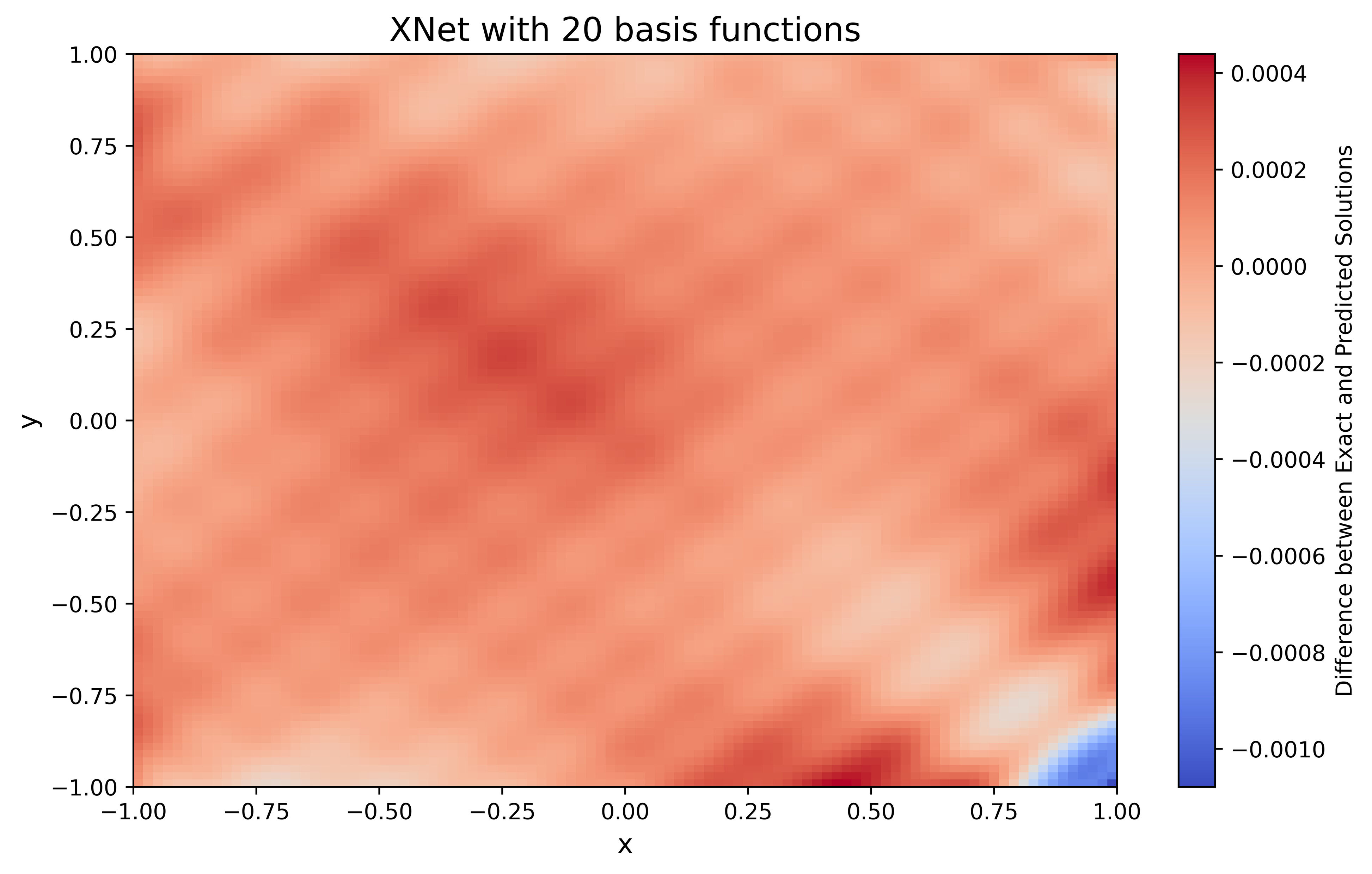}
    \end{minipage}
    \hfill
    \begin{minipage}[b]{0.23\textwidth}
        \centering
        \includegraphics[width=\textwidth]{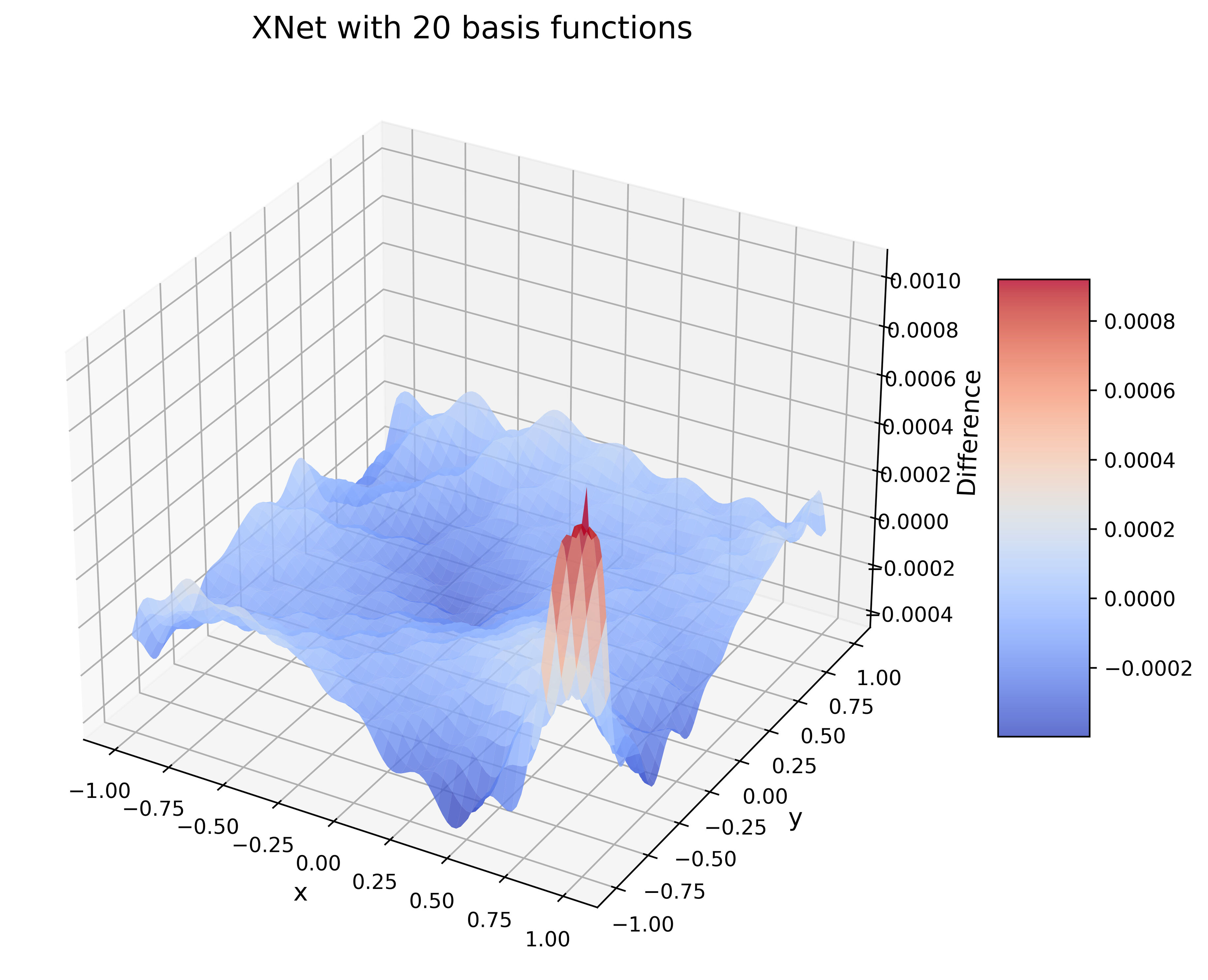}
    \end{minipage}
    \hfill
    \begin{minipage}[b]{0.23\textwidth}
        \centering
        \includegraphics[width=\textwidth]{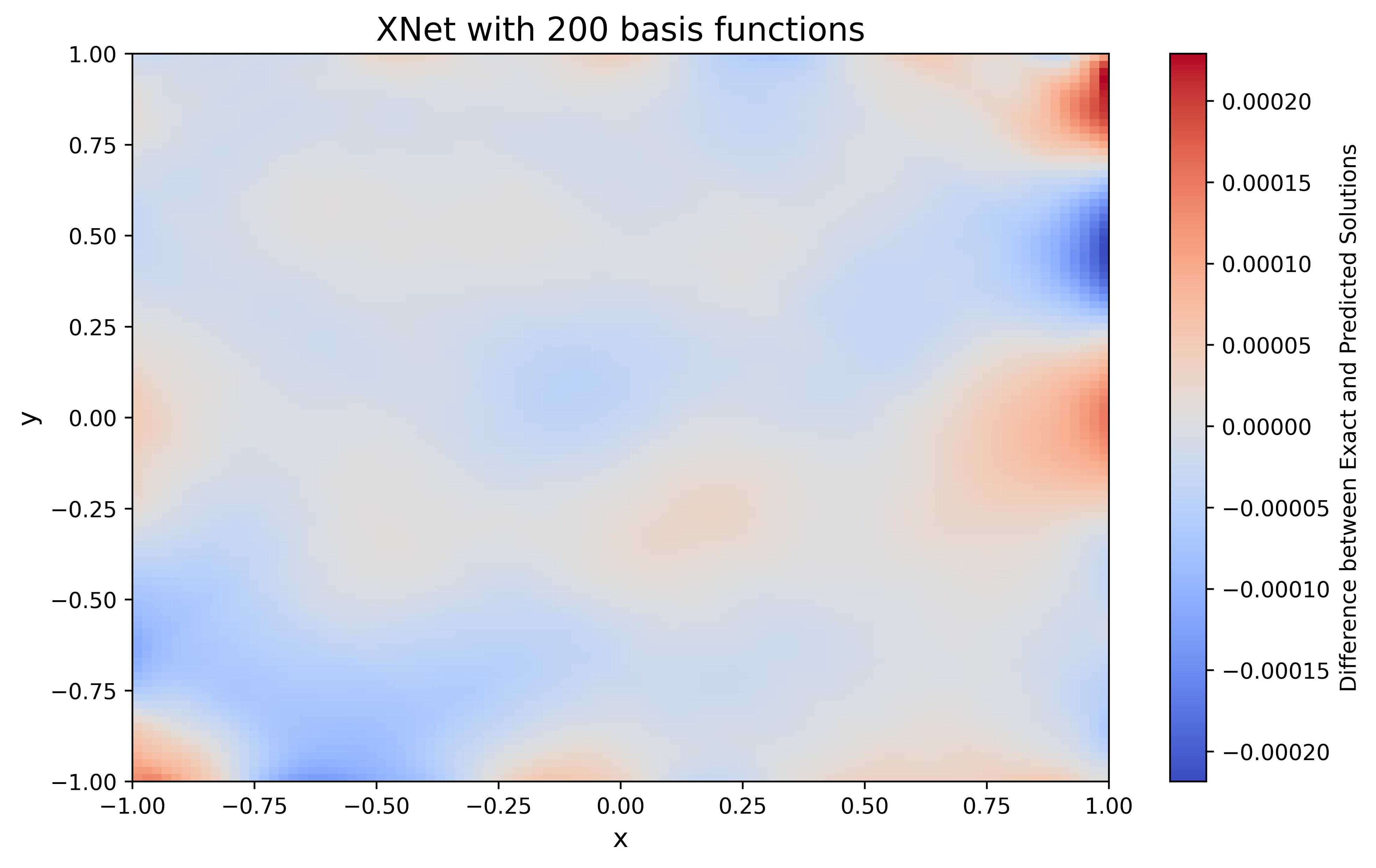}
    \end{minipage}
    \hfill
    \begin{minipage}[b]{0.23\textwidth}
        \centering
        \includegraphics[width=\textwidth]{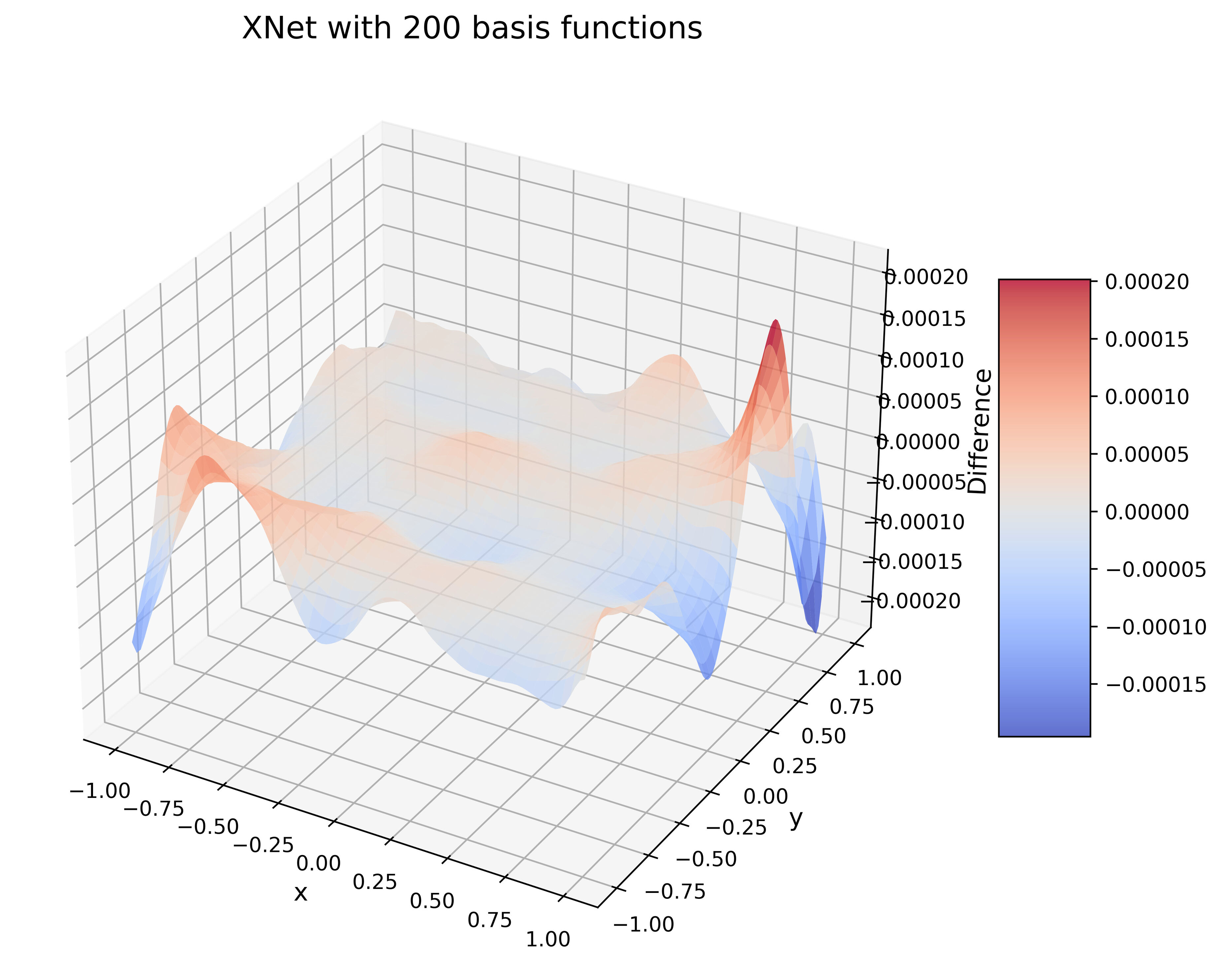}
    \end{minipage}
    \caption{XNet Performance}
\end{figure}

\begin{figure}[h!]
    \centering
    \begin{minipage}[b]{0.23\textwidth}
        \centering
        \includegraphics[width=\textwidth]{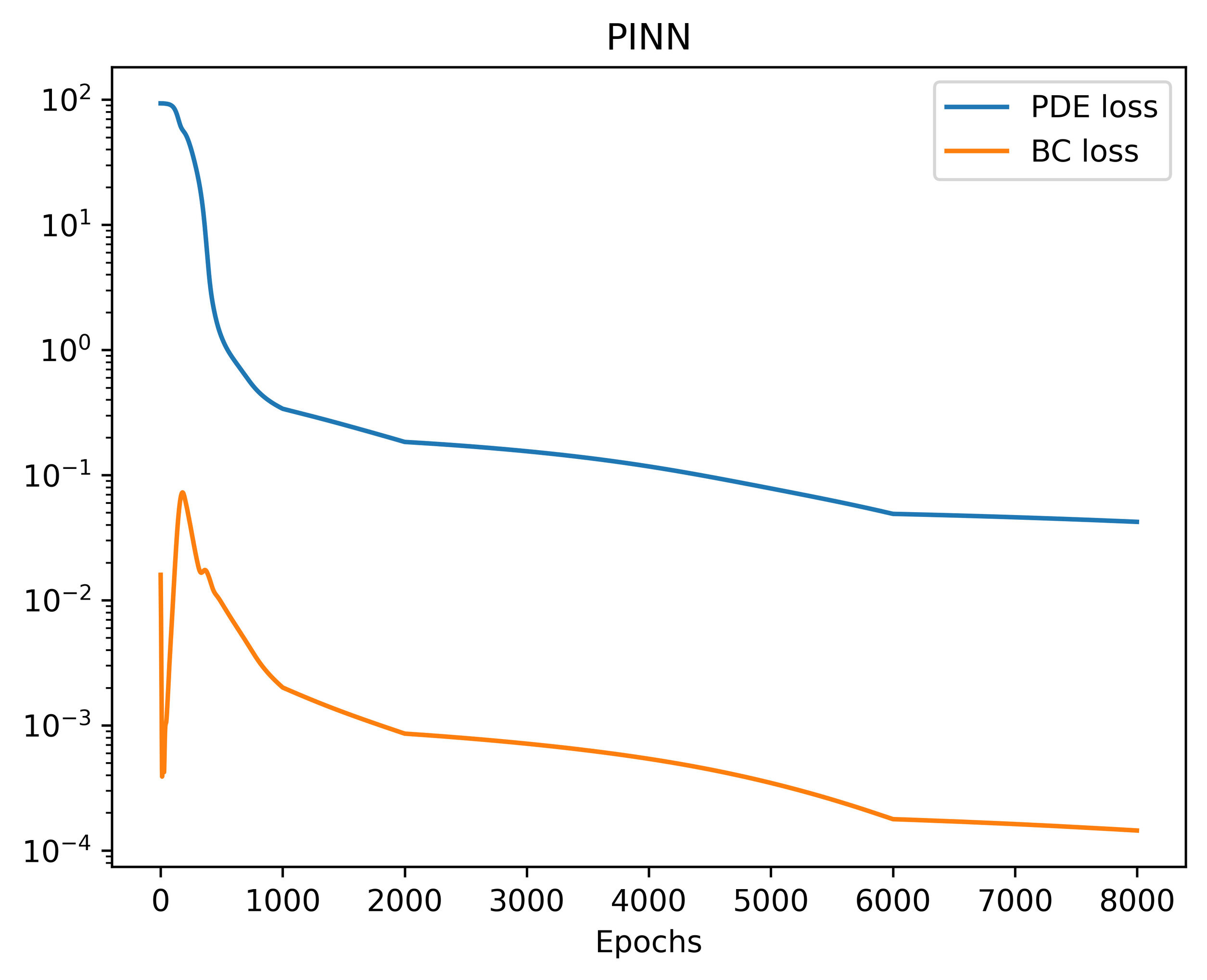}
    \end{minipage}
    \hfill
    \begin{minipage}[b]{0.23\textwidth}
        \centering
        \includegraphics[width=\textwidth]{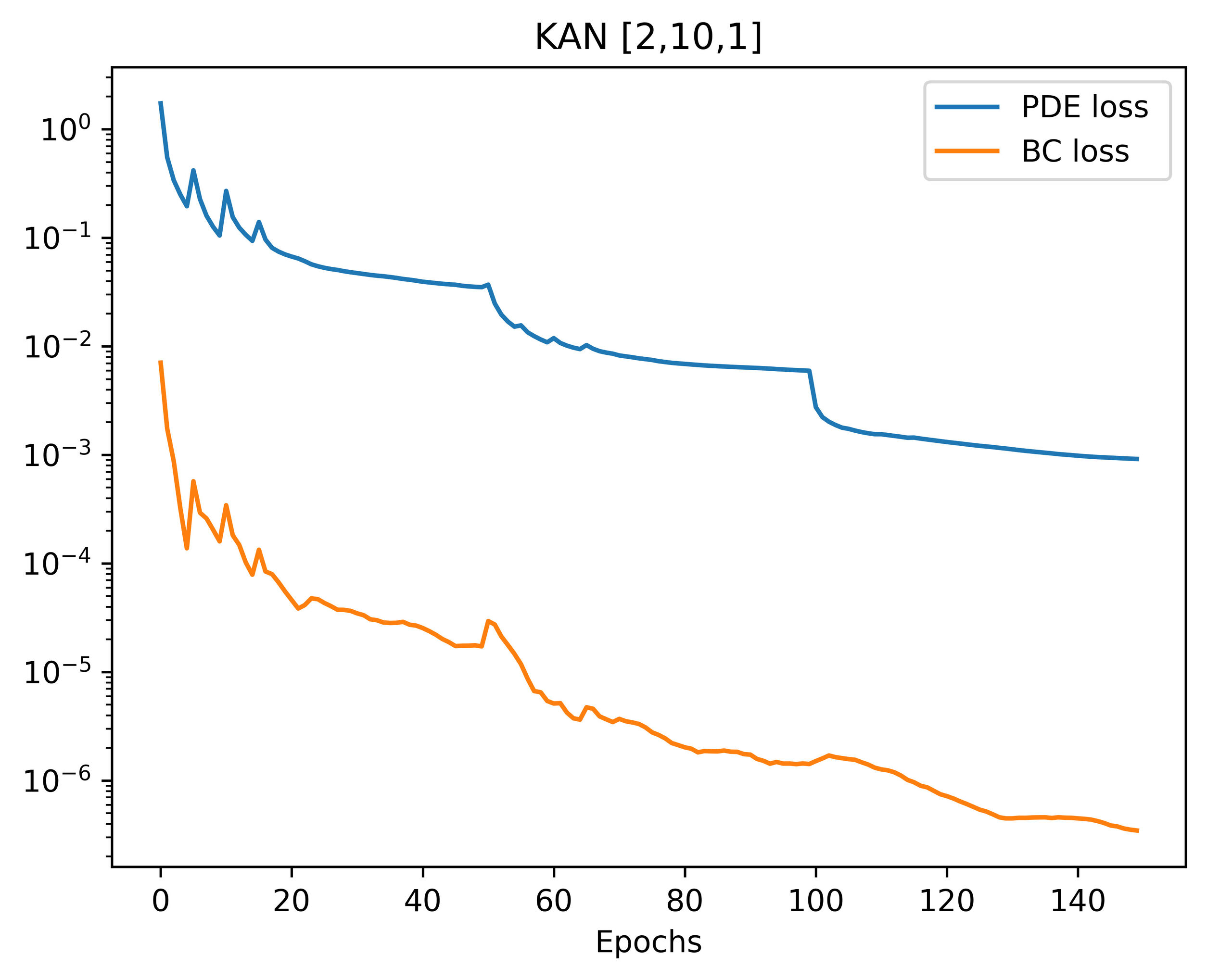}
    \end{minipage}
    \hfill
    \begin{minipage}[b]{0.23\textwidth}
        \centering
        \includegraphics[width=\textwidth]{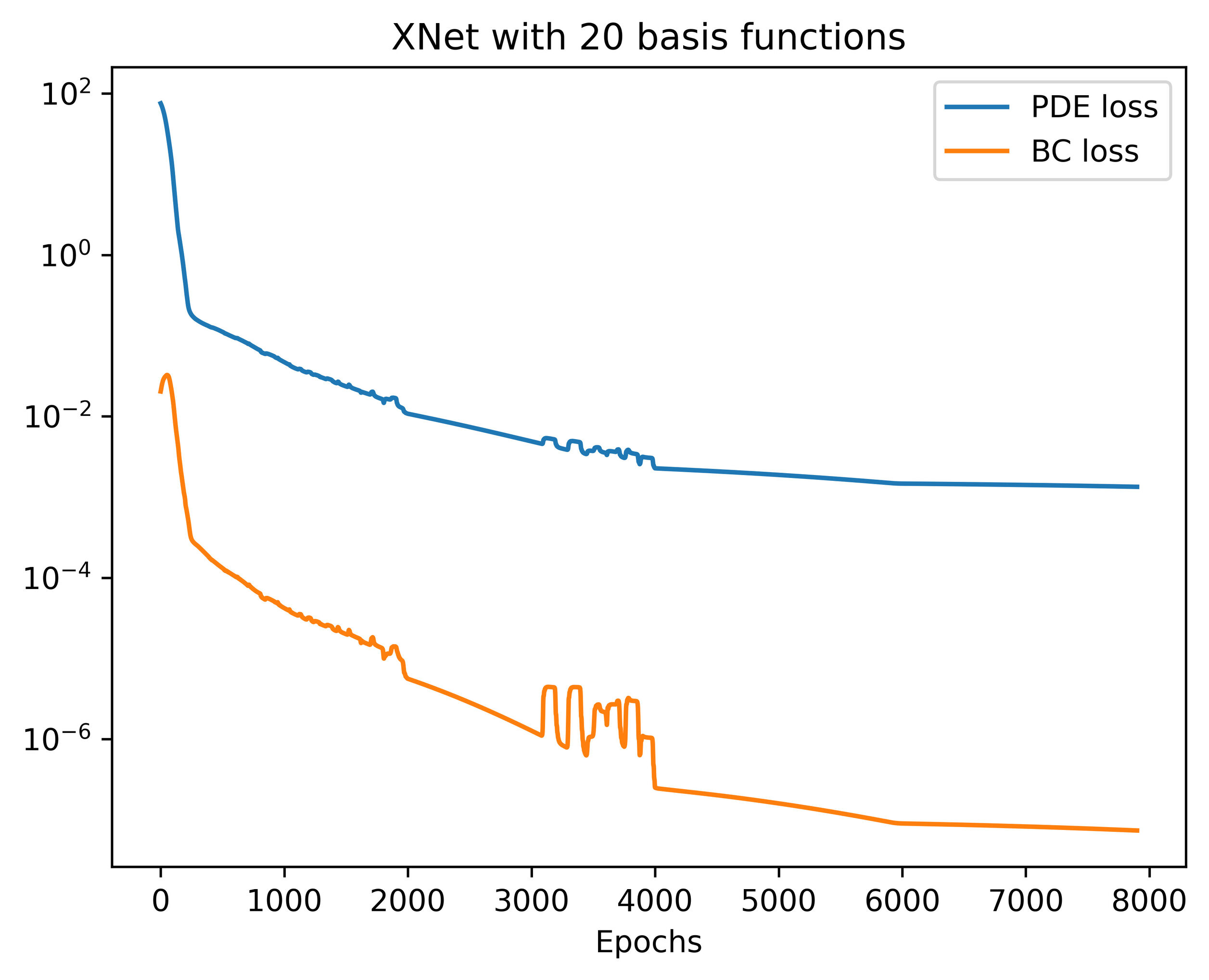}
    \end{minipage}
    \hfill
    \begin{minipage}[b]{0.23\textwidth}
        \centering
        \includegraphics[width=\textwidth]{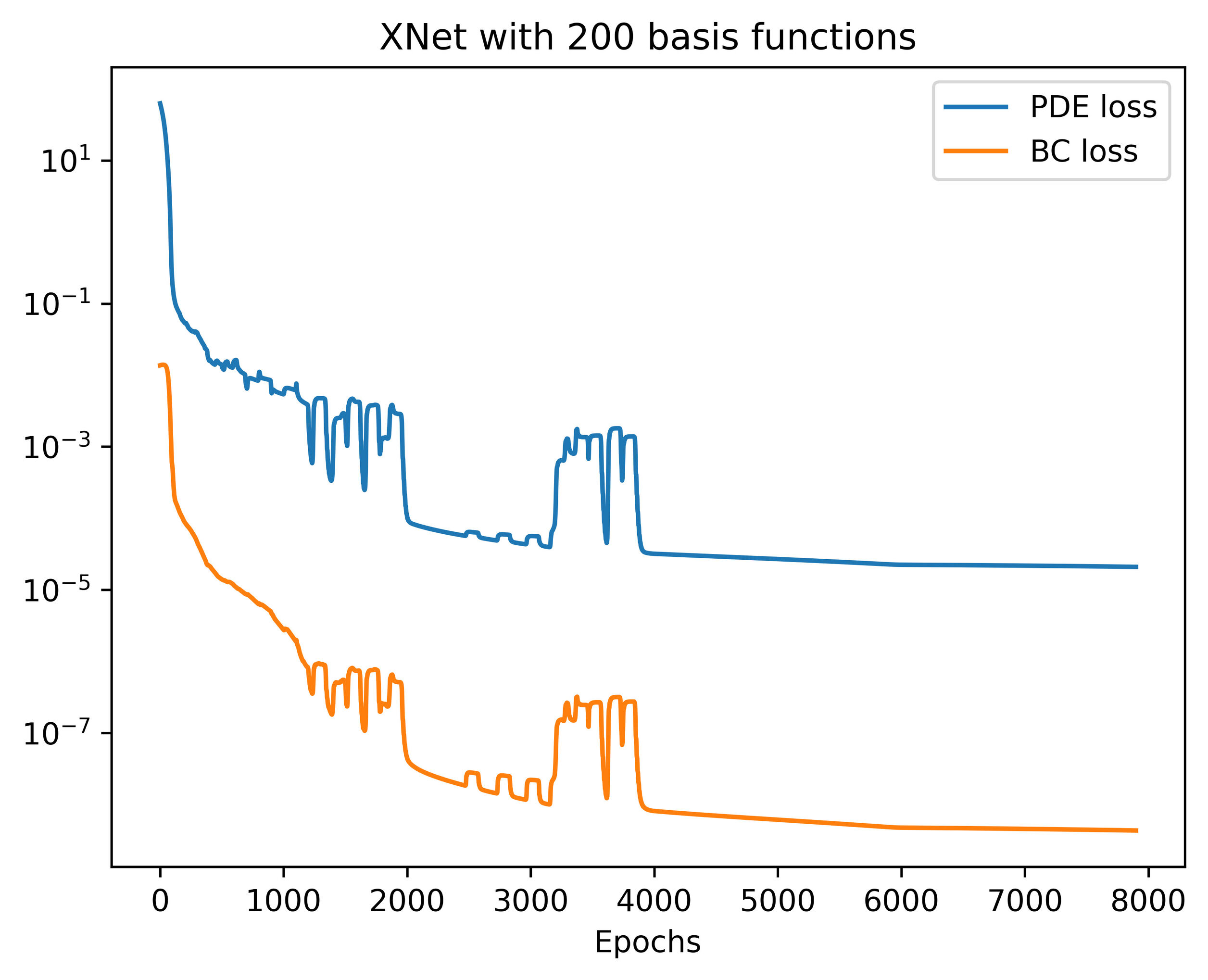}
    \end{minipage}
    \caption{Comparison of KAN, PINN and XNet approximations on PDE loss.}
    \label{fig:comparison}
\end{figure}

We compare the KAN, XNet and PINNs using the same hyperparameters $n_i=2500$, $n_b=200$, and $\alpha=0.01.$ We measured the error in the $L^2$ norm (MSE) and observed that XNet achieved a smaller error, requiring less computational time, as shown in Figure \ref{fig:comparison}. 
A width-200 XNet is 50 times more accurate and 2 times faster than a 2-Layer width-10 KAN;
a width-20 XNet is 3 times more accurate and 5 times faster than a 2-Layer width-10 KAN (see Table \ref{table:poison_compare}).
Therefore we speculate that the XNet might have the potential of serving as a good neural network representation for model reduction of PDEs.  In general, KANs and PINNs are good at representing different function classes of PDE solutions, which needs detailed future study to understand their respective boundaries.

\begin{table}[!ht]
\centering
\small 
\caption{Comparison of XNet and KAN on the Poisson equation.}
\label{table:poison_compare}
\begin{tabular}{ccccc} 
\toprule
\textbf{Metric} & \textbf{MSE} & \textbf{RMSE} & \textbf{MAE} & \textbf{Time (s)} \\
\midrule
\textbf{PINN [2,20,20,1]} & 1.7998e-05 & 4.2424e-03 & 2.3300e-03 & 48.9 \\ 
\textbf{XNet (20)} & 1.8651e-08 & 1.3657e-04 & 1.0511e-04 & 57.2 \\ 
\textbf{KAN [2,10,1]} & 5.7430e-08 & 2.3965e-04 & 1.8450e-04 & 286.3 \\ 
\textbf{XNet (200)} & 1.0937e-09 & 3.3071e-05 & 2.1711e-05 & 154.8 \\ 
\bottomrule
\end{tabular}
\end{table}

\begin{figure}[h!]
    \centering
    \begin{minipage}[b]{0.5\textwidth}
        \centering
        \includegraphics[width=\linewidth]{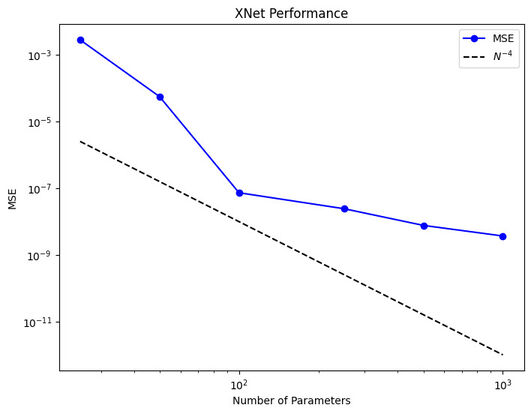}
    \end{minipage}\hfill
\caption{XNet Performance with Number of Parameters}
\label{fig:XNet_pde_params}
\end{figure}

\subsection{XNet enhance the LSTM} \label{sec:LSTM}
Time prediction tasks can generally be categorized into two types: those driven by mathematical and physical models, and those that are data-driven. In the former, time prediction can often be formulated as a function approximation problem, while the latter involves noisy data, cannot be easily described by deterministic partial differential equations (PDEs). In this subsection, we introduce the \textbf{XLSTM} algorithm, which enhances the standard LSTM framework by replacing its feed-forward neural network (FNN) component with \textbf{XNet}. Across various examples, XLSTM consistently demonstrates superior predictive performance compared to the traditional LSTM. In the following experiments, we will demonstrate that XLSTM also significantly outperforms the KAN model in noisy time series examples. The KAN implementation for time series prediction is sourced from this repository: https://github.com/Nixtla/neuralforecast

\textbf{Example 1: Predicting a Synthetic Time Series}

The time series is generated by the following equations:
$$x_5^i=0.1*x_0^ix_1^i+0.1*sin(x_2^ix_3^i)+ sin(x_4^i) + \mu^i, i=1,2,...,n$$
and
$$x_0^i=x_1^{i-1},x_1^i=x_2^{i-1},x_2^i=x_3^{i-1},x_4^i=x_5^{i-1},  $$
where the initial conditions \( x_0^0, x_1^0, x_2^0, x_3^0, x_4^0 \thicksim \text{rand}(0, 0.2) \) are randomly sampled in the range \([0, 0.2]\), and the noise term \( \mu^i \) is sampled from a normal distribution, \( \mu^i \sim N(0, \text{noise}) \). This generates a time series \( \{ f^i = x_5^i \}_{i=1,\dots,n} \), with \( n = 200 \).
In this example, the time series is governed by relatively simple functions. The task of predicting the sixth data point using the first five data points becomes a high-dimensional function approximation problem.

\begin{figure}[h!]
    \centering
    \begin{minipage}[b]{0.45\textwidth}
        \centering
        \includegraphics[width=\textwidth]{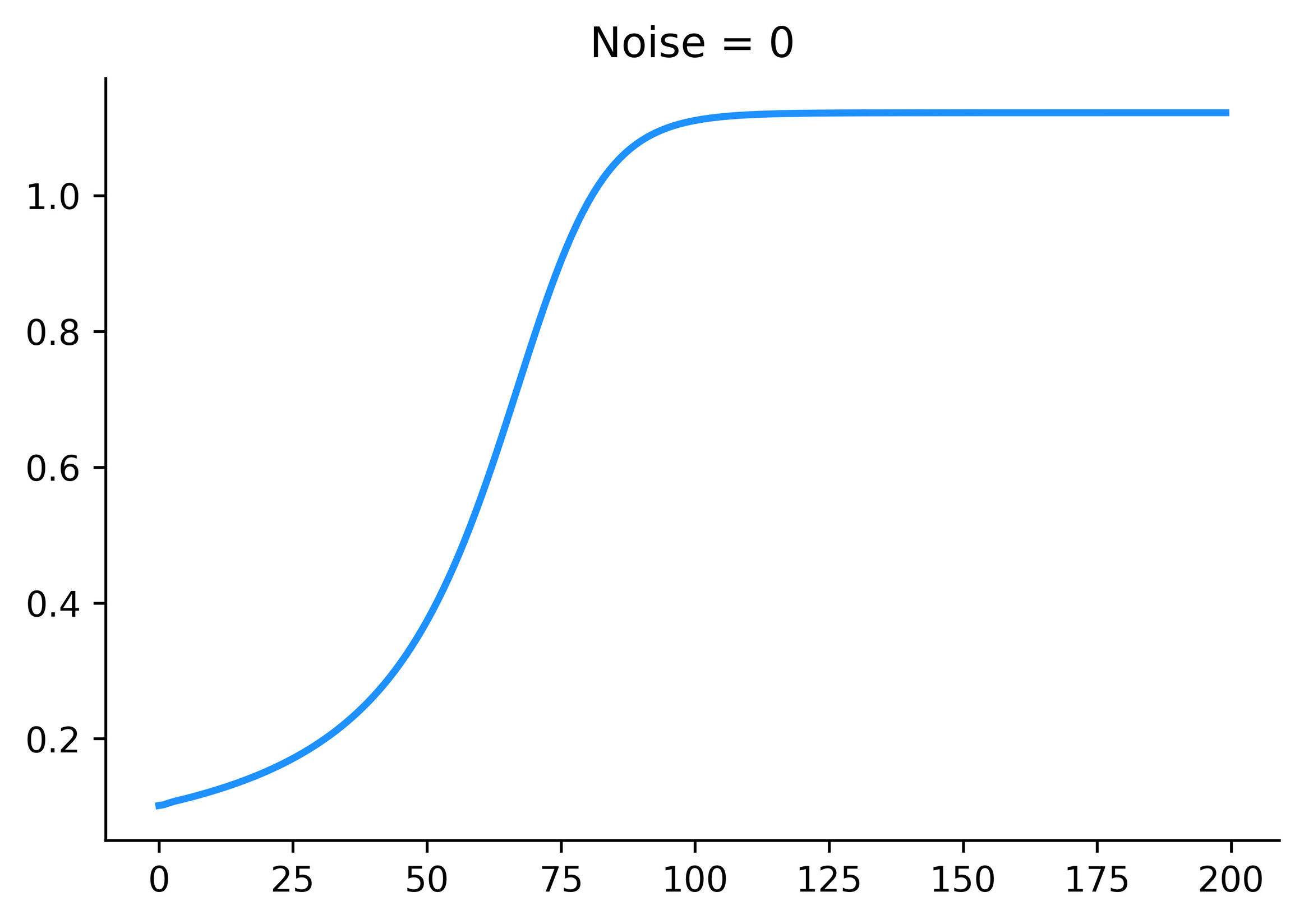}
    \end{minipage}
    \hfill
    \begin{minipage}[b]{0.45\textwidth}
        \centering
        \includegraphics[width=\textwidth]{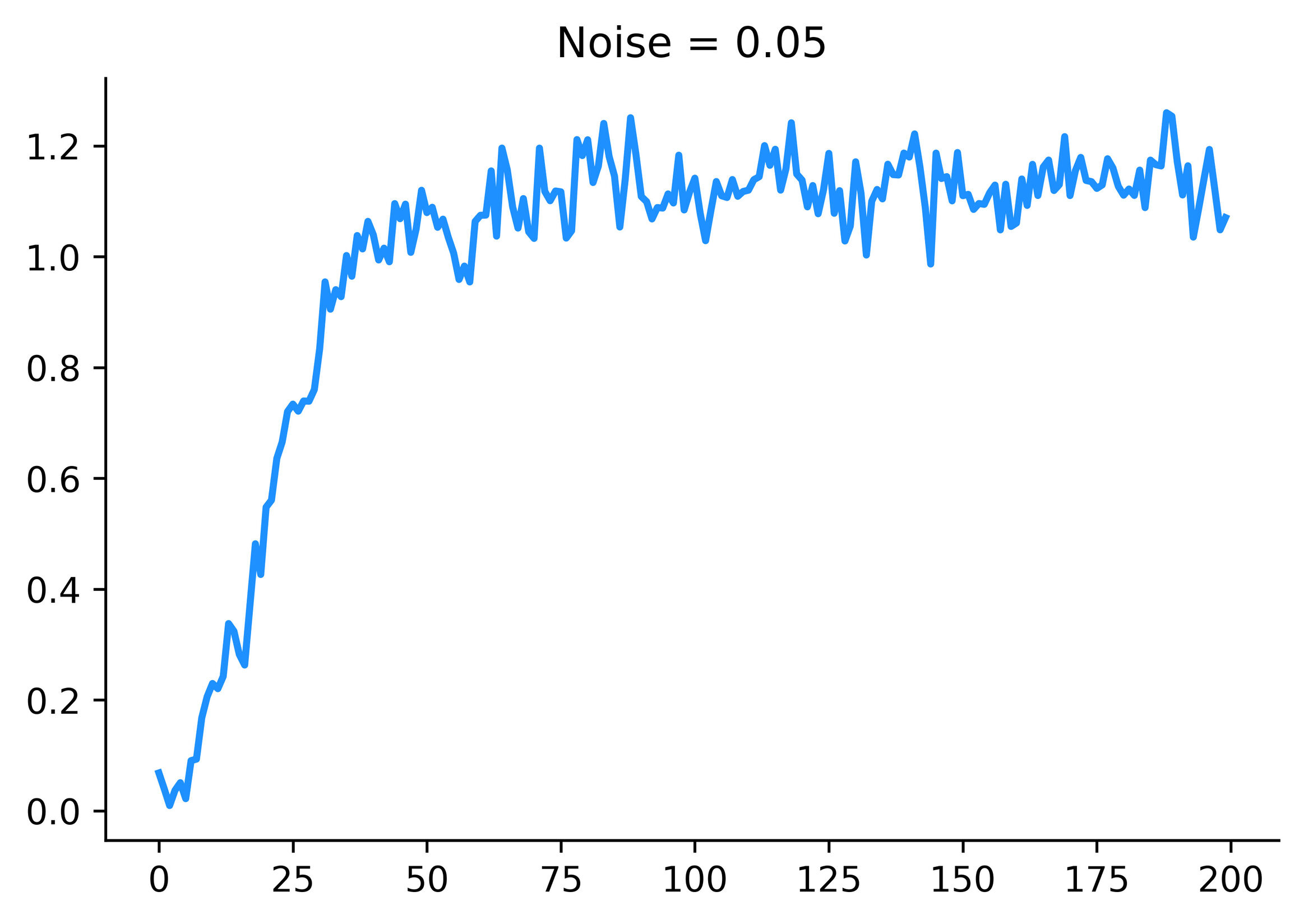}
    \end{minipage}
    \caption{noise=0,0.05}
    \label{fig:ts1_noise}
\end{figure}

\begin{figure}[h!]
    \centering
    \begin{minipage}[b]{0.30\textwidth}
        \centering
        \includegraphics[width=\textwidth]{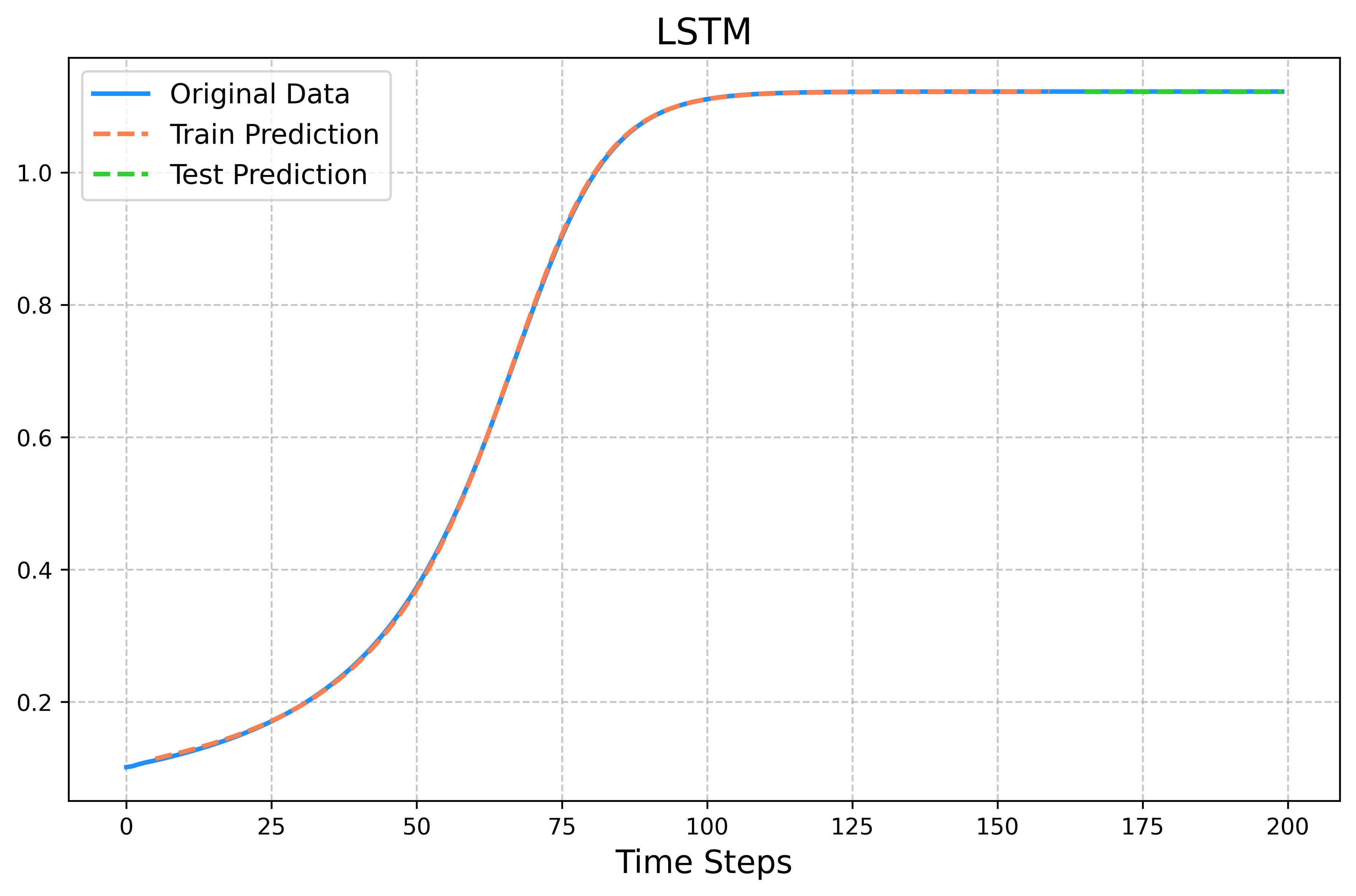}
        \caption*{LSTM (Noise = 0)}
    \end{minipage}
    \hfill
    \begin{minipage}[b]{0.30\textwidth}
        \centering
        \includegraphics[width=\textwidth]{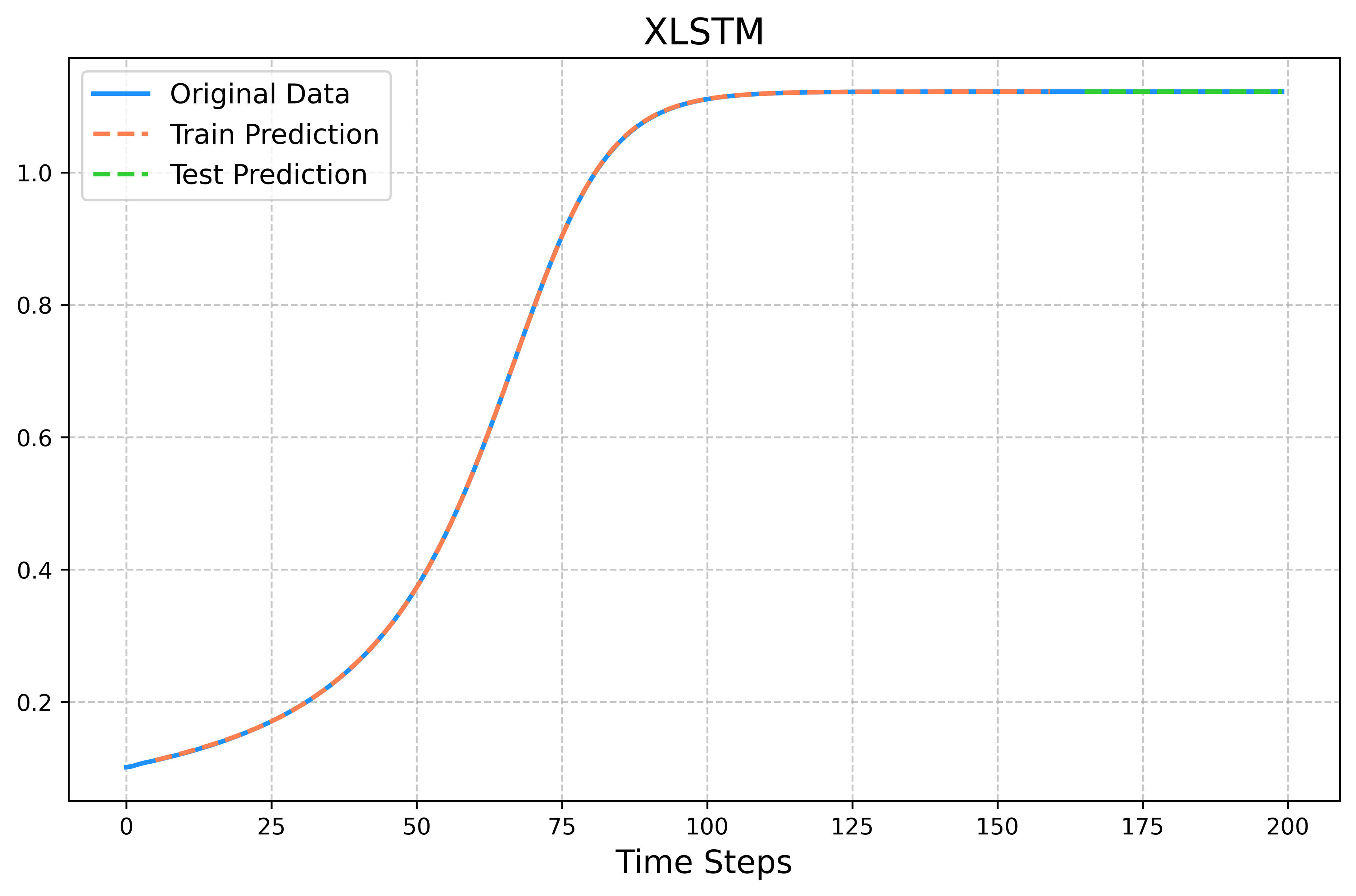}
        \caption*{XLSTM (Noise = 0)}
    \end{minipage}
    \hfill
    \begin{minipage}[b]{0.30\textwidth}
        \centering
        \includegraphics[width=\textwidth]{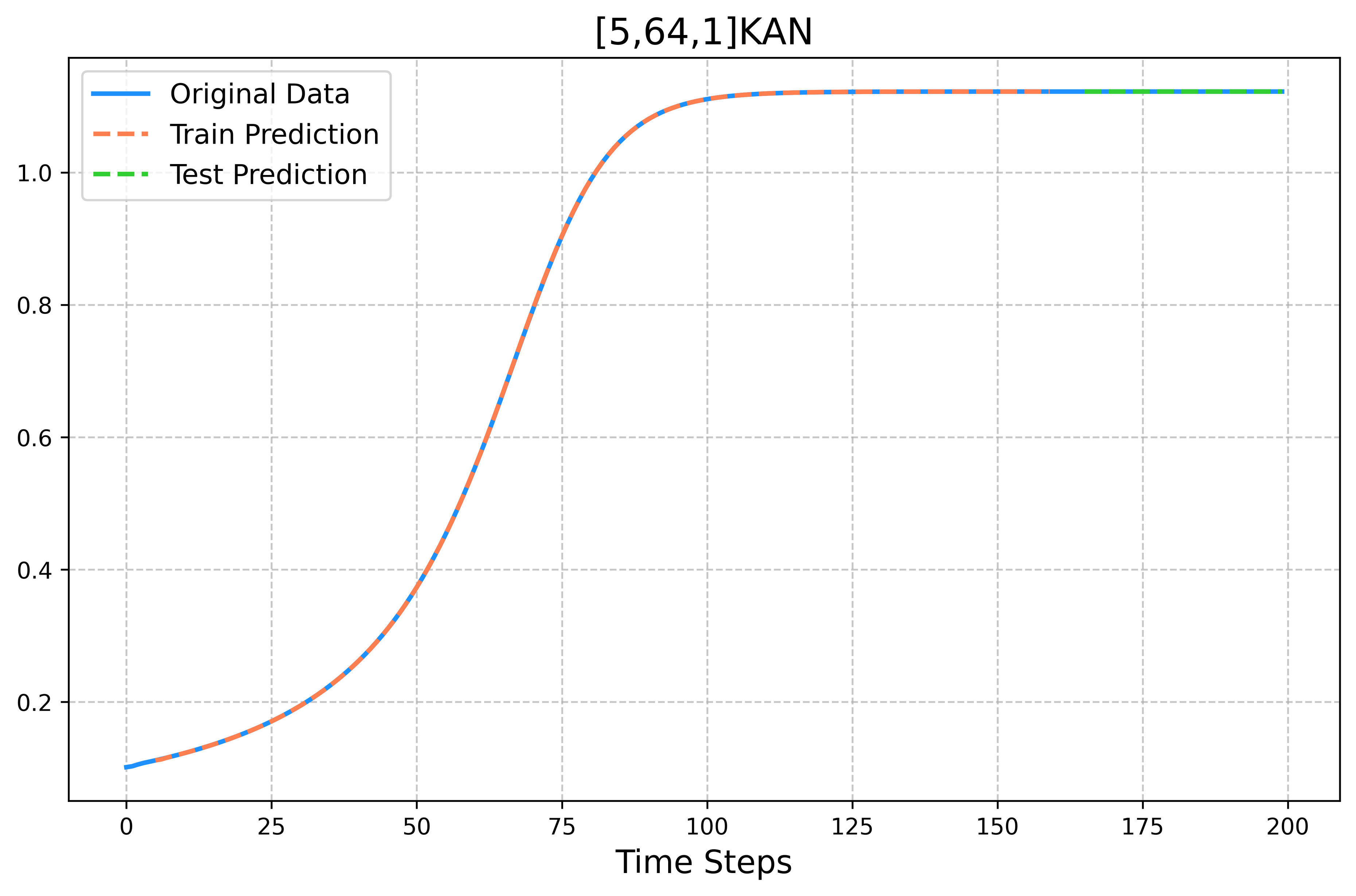}
        \caption*{KAN (Noise = 0)}
    \end{minipage}
    
    \vspace{0.5cm}
    
    \begin{minipage}[b]{0.30\textwidth}
        \centering
        \includegraphics[width=\textwidth]{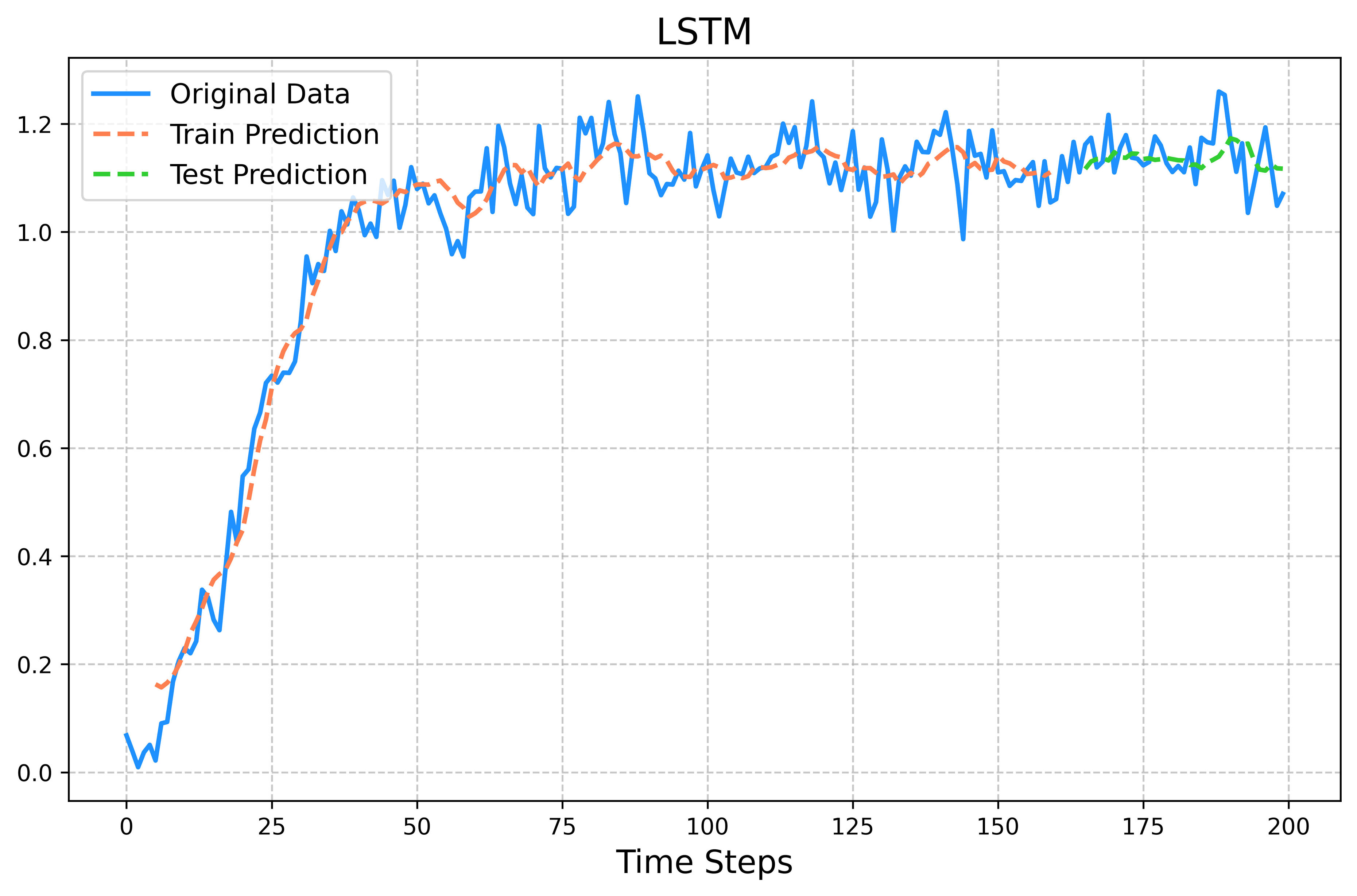}
        \caption*{LSTM (Noise = 0.05)}
    \end{minipage}
    \hfill
    \begin{minipage}[b]{0.30\textwidth}
        \centering
        \includegraphics[width=\textwidth]{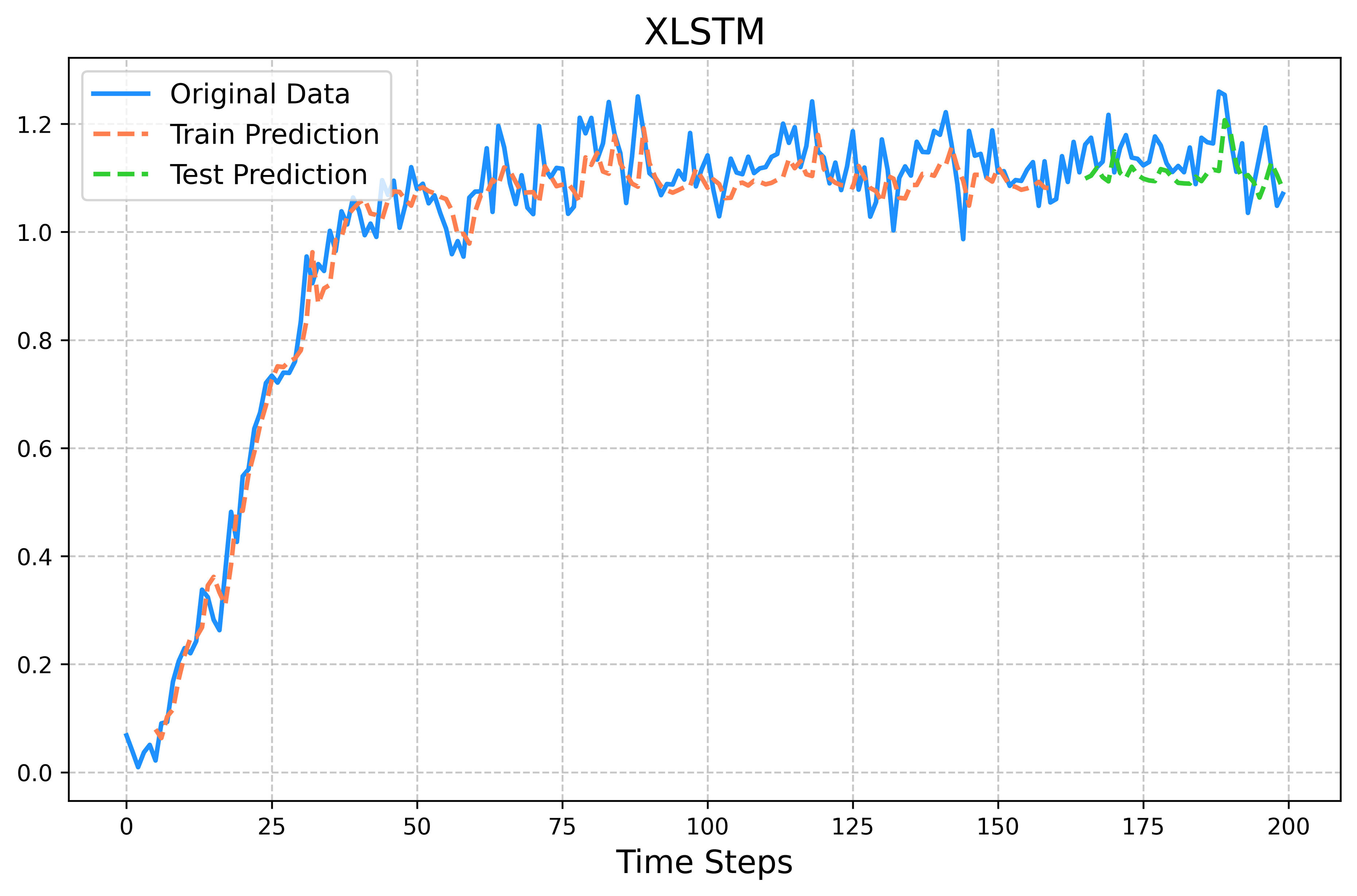}
        \caption*{XLSTM (Noise = 0.05)}
    \end{minipage}
    \hfill
    \begin{minipage}[b]{0.30\textwidth}
        \centering
        \includegraphics[width=\textwidth]{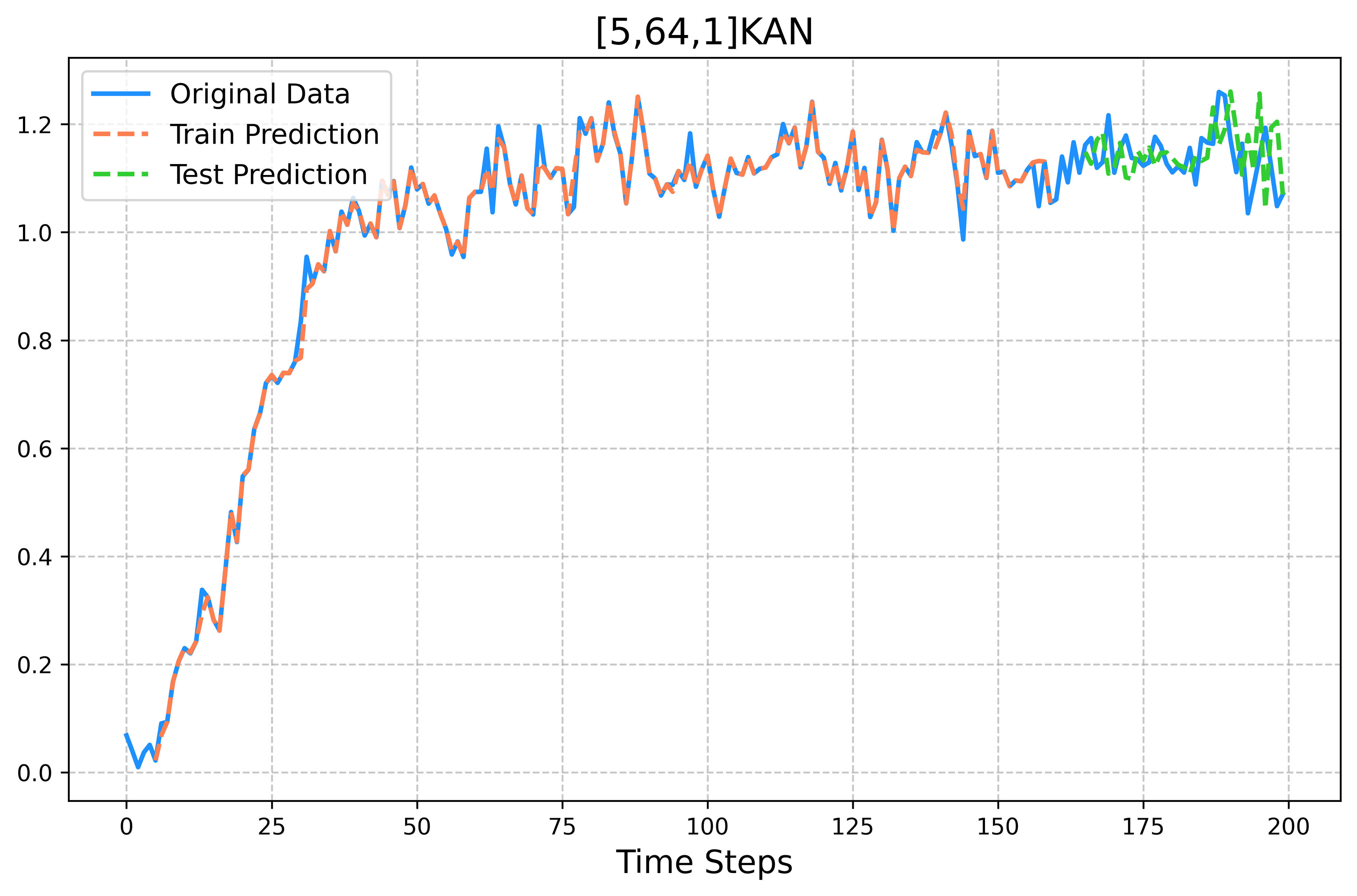}
        \caption*{KAN (Noise = 0.05)}
    \end{minipage}
    
    \caption{Comparison of the performance of LSTM, XLSTM, and KAN under different noise levels. The first row shows the results for noise level 0, while the second row corresponds to noise level 0.05.}
    \label{fig:comparison_LSTM_XLSTM}
\end{figure}

Figures [\ref{fig:ts1_noise}] and [\ref{fig:comparison_LSTM_XLSTM}] show a comparison of the predictive performance of LSTM and XLSTM on two scenarios: one with no noise (noise = 0) and one with moderate noise (noise = 0.05). The results indicate that XLSTM significantly outperforms LSTM in both settings, particularly under non-noisy conditions. When there is no noise, XLSTM achieves an MSE of \( 3.4252 \times 10^{-11} \), which is lower than that of LSTM (\( 1.5925 \times 10^{-7} \)). Similarly, XLSTM's \textbf{RMSE} and \textbf{MAE} are drastically lower than LSTM's, while the computation time remains comparable.
In the presence of moderate noise (noise = 0.05), although XLSTM does not show a significant advantage in metrics such as MSE, it is clear from Figure (\ref{fig:ts1_noise}) that XLSTM captures the underlying patterns of the data better than LSTM.

\begin{table}[!ht]
\centering
\small 
\caption{Comparison of LSTM and XLSTM on the example1 (noise=0).}
\label{table:noise0}
\begin{tabular}{ccccc} 
\toprule
\textbf{Metric} & \textbf{MSE} & \textbf{RMSE} & \textbf{MAE} & \textbf{Time (s)} \\
\midrule
\textbf{LSTM} & 1.5925e-07 &3.9906e-04& 3.9906e-04 & 9.01 \\ 
\textbf{XLSTM} & 3.4252e-11& 5.8525e-06& 5.8457e-06 & 9.42 \\ 
\textbf{[5,64,1]KAN} & 9.8281e-13& 9.9137e-07& 8.0000e-07 & 11.63 \\ 
\bottomrule
\end{tabular}
\end{table}

\begin{table}[!ht]
\centering
\small 
\caption{Comparison of LSTM and XLSTM on the example1 (noise=0.05).}
\label{table:noise1}
\begin{tabular}{ccccc} 
\toprule
\textbf{Metric} & \textbf{MSE} & \textbf{RMSE} & \textbf{MAE} & \textbf{Time (s)} \\
\midrule
\textbf{LSTM} & 2.5919e-03 &5.0911e-02 &3.8814e-02 & 9.07 \\ 
\textbf{XLSTM} & 2.2080e-03&4.6990e-02&3.7182e-02& 9.56 \\ 
\textbf{[5,64,1]KAN} & 4.6537e-03&6.8218e-02&5.3703e-02& 11.59 \\ 
\bottomrule
\end{tabular}
\end{table}

In this example of a mathematical model-driven time series, XLSTM clearly outperforms LSTM, particularly in noisy and noise-free environments. Given these results, we hypothesize that XLSTM will also exhibit superior performance in highly noisy, real-world datasets, such as financial time series, where traditional LSTM models may struggle. The [5,64,1] KAN model, however, shows signs of overfitting, with excellent performance on the training set but noticeable degradation on the test set.

\textbf{Example 2: Predicting a Financial Time Series}

This is a toy model case with extremely noisy data. Stock price
patterns are notoriously unpredictable, and we do not claim that our
simplistic model outperforms others. We included this case merely to
demonstrate the models potential. In this experiment, we focus on
Apple's stock price from the U.S. market, with the test period
spanning from July 1, 2016 to July 1, 2017. The entire set of 252 data
points is divided into two parts: 201 for training and 51 for
testing. We consider using LSTM and XLSTM for time series prediction,
where the model uses the first 10 data points and predicts the
11th. After 500 iterations, training was deemed complete.

\begin{figure}[h!]
    \centering
    \begin{minipage}[b]{0.3\textwidth}
        \centering
        \includegraphics[width=\textwidth]{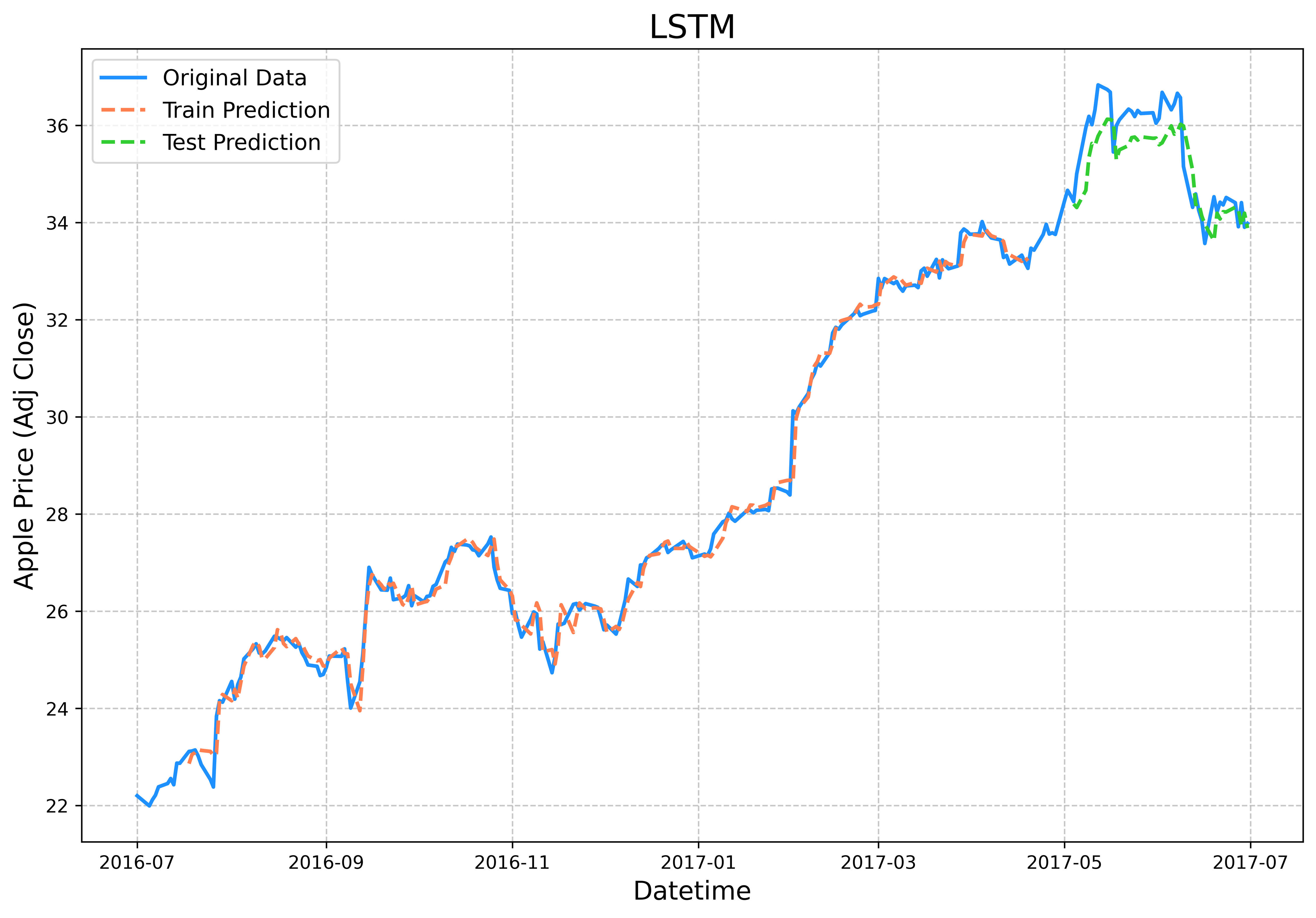}
    \end{minipage}
    \hfill
    \begin{minipage}[b]{0.3\textwidth}
        \centering
        \includegraphics[width=\textwidth]{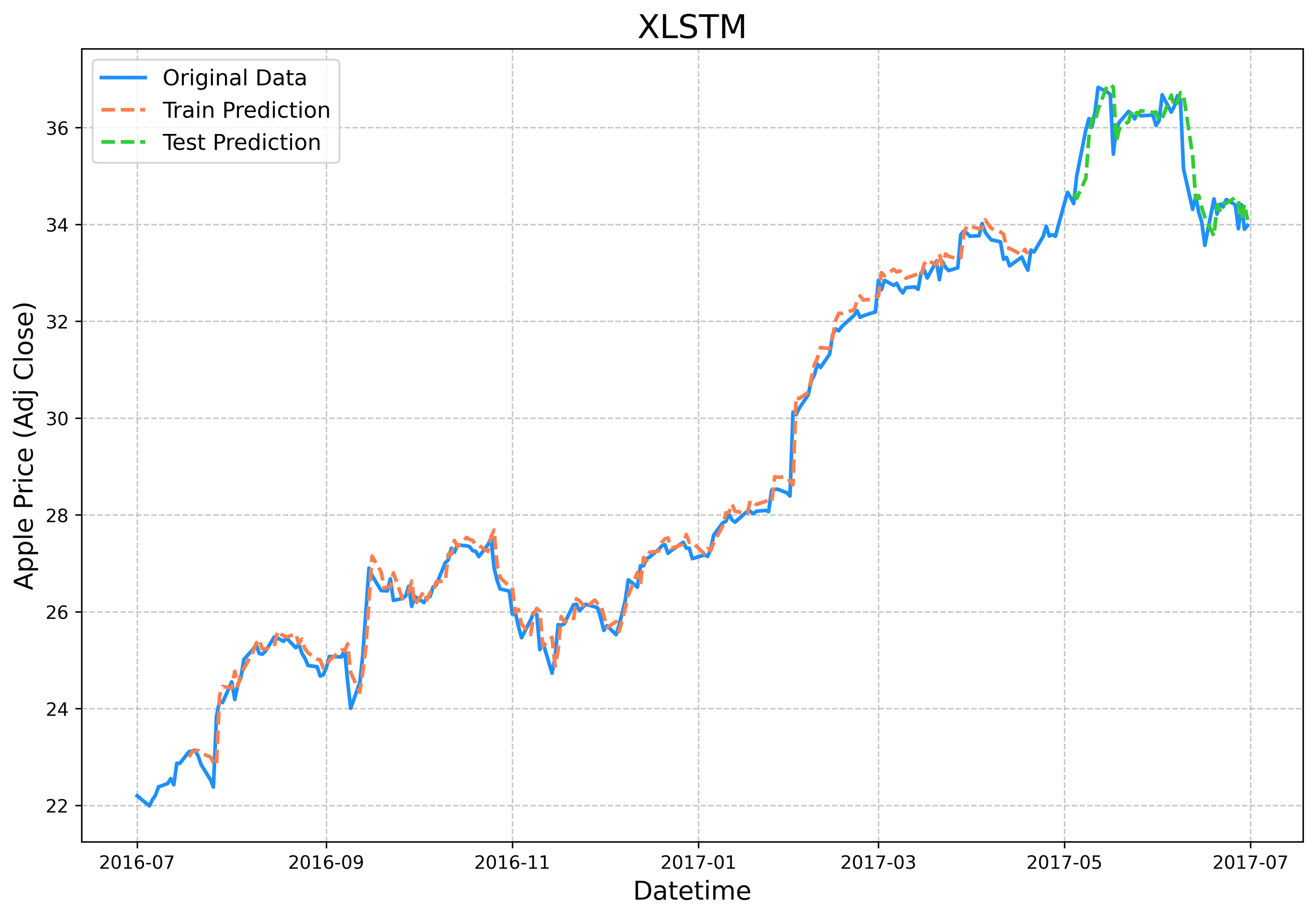}
    \end{minipage}
    \hfill
    \begin{minipage}[b]{0.3\textwidth}
        \centering
        \includegraphics[width=\textwidth]{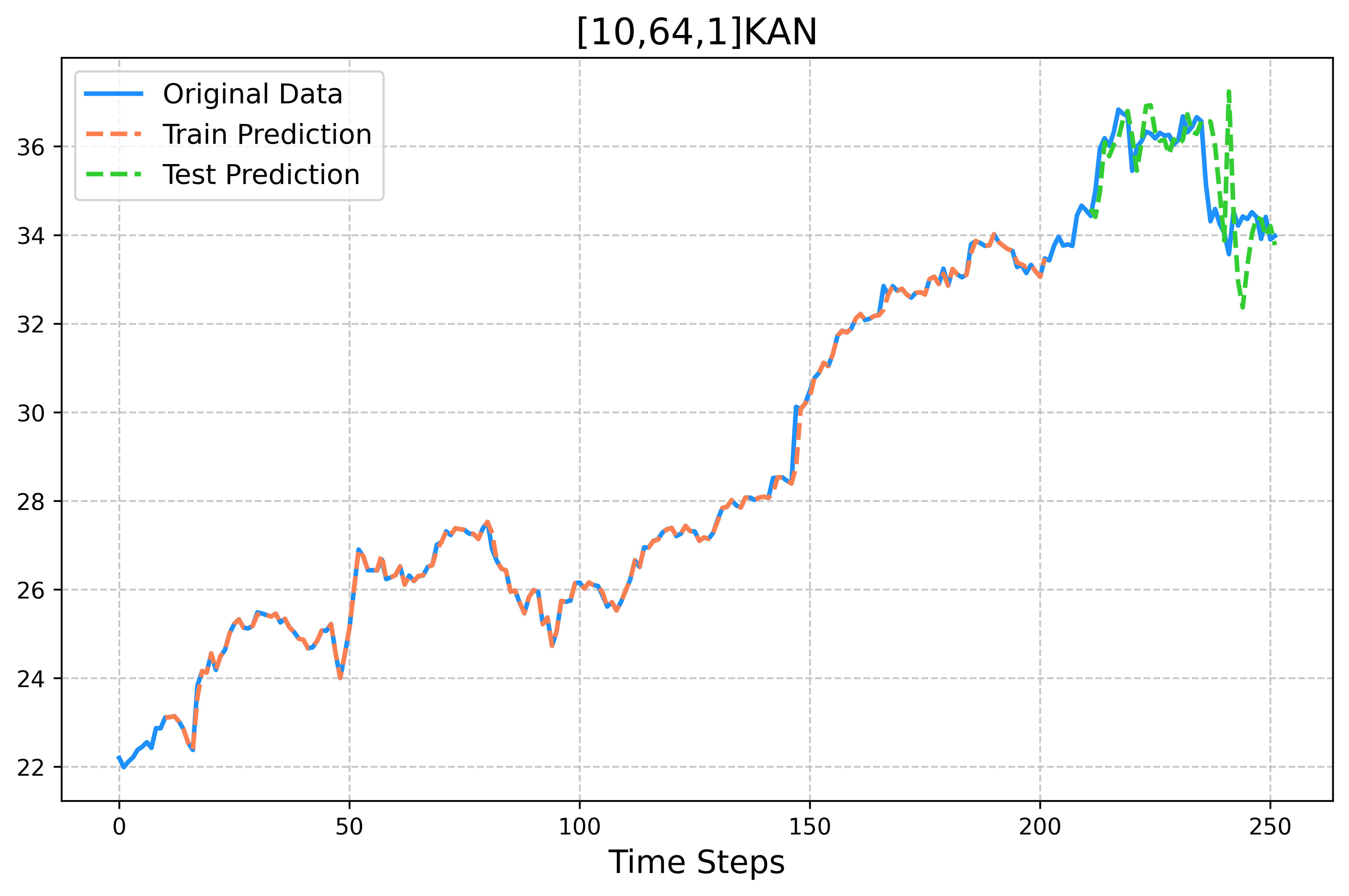}
    \end{minipage}
     \caption{Comparison of the performance of LSTM, XLSTM, and KAN on Apple's stock price}
     \label{fig:stock_Model}
\end{figure}

As shown in Figure \ref{fig:stock_Model}, XLSTM aligns more closely with the original data, outperforming LSTM by a significant margin. In this example, the KAN model continues to exhibit overfitting, making it unsuitable for direct application to time series prediction with significant noise.

\begin{table}[!ht]
\centering
\small 
\caption{Comparison of LSTM, XLSTM and KAN on the Financial Time Series.}
\label{table:stock_compare}
\begin{tabular}{ccccc} 
\toprule
\textbf{Metric} & \textbf{MSE} & \textbf{RMSE} & \textbf{MAE} & \textbf{Time (s)} \\
\midrule
\textbf{LSTM} & 3.3768E-01 &5.8110E-01 & 4.8787E-01 & 8.9574 \\ 
\textbf{XLSTM} & 2.3878E-01&4.8865E-01& 3.3764E-01& 10.1159 \\ 
\textbf{[10,64,1]KAN} & 8.5918e-01& 9.2692e-01& 5.9108e-01 & 11.7505  \\ 
\bottomrule
\end{tabular}
\end{table}

\section{Summary and Outlook} \label{sec:conclusion}

1. \textbf{XNet vs. KAN for Function Approximation}
   Recently, KAN has gained popularity as a function approximator. However, our experiments demonstrate that XNet outperforms Kan, particularly when approximating discontinuous or high-dimensional functions.

2. \textbf{XNet in the PINN Framework}
   Within the Physics-Informed Neural Networks (PINN) framework, we verified that using KAN significantly improves the accuracy of traditional PINNs. Moreover, implementing XNet further enhances both accuracy and computational efficiency. We hypothesize this is due to XNet's superior approximation capabilities.

3. \textbf{Enhancing LSTM with XNet }
   Given XNet's ability to capture complex data features, we found that XNet can enhance LSTM performance by replacing the embedded feed-forward neural network (FNN) within the LSTM structure.

4. \textbf{Potential Applications of XNet }
   We believe that XNet can improve the performance of models in other machine learning domains, including image recognition, image generation, computer vision, and more.

\appendix
\section{Appendix}

\subsection{ADDITIONAL EXPERIMENT DETAILS} 
The numerical experiments presented below were performed in Python using the Tensorflow-CPU processor on a Dell computer equipped with a 3.00 Gigahertz (GHz) Intel Core i9-13900KF.
When detailing grids ans k for KAN models, we always use values provided by respective authors (Kan).

\subsection{A.1 FUNCTION APPROXIMATION} \label{A.1}
For 1d heaciside function, we set different configurations. The results are shown as follows
\begin{table}[h!]
\centering
\caption{B-Spline Performance metrics comparison for different G and K values. reference}
\begin{tabular}{|c|c|c|c|}
\hline
\multicolumn{4}{c|}{B-Spline} \\
\hline
\textbf{k, G} & \textbf{MSE} & \textbf{RMSE} & \textbf{MAE} \\
\hline
k=50, G=200 & 5.8477e-01 & 7.6470e-01 & 6.1076e-01 \\
k=3, G=10 & 9.2871e-03 & 9.6369e-02 & 4.7923e-02 \\
k=3, G=50 & 2.3252e-03 & 4.8221e-02 & 1.2255e-02 \\
k=10, G=50 & 1.9881e-03 & 4.4588e-02 & 1.0879e-02 \\
k=3, G=200 & 1.1252e-03 & 3.3544e-02 & 4.4737e-03 \\
k=10, G=200 & 1.1029e-03 & 3.3210e-02 & 5.1904e-03 \\
\hline
\end{tabular}

\end{table}

\begin{table}[htbp]
	\centering
	\caption{KAN  reference}
	\label{table:kan_1d}
	\resizebox{\textwidth}{!}{%
		\begin{tabular}{ccccccc}
			\hline \hline
			& \multicolumn{3}{c}{[1,1]KAN} & \multicolumn{3}{c}{[1,3,1]KAN} \\
			\midrule
			k,G  & MSE & RMSE  & MAE & MSE & RMSE  & MAE \\
			\midrule
			k=3, G=3 & 2.20E-02 & 1.48E-01 & 9.89E-02  & 3.50E-04 & 1.87E-02 & 5.56E-03 \\
			k=3, G=10 & 1.22E-02 & 1.10E-01 & 5.91E-02 & 1.84E-04 & 1.36E-02 & 2.54E-03  \\
			k=3, G=50 & 2.44E-03 & 4.94E-02 & 1.22E-02 & 4.28E-05 & 6.55E-03 & 2.71E-03 \\
			k=3, G=200 & 5.98E-04 & 2.45E-02 & 3.03E-03 & 3.79E-04 & 1.95E-02 & 1.24E-02 \\
			\hline \hline
		\end{tabular}%
	}
	\label{dt_diffrate_results}
\end{table}

For 2d functions, loss function
\begin{figure}[h!]
    \centering
    \begin{minipage}[b]{0.48\textwidth}
        \centering
        \includegraphics[width=0.48\textwidth]{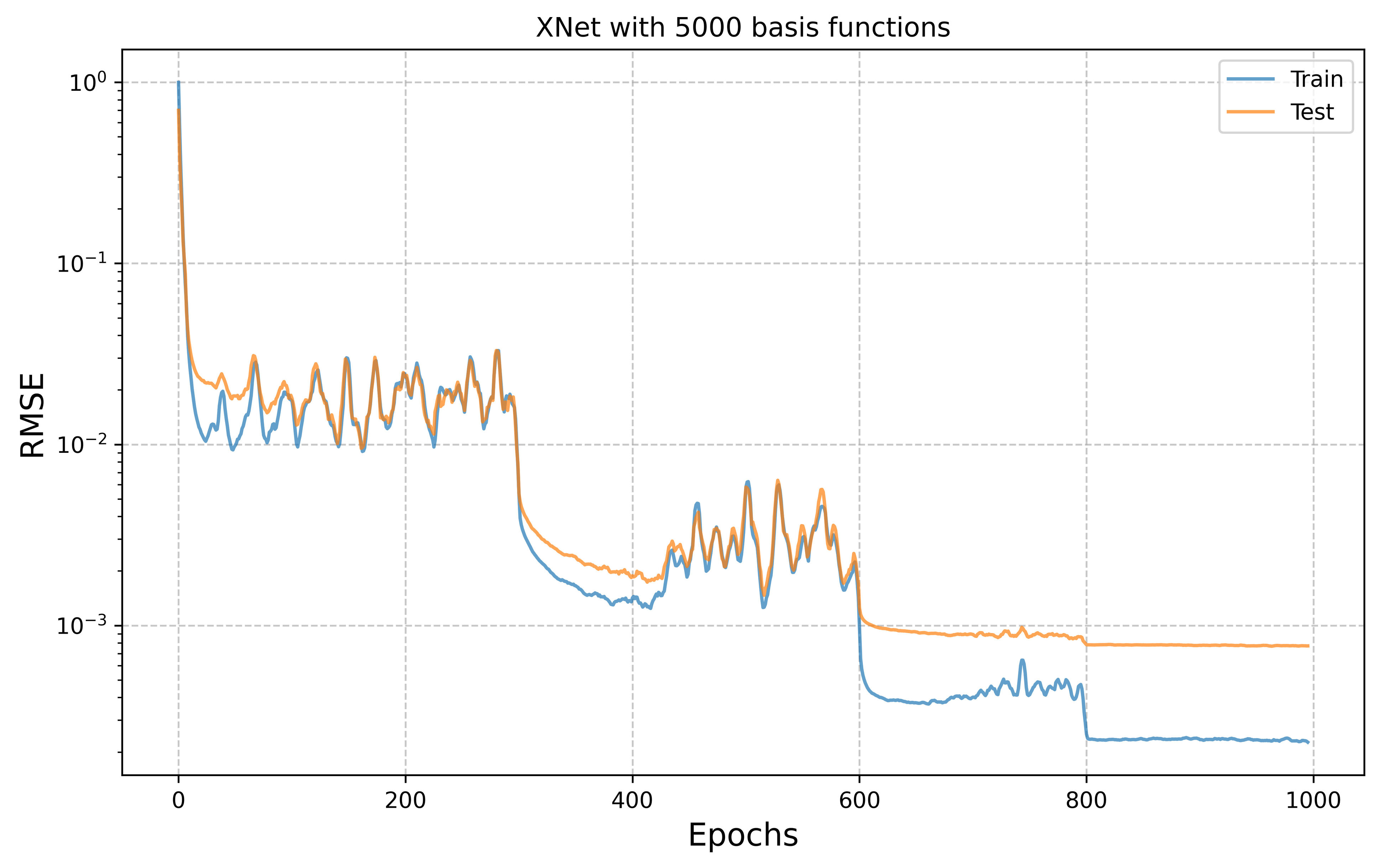}
        \hfill
        \includegraphics[width=0.48\textwidth]{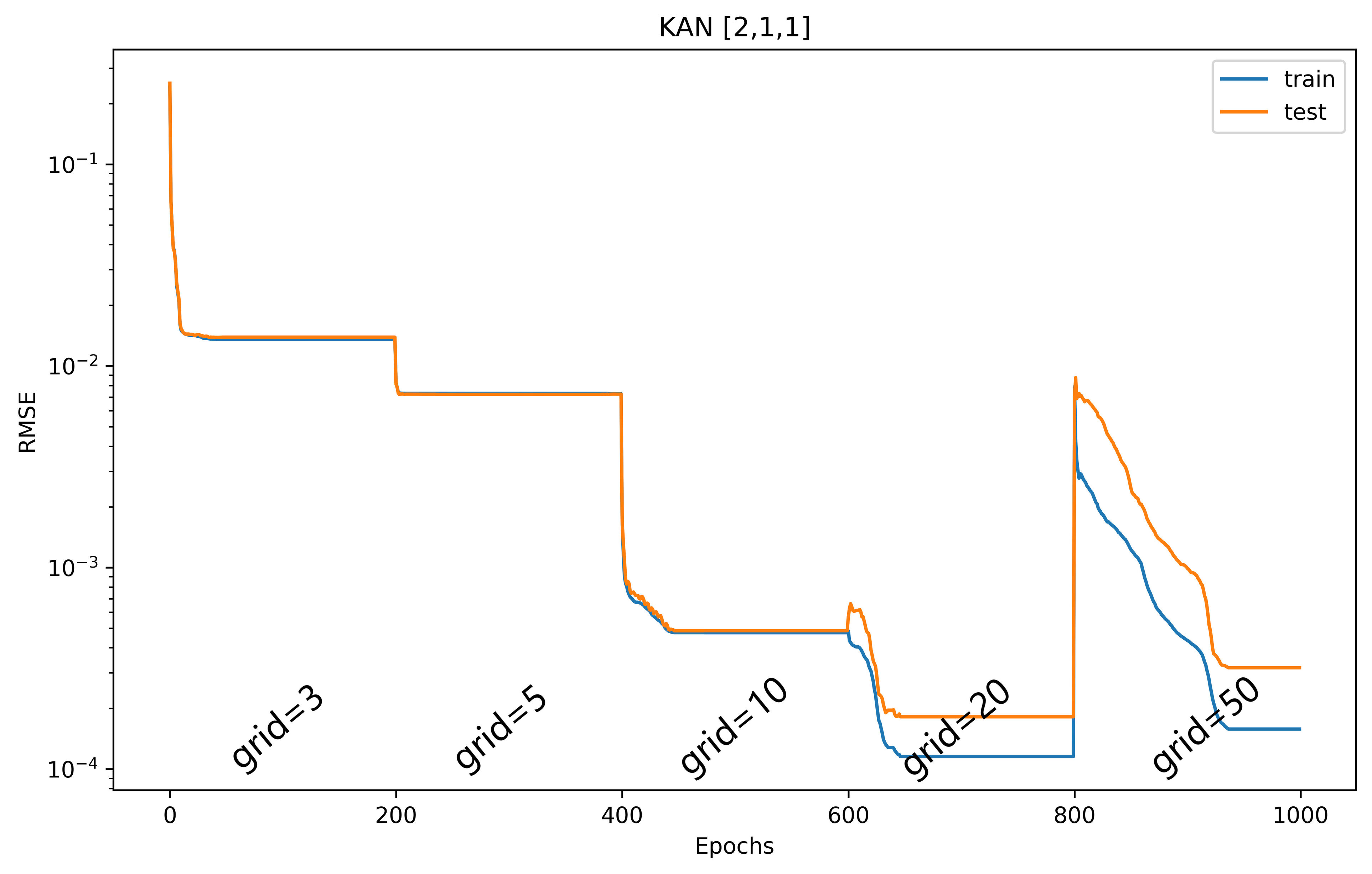}
        \captionof{figure}{Loss on \(\exp(\sin(\pi x) + y^2)\)}
    \end{minipage}
    \hfill
    \begin{minipage}[b]{0.48\textwidth}
        \centering
        \includegraphics[width=0.48\textwidth]{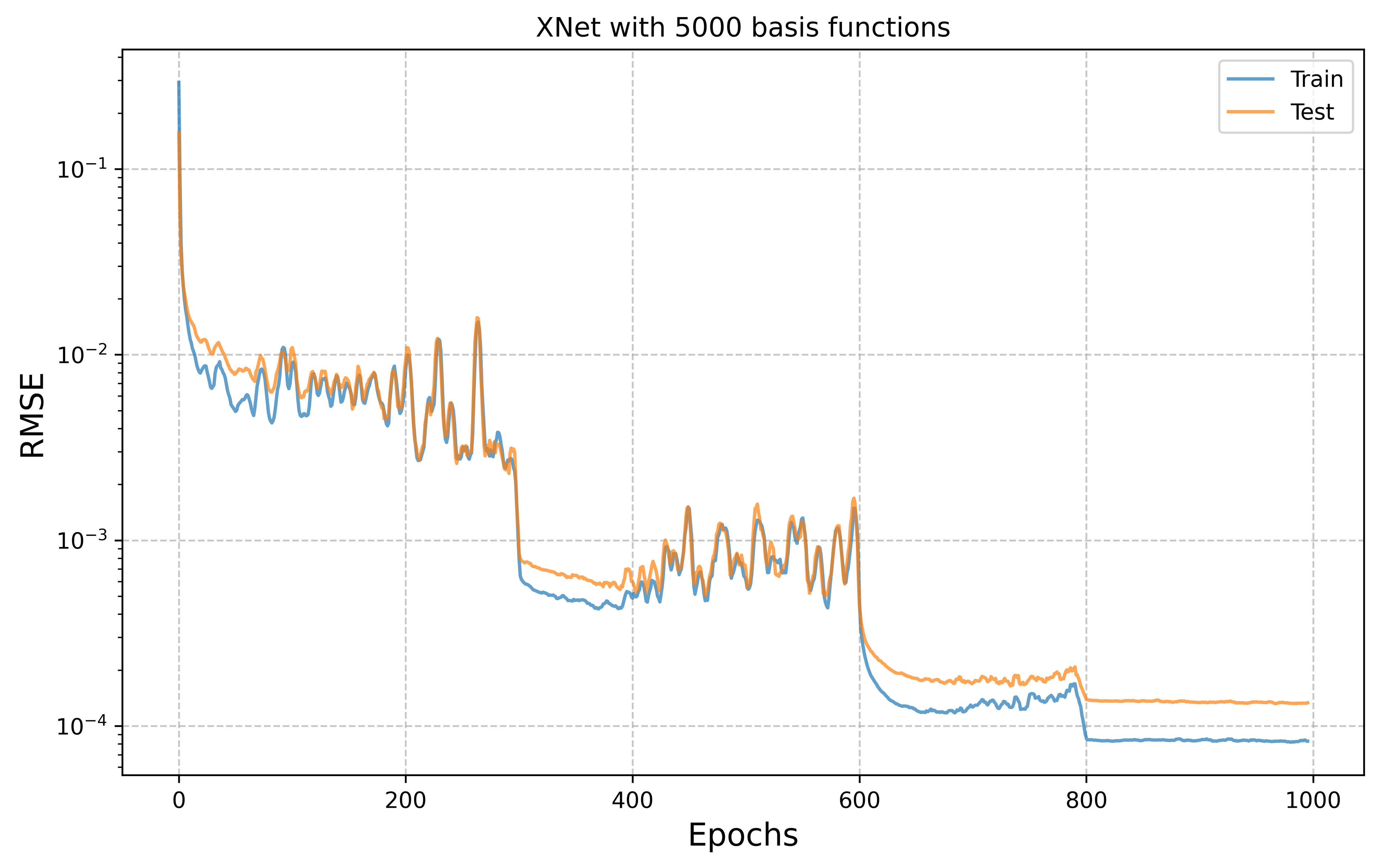}
        \hfill
        \includegraphics[width=0.48\textwidth]{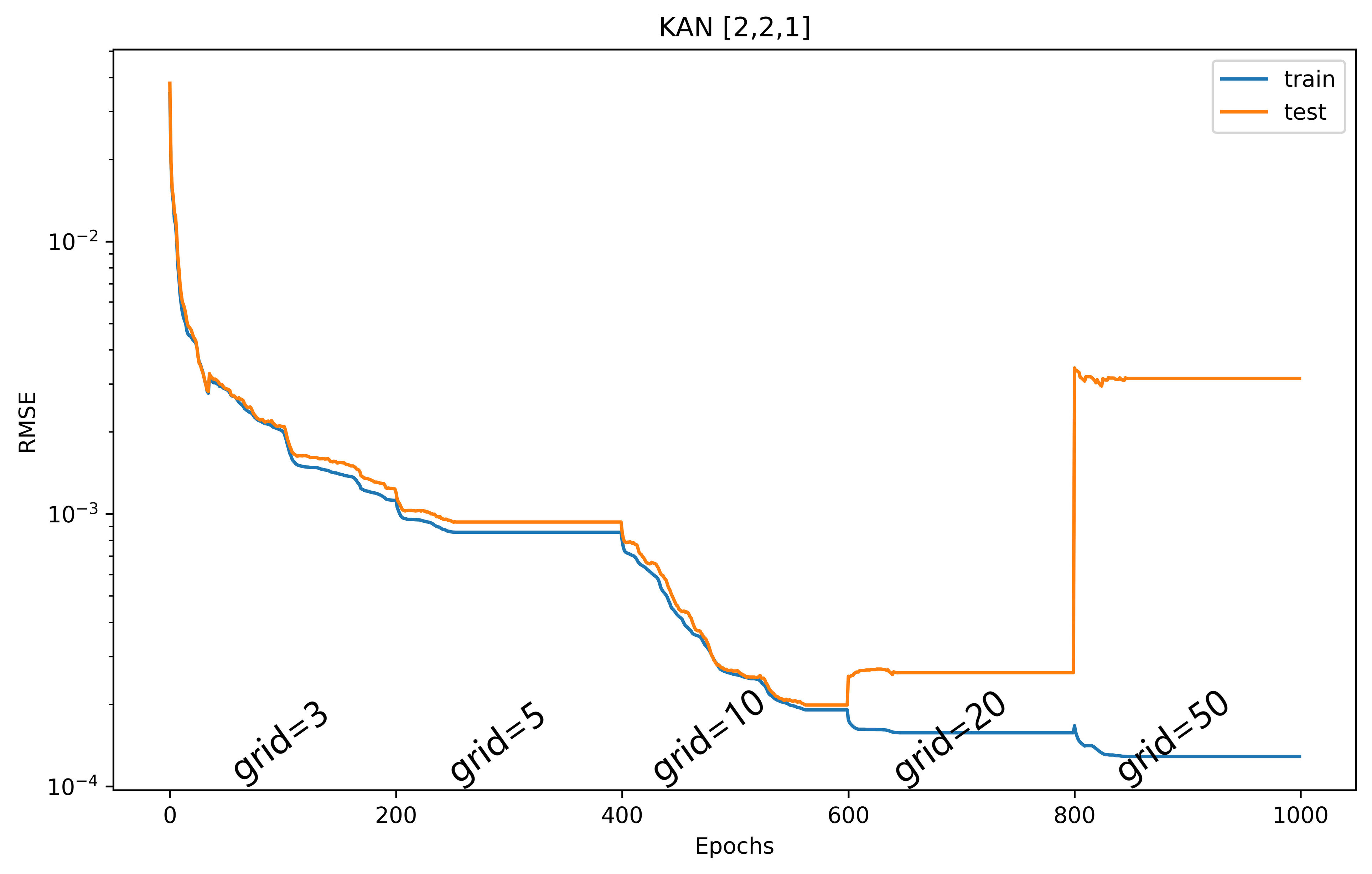}
        \captionof{figure}{Loss on \(xy\)}
    \end{minipage}
\end{figure}

for high-dimensional functions, loss functions
\begin{figure}[h!]
    \centering
    \begin{minipage}[b]{0.48\textwidth}
        \centering
        \includegraphics[width=0.48\textwidth]{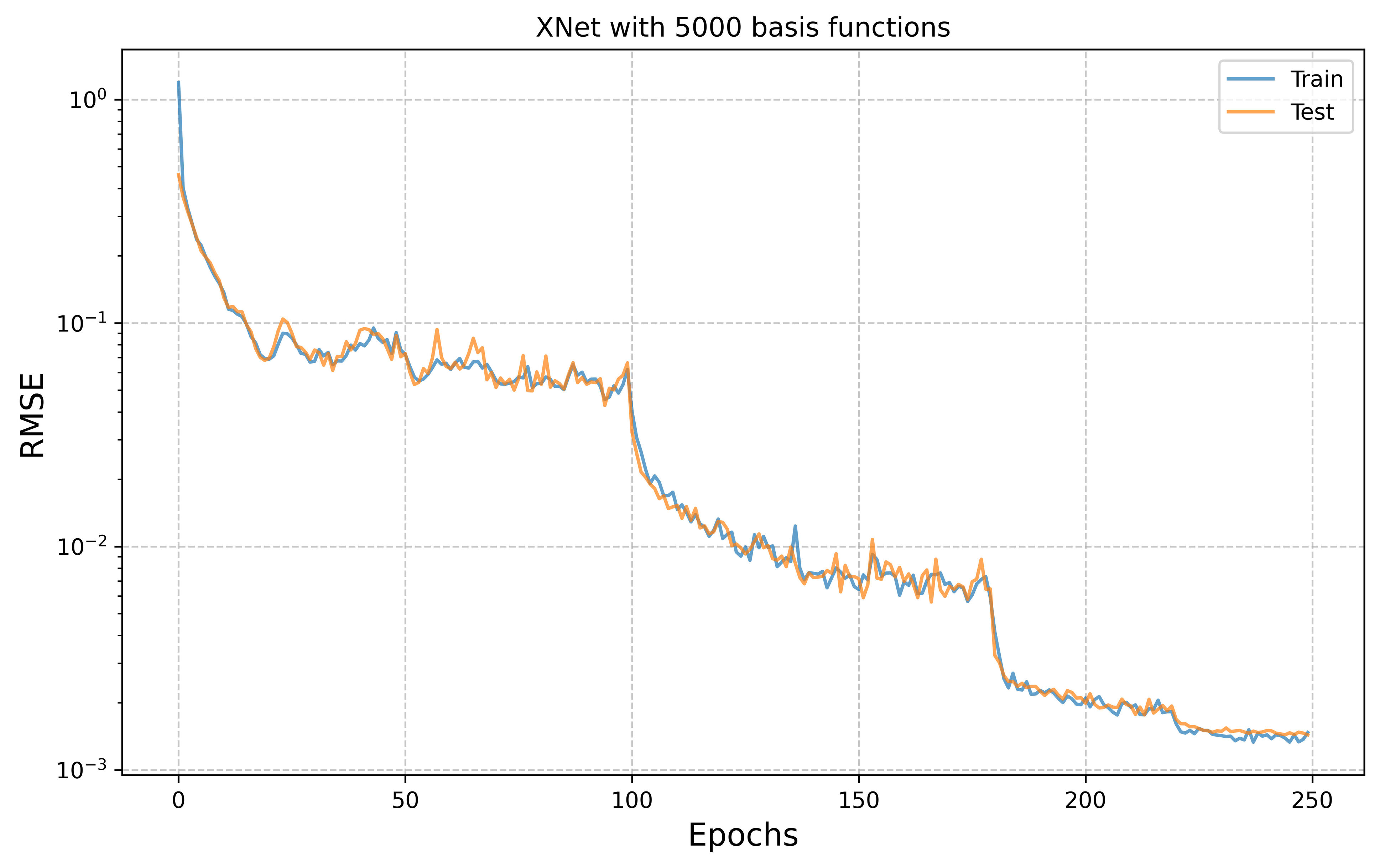}
        \hfill
        \includegraphics[width=0.48\textwidth]{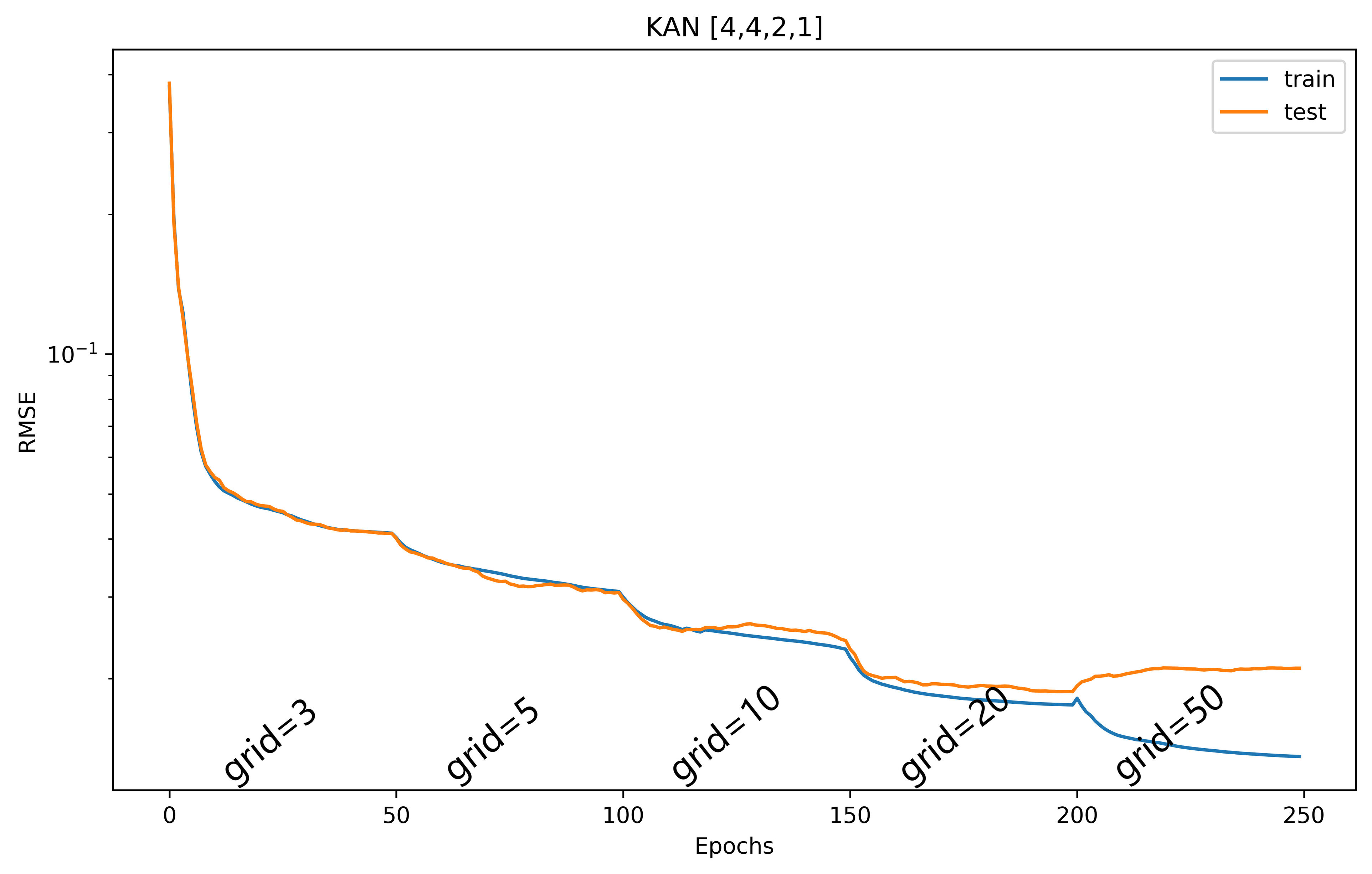}
        \caption*{ \( \exp\left(\frac{1}{2}\left(\sin\left(\pi(x_{1}^{2}+x_{2}^{2})\right) + x_{3}x_{4}\right)\right) \)}
    \end{minipage}
    \hfill
    \begin{minipage}[b]{0.48\textwidth}
        \centering
        \includegraphics[width=0.48\textwidth]{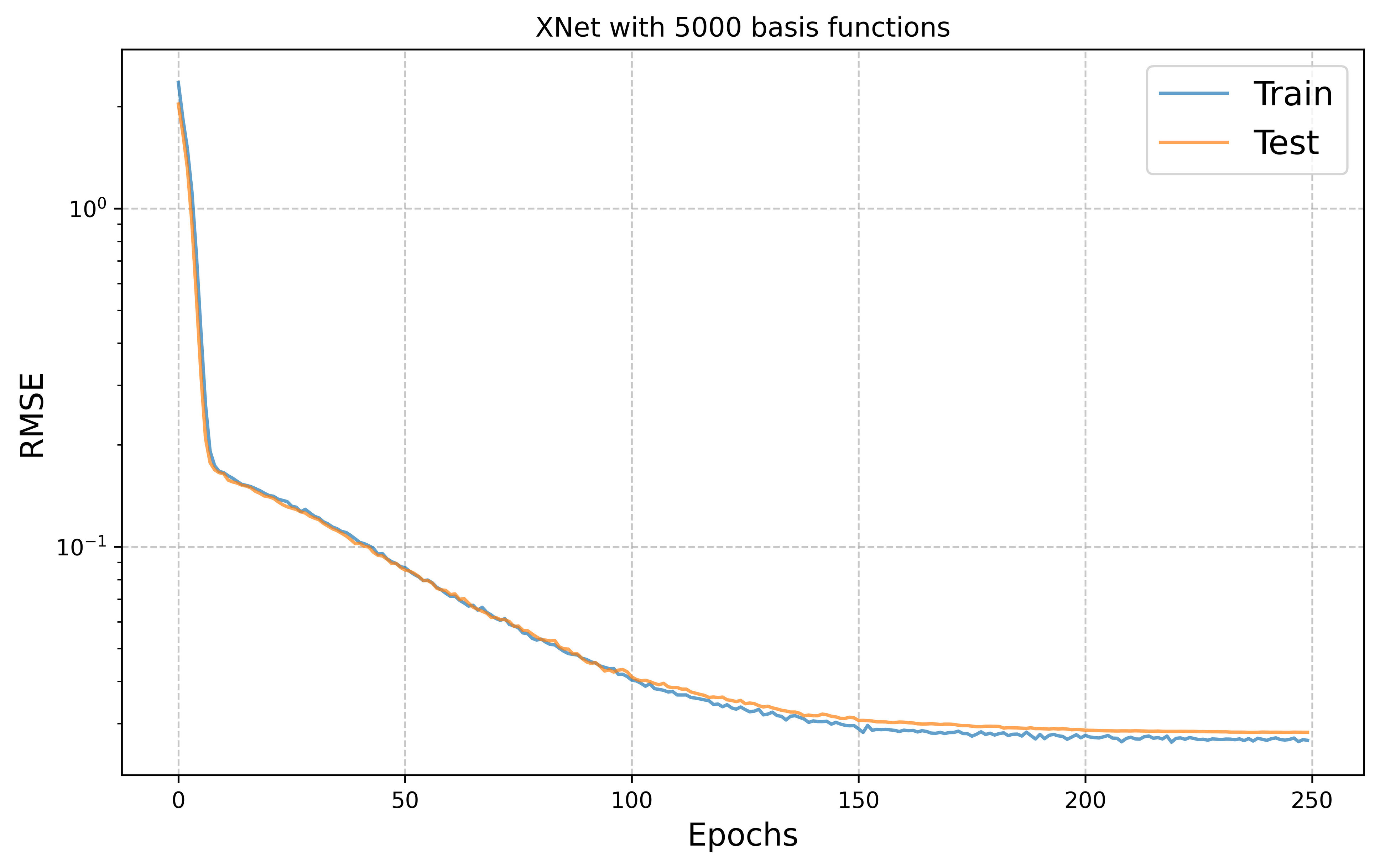}
        \hfill
        \includegraphics[width=0.48\textwidth]{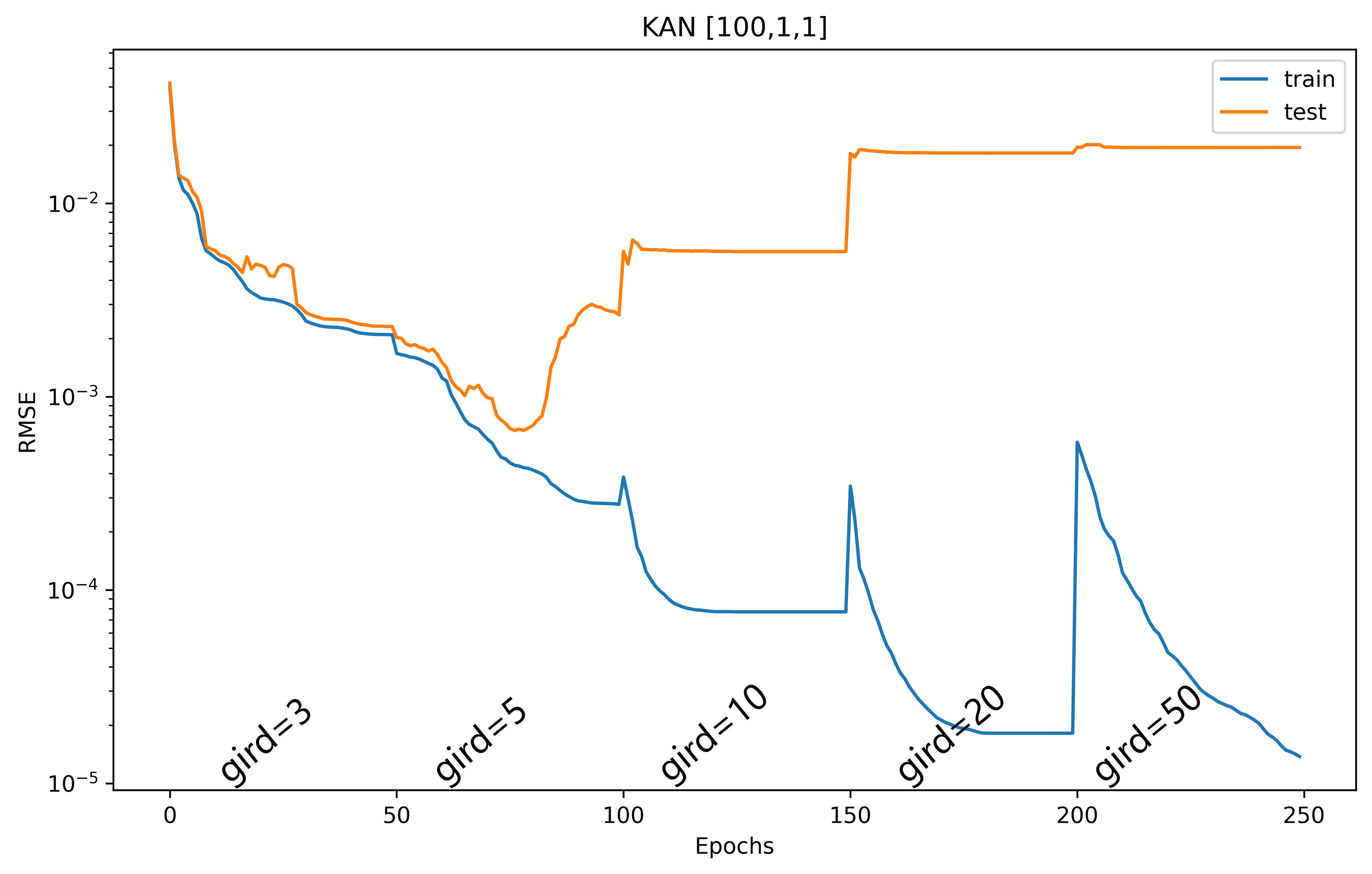}
        \caption*{\( \exp(\frac{1}{100}\sum_{i=1}^{100}\sin^2(\frac{\pi x_i}{2})) \)}
    \end{minipage}
	\caption{Loss on high-dimensional functions}
\end{figure}

\subsection{A.2 Time sereis} \label{A.2}
\begin{figure}[h!]
    \centering
    \begin{minipage}[b]{0.23\textwidth}
        \centering
        \includegraphics[width=\textwidth]{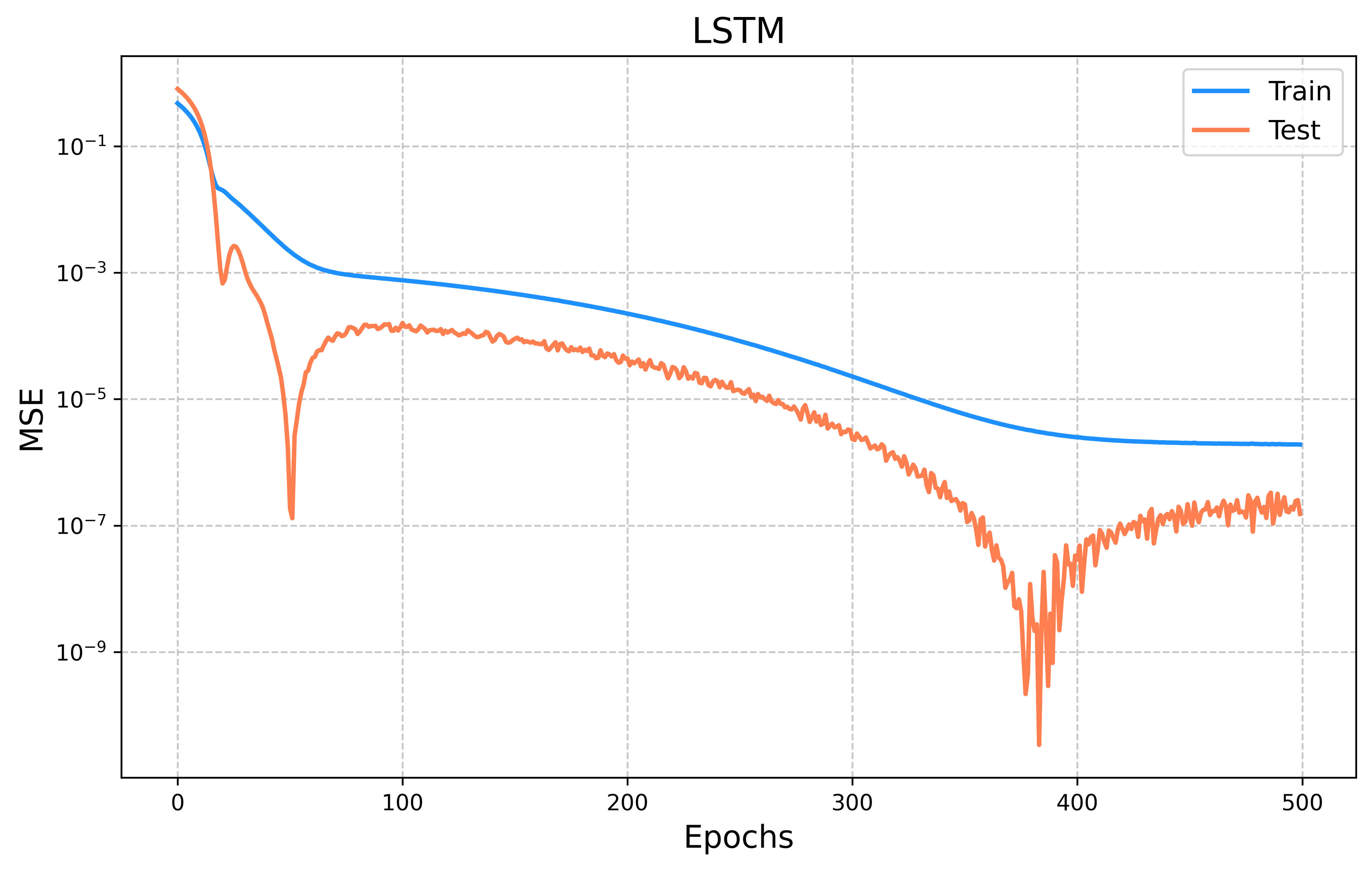}
    \end{minipage}
    \hfill
    \begin{minipage}[b]{0.23\textwidth}
        \centering
        \includegraphics[width=\textwidth]{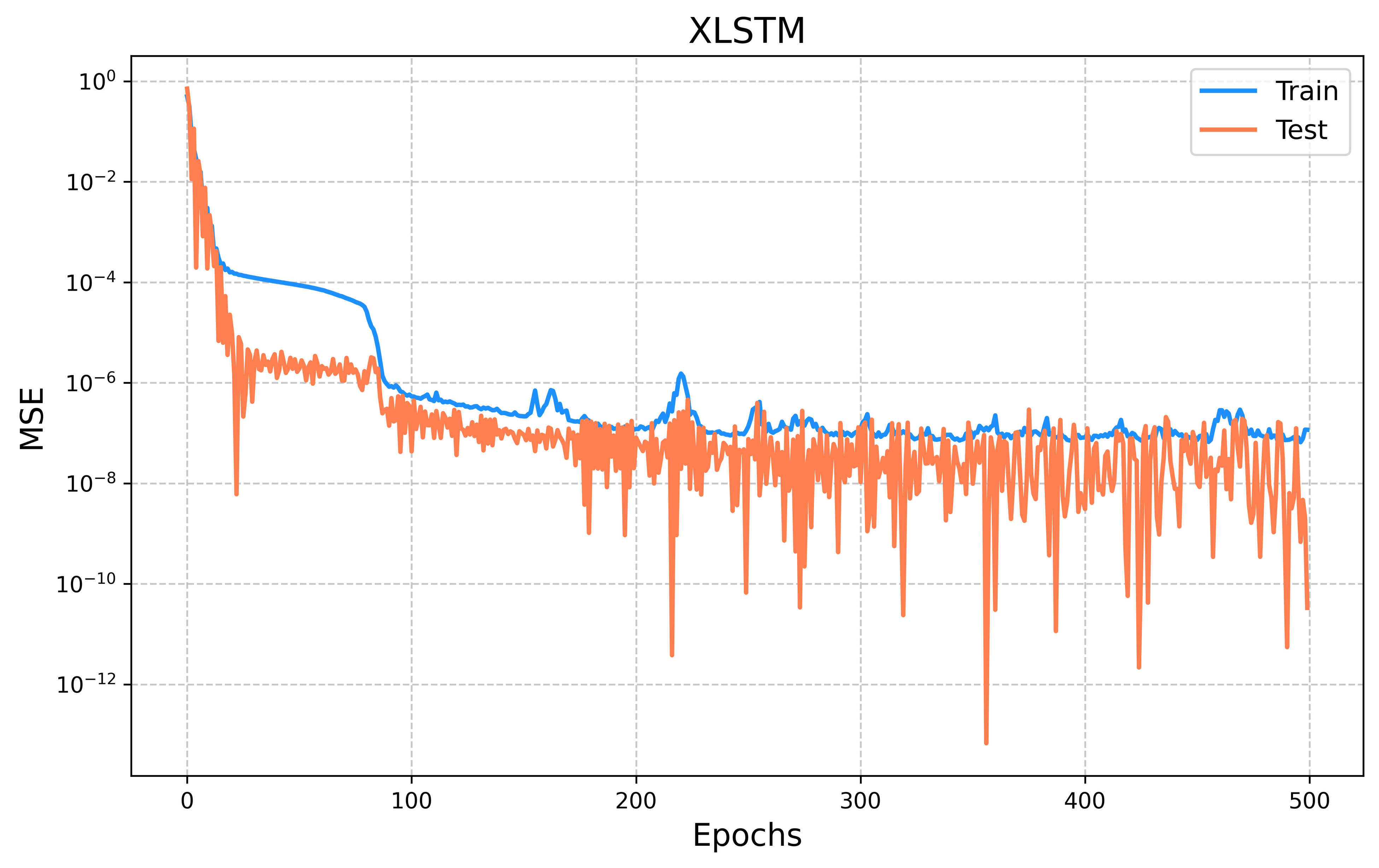}
    \end{minipage}
    \hfill
    \begin{minipage}[b]{0.23\textwidth}
        \centering
        \includegraphics[width=\textwidth]{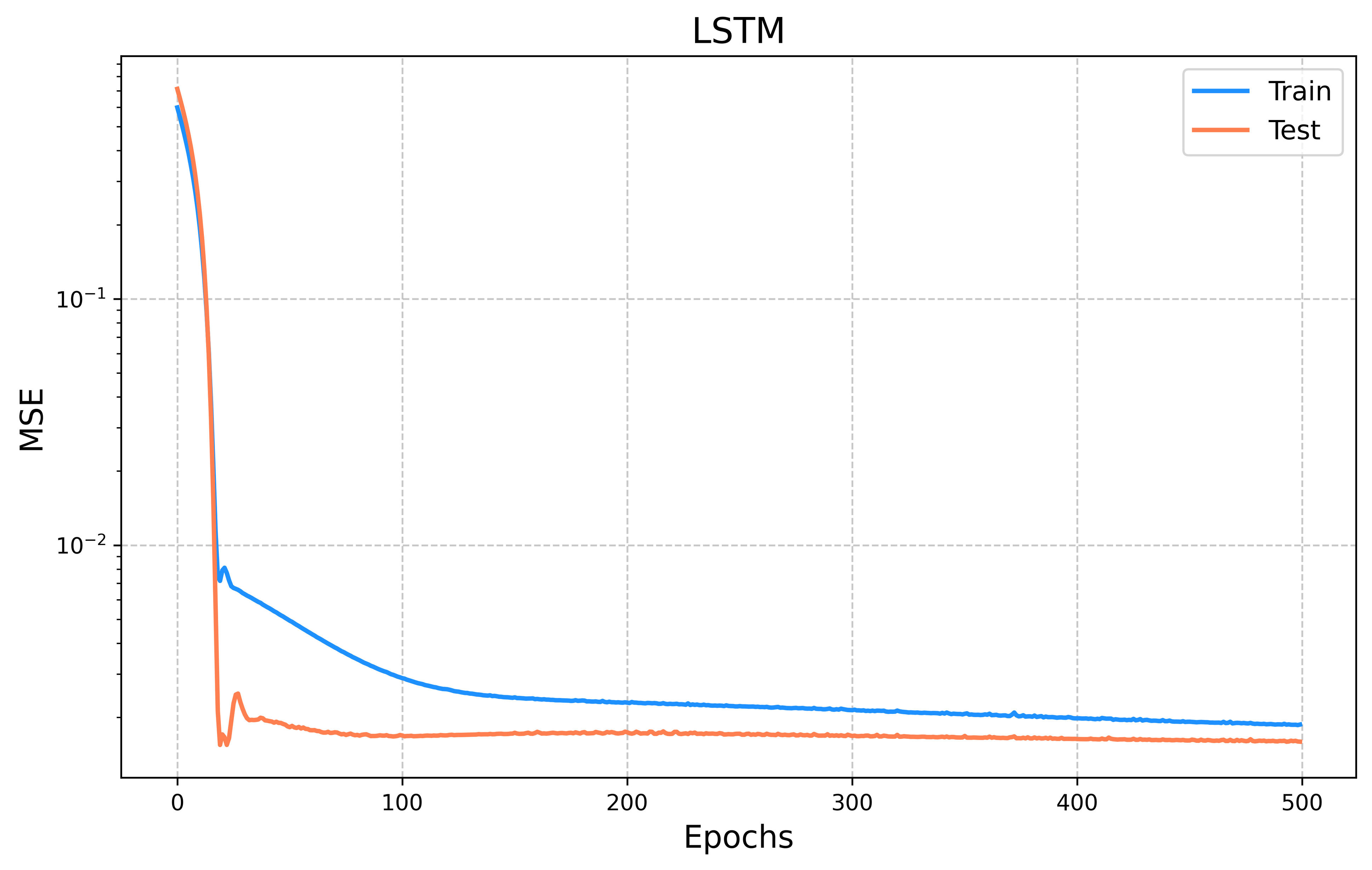}
    \end{minipage}
    \hfill
    \begin{minipage}[b]{0.23\textwidth}
        \centering
        \includegraphics[width=\textwidth]{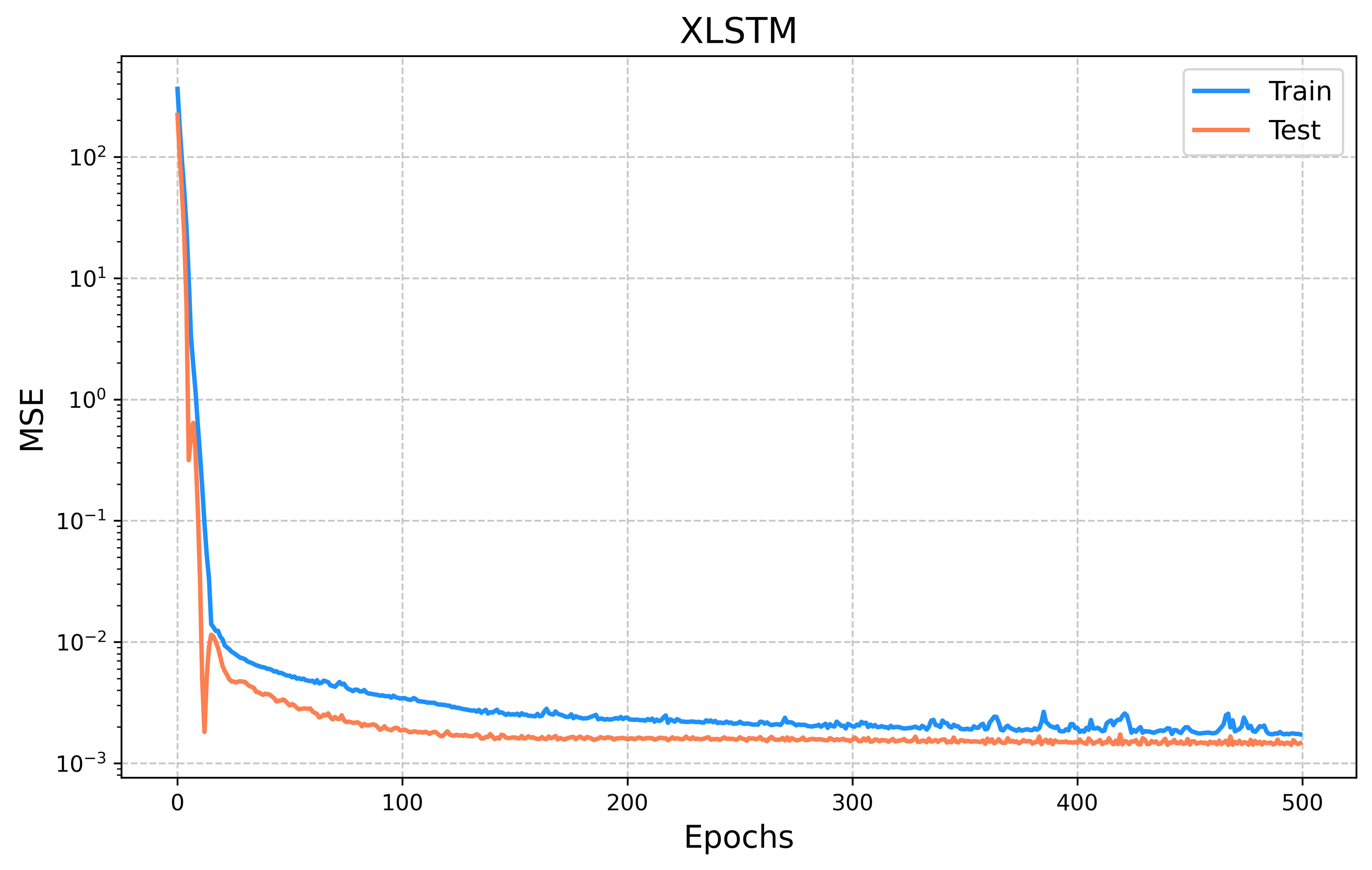}
    \end{minipage}
    \caption{Loss of LSTM and XLSTM on Example 1.}
    \label{fig:comparison_LSTM_XLSTM_loss}
\end{figure}

\begin{figure}[h!]
    \centering
    \begin{minipage}[b]{0.45\textwidth}
        \centering
        \includegraphics[width=\textwidth]{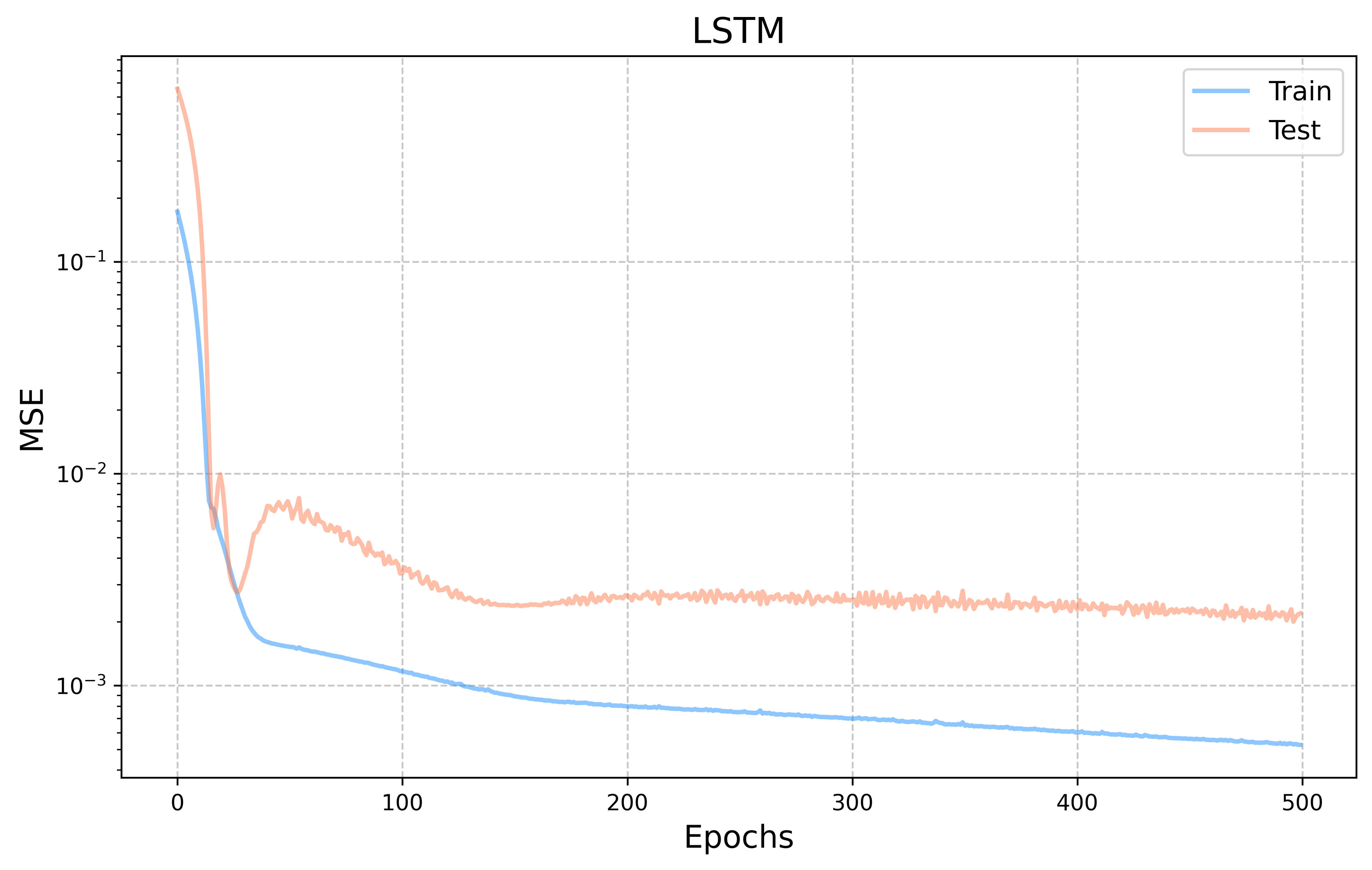}
    \end{minipage}
    \hfill
    \begin{minipage}[b]{0.45\textwidth}
        \centering
        \includegraphics[width=\textwidth]{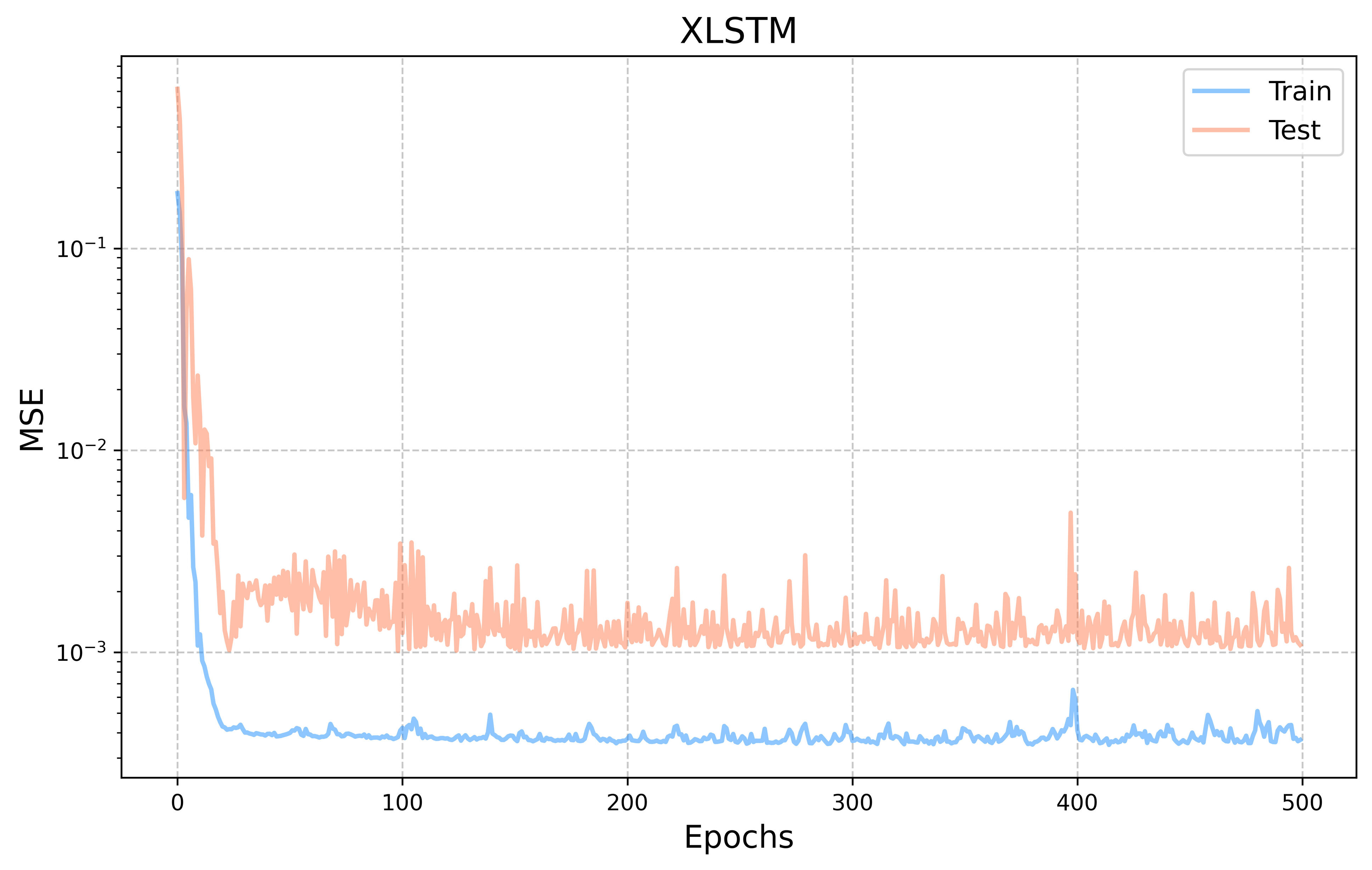}
    \end{minipage}
 \caption{loss of LSTM and XLSTM on Apple's stock price}
	\label{fig:stock_Model_loss}
\end{figure}

In Section \ref{sec:LSTM}, we present two examples to forecast future unknown data using LSTM and XLSTM. In the function-driven example (\ref{fig:ts1_noise}), the loss functions of LSTM and XLSTM are shown in Figure \ref{fig:comparison_LSTM_XLSTM_loss}; for the task of predicting Appleâs stock price, the loss functions of LSTM and XLSTM are illustrated in Figure \ref{fig:stock_Model_loss}.

Example 2 loss

\subsection{Time series} \label{sec:ts}
There exists two types of time prediction applications. One is driven by mathematical and physical models, where time prediction can essentially be viewed as a function approximation. The other is data-driven, where the data often contains significant noise and cannot be easily described by PDEs.
In this section, we introduce the XLSTM algorithm, which replaces the FNN component in the standard LSTM framework with XNet. In the following examples, XLSTM consistently demonstrates superior predictive performance.

\subsection{high-dimensional function 1}
The time series is generated by the following equations:
$$x_5^i=0.1*x_0^i*x_1^i+0.5*sin(x_2^i*x_3^i)+1*sin(x_4^i),i=1,2,...,n$$
and
$$x_0^i=x_1^{i-1},x_1^i=x_2^{i-1},x_2^i=x_3^{i-1},x_4^i=x_5^{i-1}, $$
with
$$x_0^0,x_1^0,x_2^0,x_3^0,x_4^0\thicksim rand(0,0.2). $$
This generates the time series $\{ f^i=x_5^i \}_{i=1,...,n}$. We consider the data n=200.
In this example, the time series is driven by simple functions. Specifically, when the task is to predict the sixth data point using the first five, it essentially becomes a high-dimensional function approximation problem.

We first split the data into a training set (80\%) and a validation set (20\%) and performed predictions using different models including 2-Layer width-10 FNN, 1-layer width-10 LSTM, width-10 XNet and width-10 XLSTM.

For each training iteration, the first five data points were used as input, and the model predicted the sixth data point, which was then compared with the target values. After five thousand iterations, the training process was considered complete. On the test set, we used the first five data points as input to predict the sixth, sliding through the sequence until all predictions were made. In essence, this can be viewed as a function-fitting problem.

\begin{figure}[h!]
    \centering
    \begin{minipage}[b]{0.48\textwidth}
        \centering
        \includegraphics[width=0.48\textwidth]{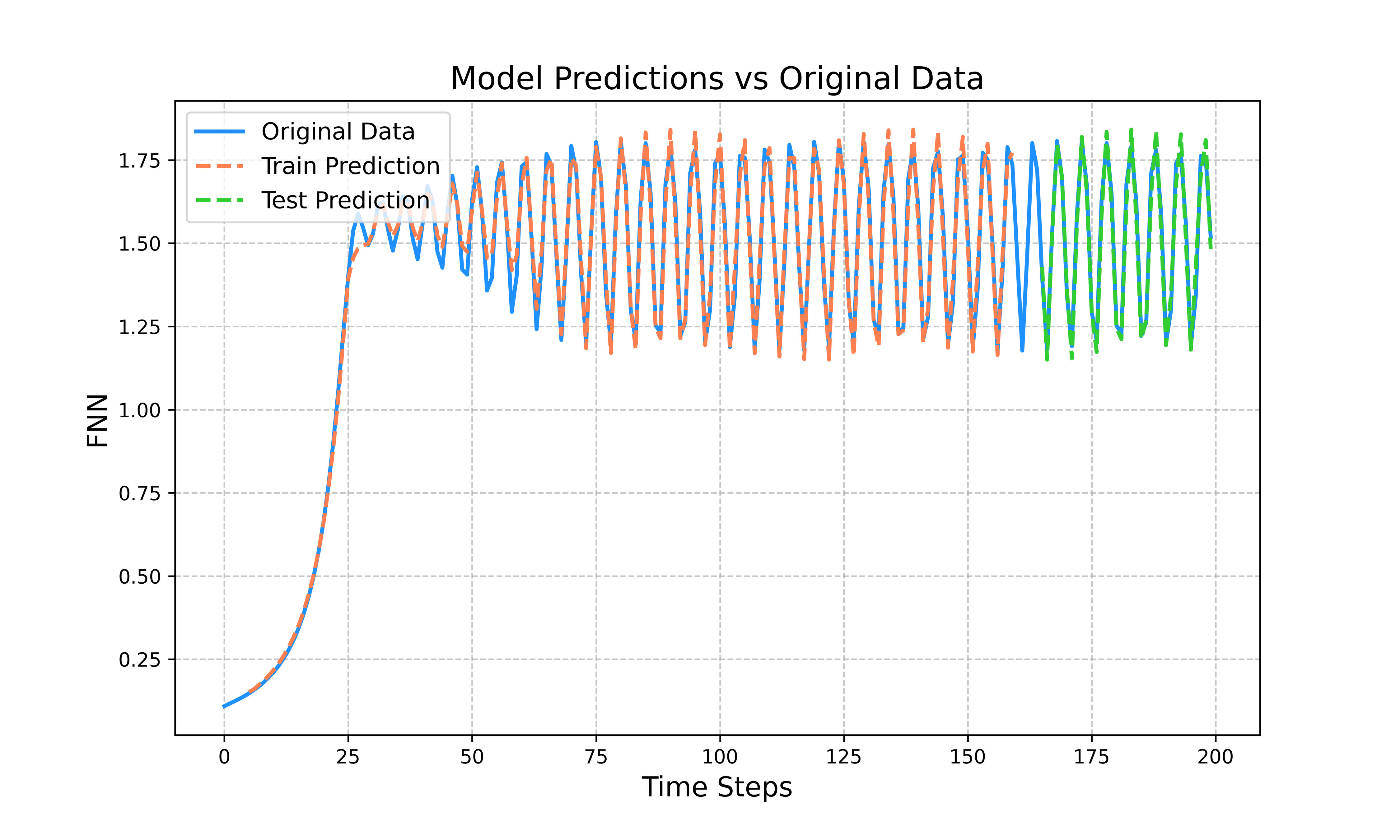}
        \hfill
        \includegraphics[width=0.48\textwidth]{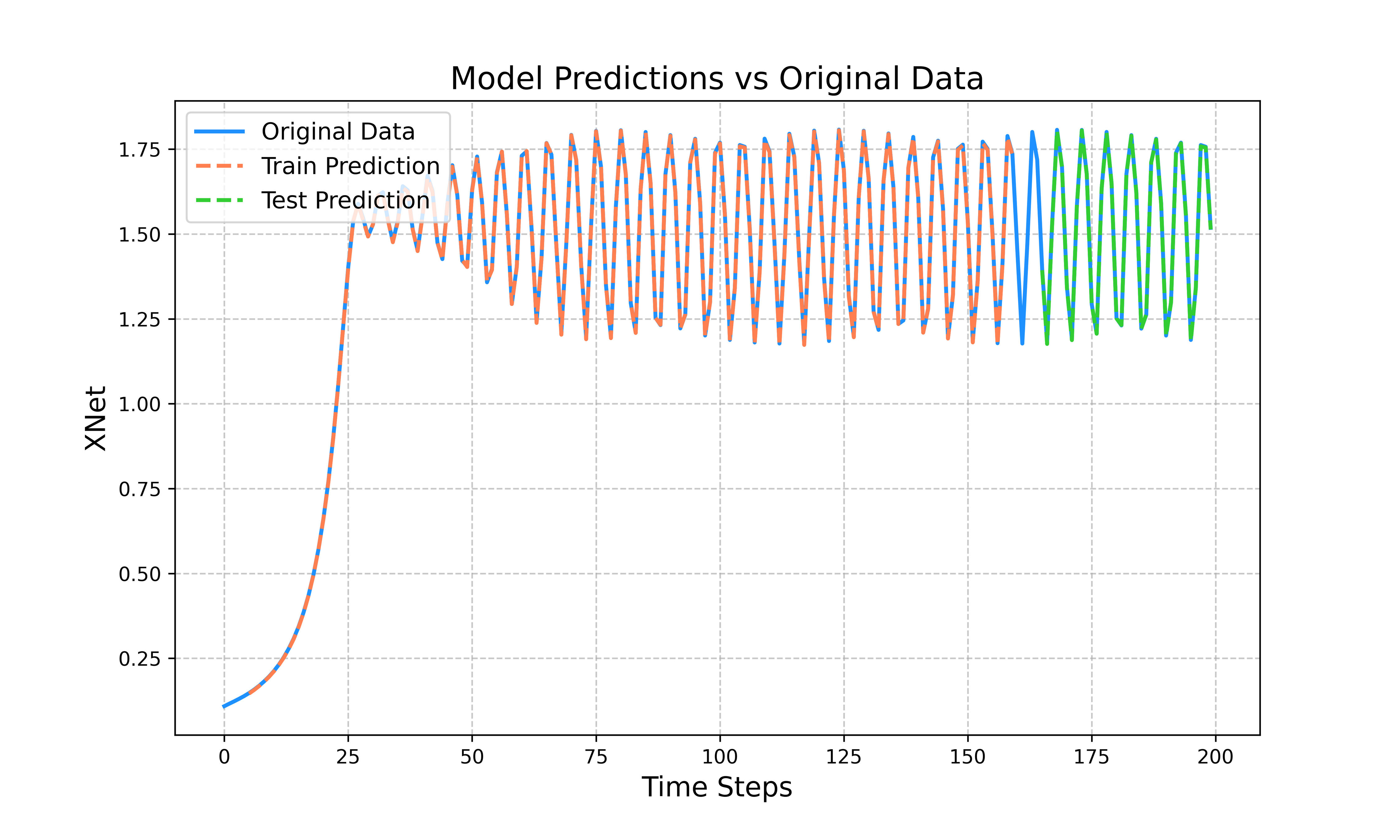}
    \end{minipage}
    \hfill
    \begin{minipage}[b]{0.48\textwidth}
        \centering
        \includegraphics[width=0.48\textwidth]{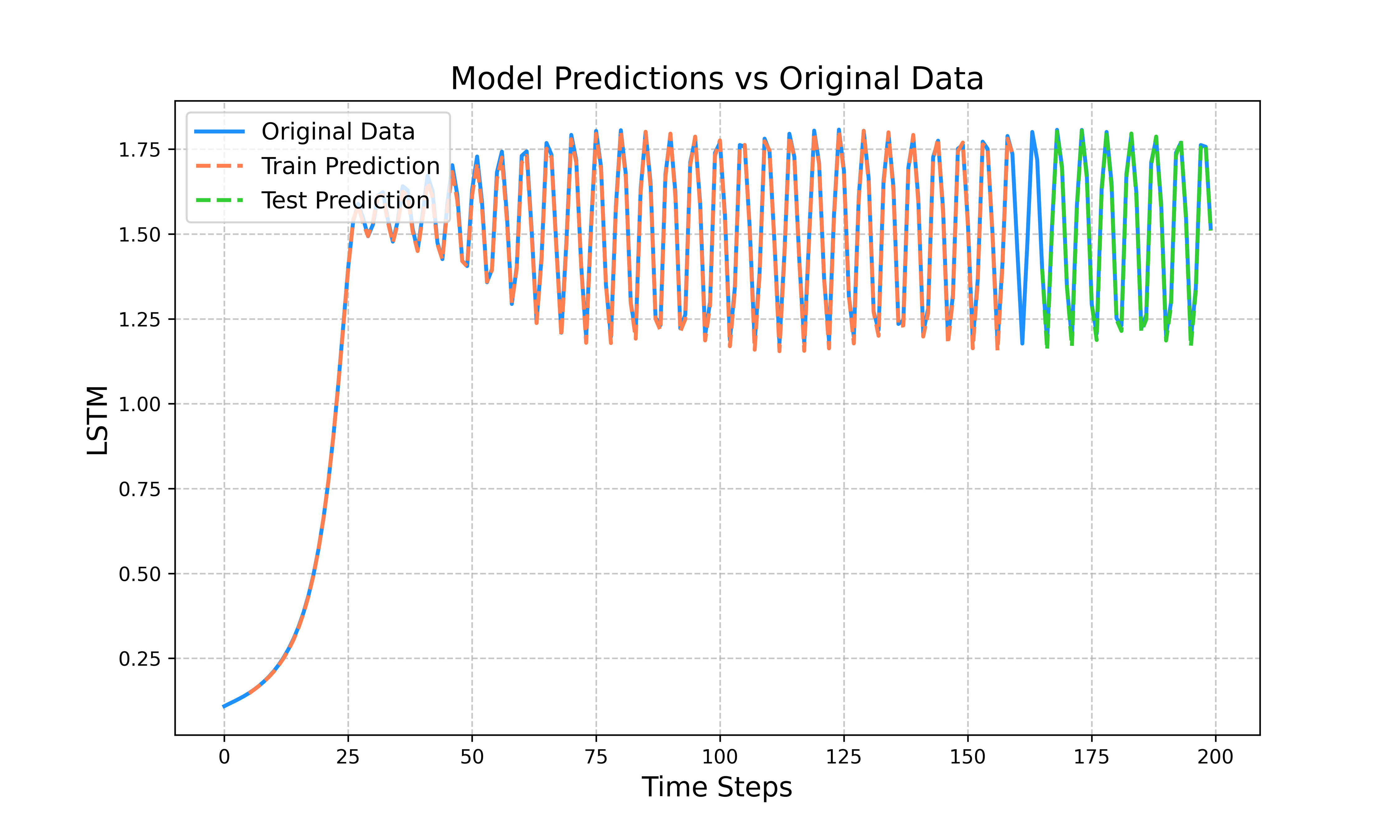}
        \hfill
        \includegraphics[width=0.48\textwidth]{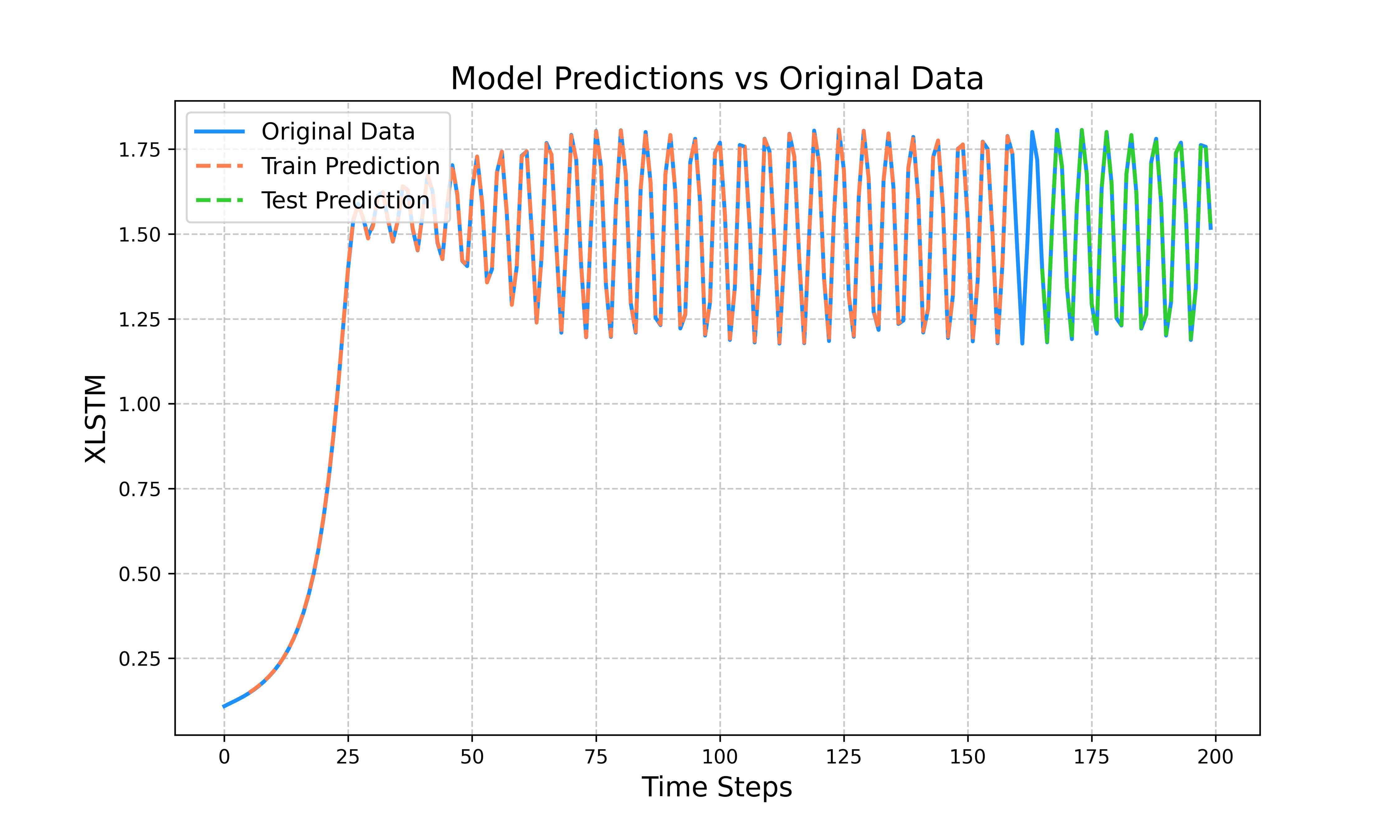}
    \end{minipage}
    \caption{different models}
\end{figure}

\begin{figure}[h!]
     \centering
     \includegraphics[width=0.5 \textwidth]{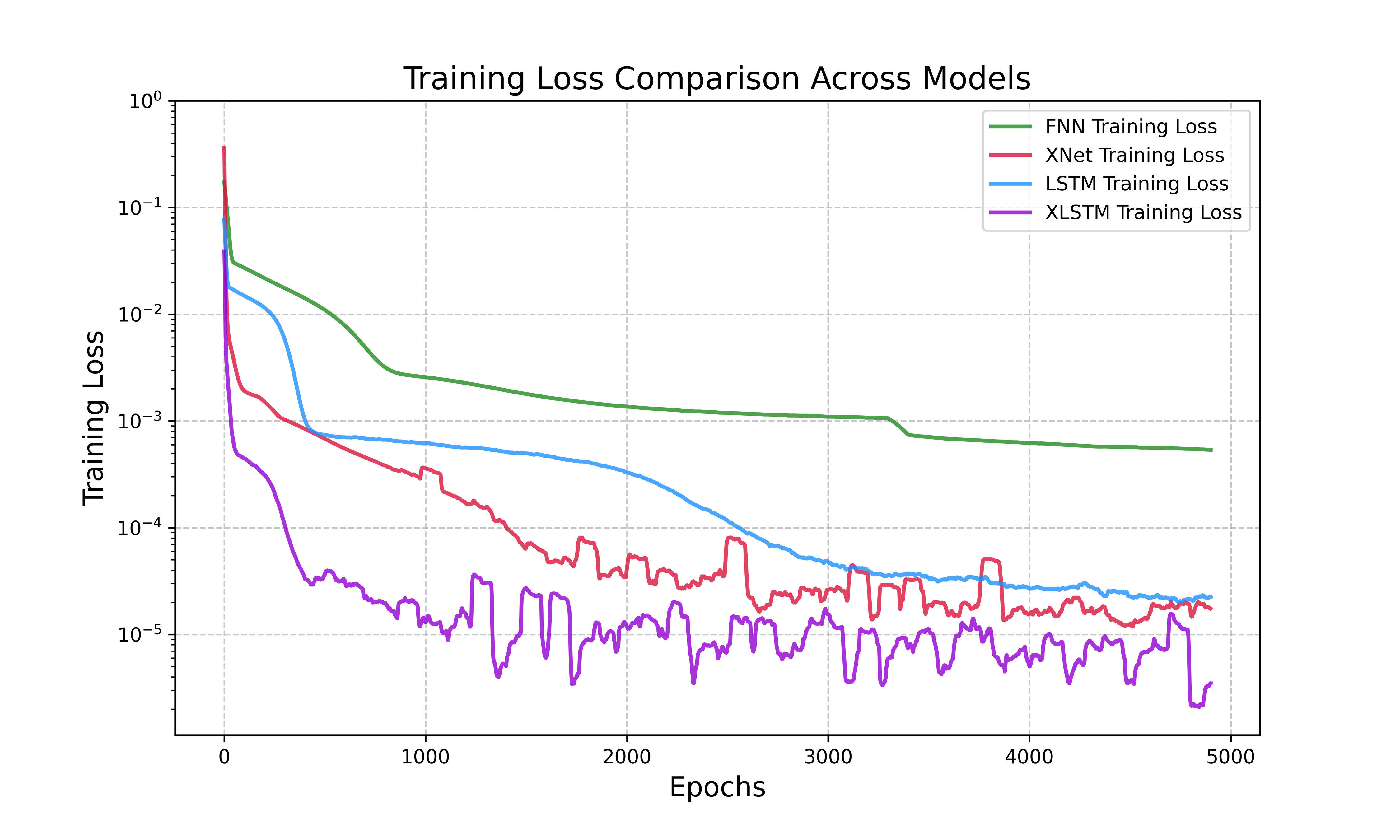}
      \caption{loss}
\end{figure}

XLSTM demonstrates stronger predictive capabilities compared to standard LSTM. With the same training cost, XLSTM improves accuracy by a factor of fifty.

\begin{table}[!ht]
    \centering
    \begin{tabular}{|l|l|l|l|l|}
    \hline
        ~ & FNN & XNet & LSTM & X-LSTM  \\ \hline
        MSE (Val) & 1.6253E-03 & 1.0758E-05 & 1.1187E-04 & 2.5222E-06  \\ \hline
        RMSE (Val) & 4.0315E-02 & 3.2800E-03 & 1.0577E-02 & 1.5881E-03  \\ \hline
        MAE (Val) & 3.3874E-02 & 2.7836E-03 & 9.0519E-03 & 1.1279E-03  \\ \hline
        MSE (Train) & 3.0175E-02 & 3.3013E-03 & 8.2499E-03 & 1.3336E-03  \\ \hline
        Time(s) & 6 & 6 & 12 & 12  \\ \hline
    \end{tabular}
\end{table}

Next, we apply XLSTM to stock price prediction and power consumption forecasting, where it again demonstrates stronger predictive capabilities compared to LSTM.

\subsection{electric power}
In this experiment, the time series represents electricity consumption in Zone 1 of the United States, with the test period from 01/01/2017 00:00 to 01/14/2017 21:20. The data is sourced from https://www.kaggle.com/datasets/fedesoriano/electric-power-consumption. The 2,000 data points are divided into two parts: 1,602 for training and 398 for testing. During training, the model takes the first 10 data points as input and predicts the 11th, comparing it with the target.

\begin{figure}[h!]
    \centering
    \begin{minipage}[b]{0.48\textwidth}
        \centering
        \includegraphics[width=0.48\textwidth]{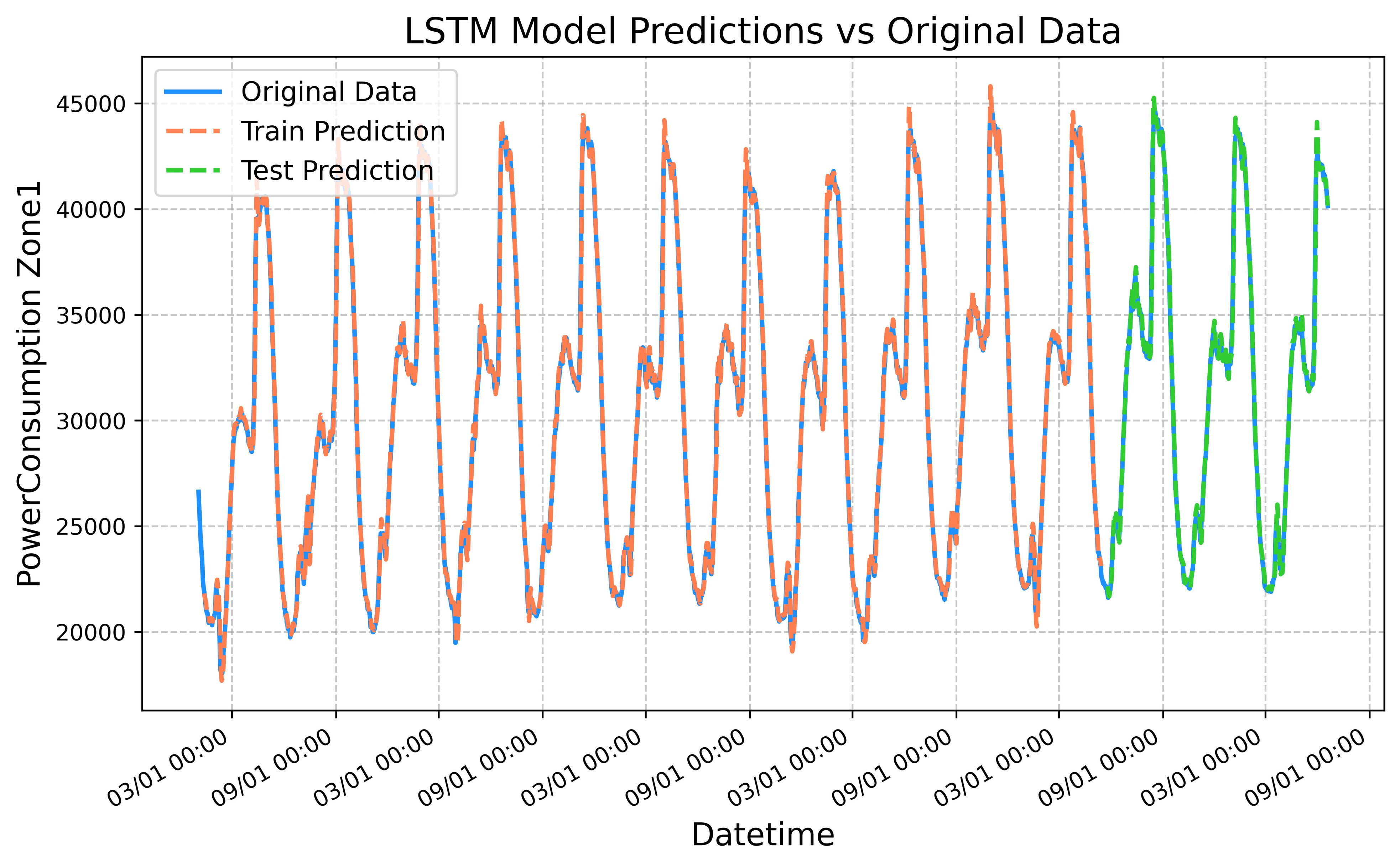}
	\hfill
        \includegraphics[width=0.48\textwidth]{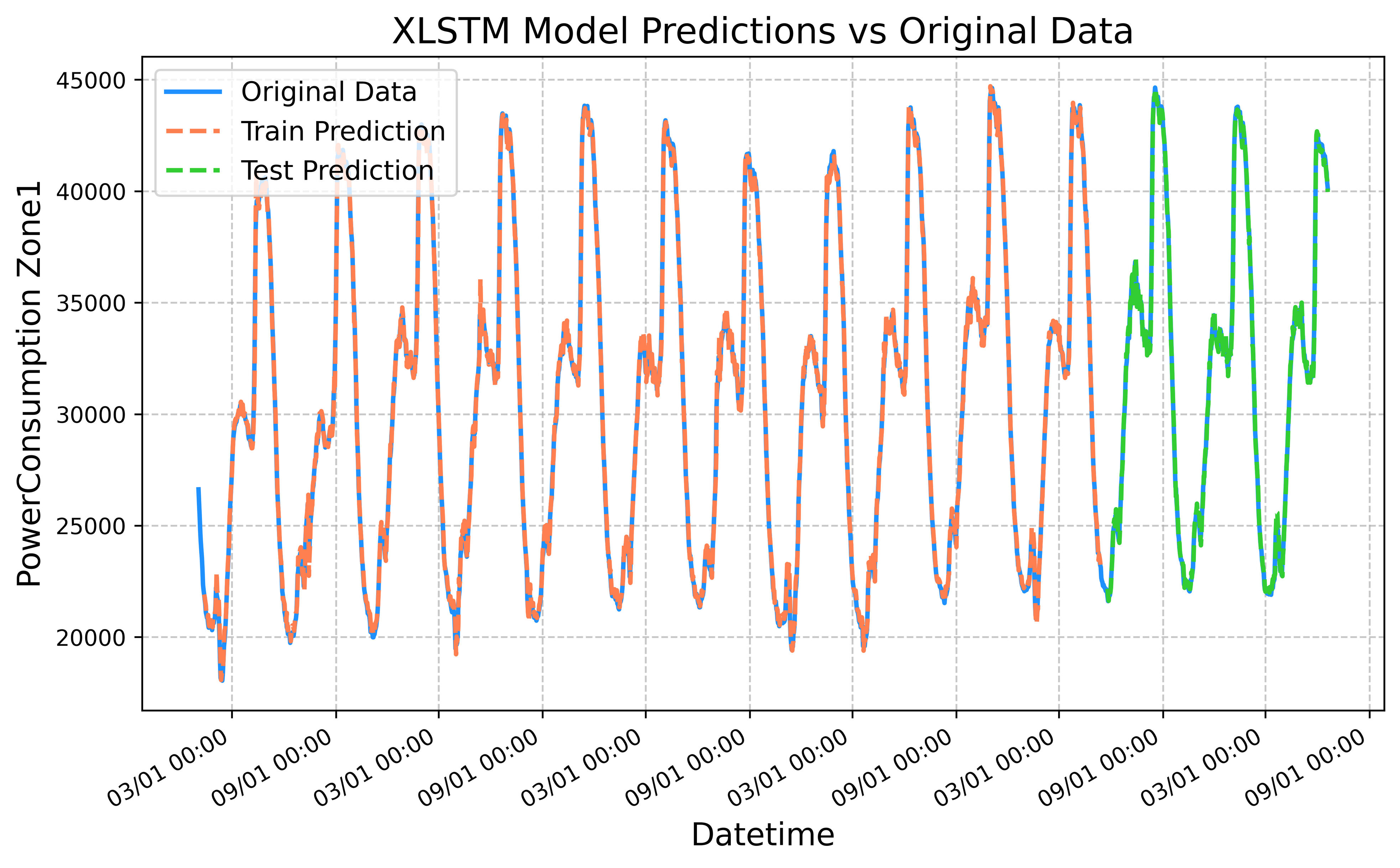}
    \end{minipage}
    \hfill
    \begin{minipage}[b]{0.48\textwidth}
        \centering
        \includegraphics[width=0.48\textwidth]{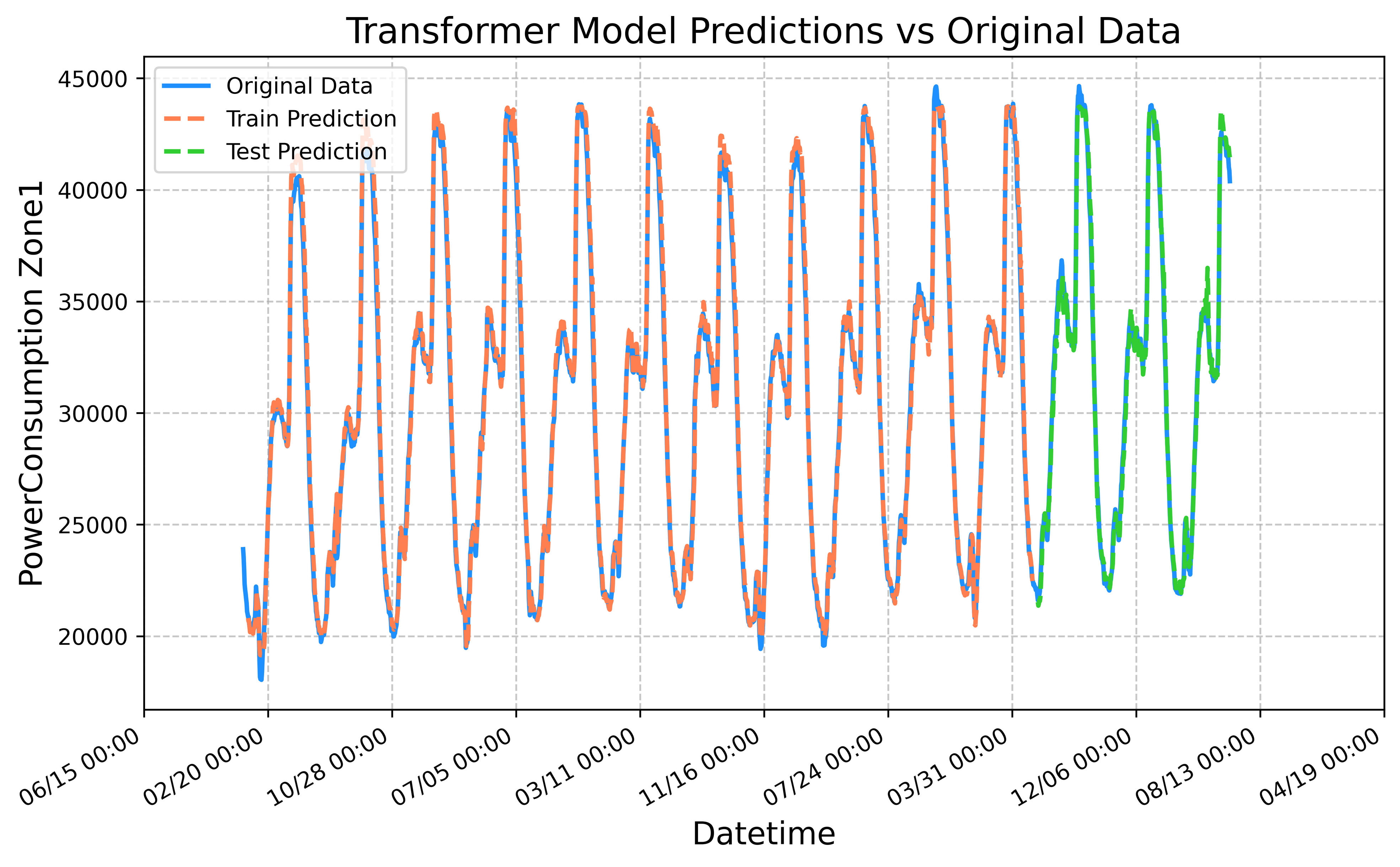}
	\hfill
        \includegraphics[width=0.48\textwidth]{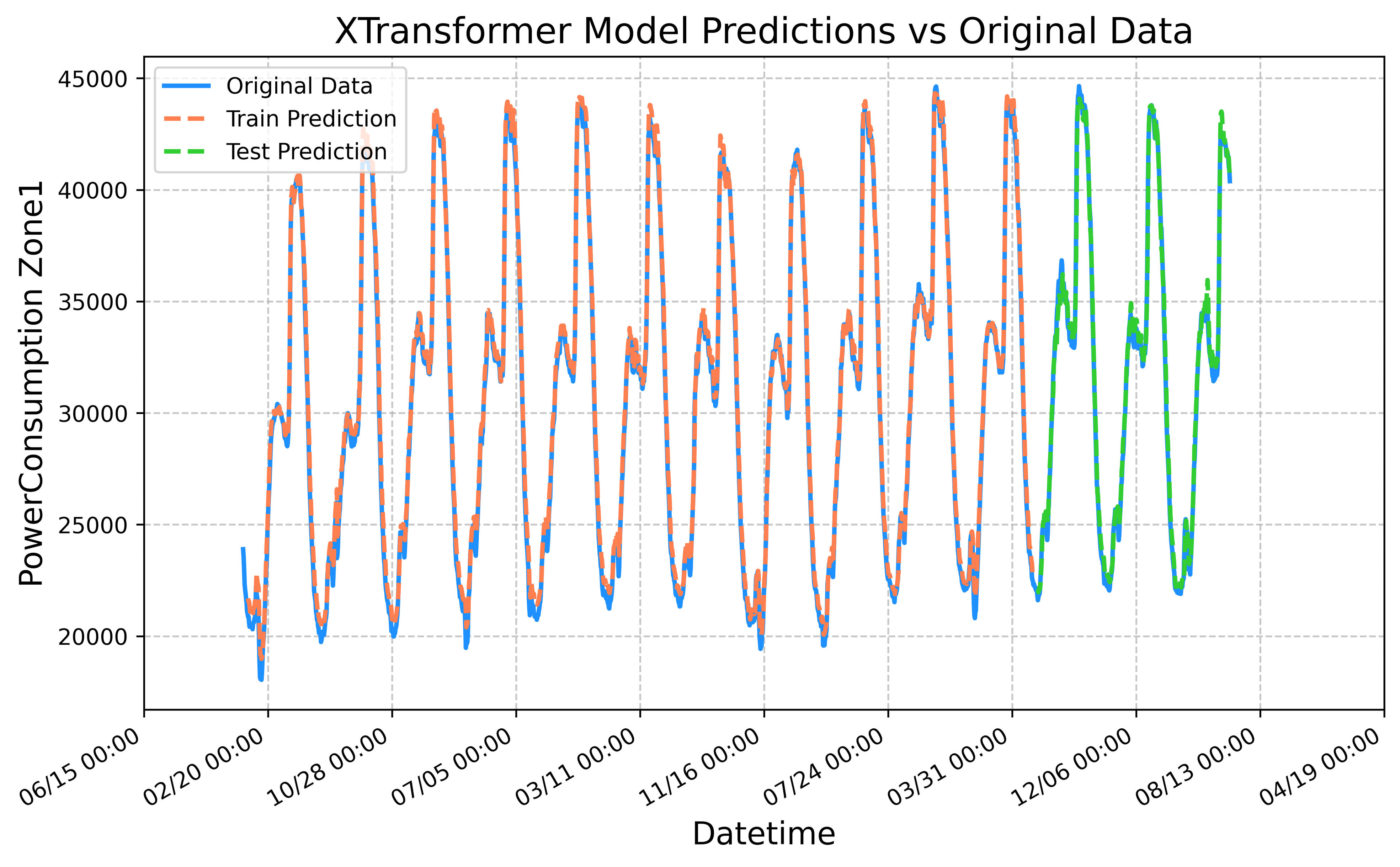}
    \end{minipage}
    \caption{electric power}
\end{figure}
XNet enhanced transformer and lstm model.
transformer has little advantage in this case

\begin{figure}[h!]
     \centering
     \includegraphics[width=0.5 \textwidth]{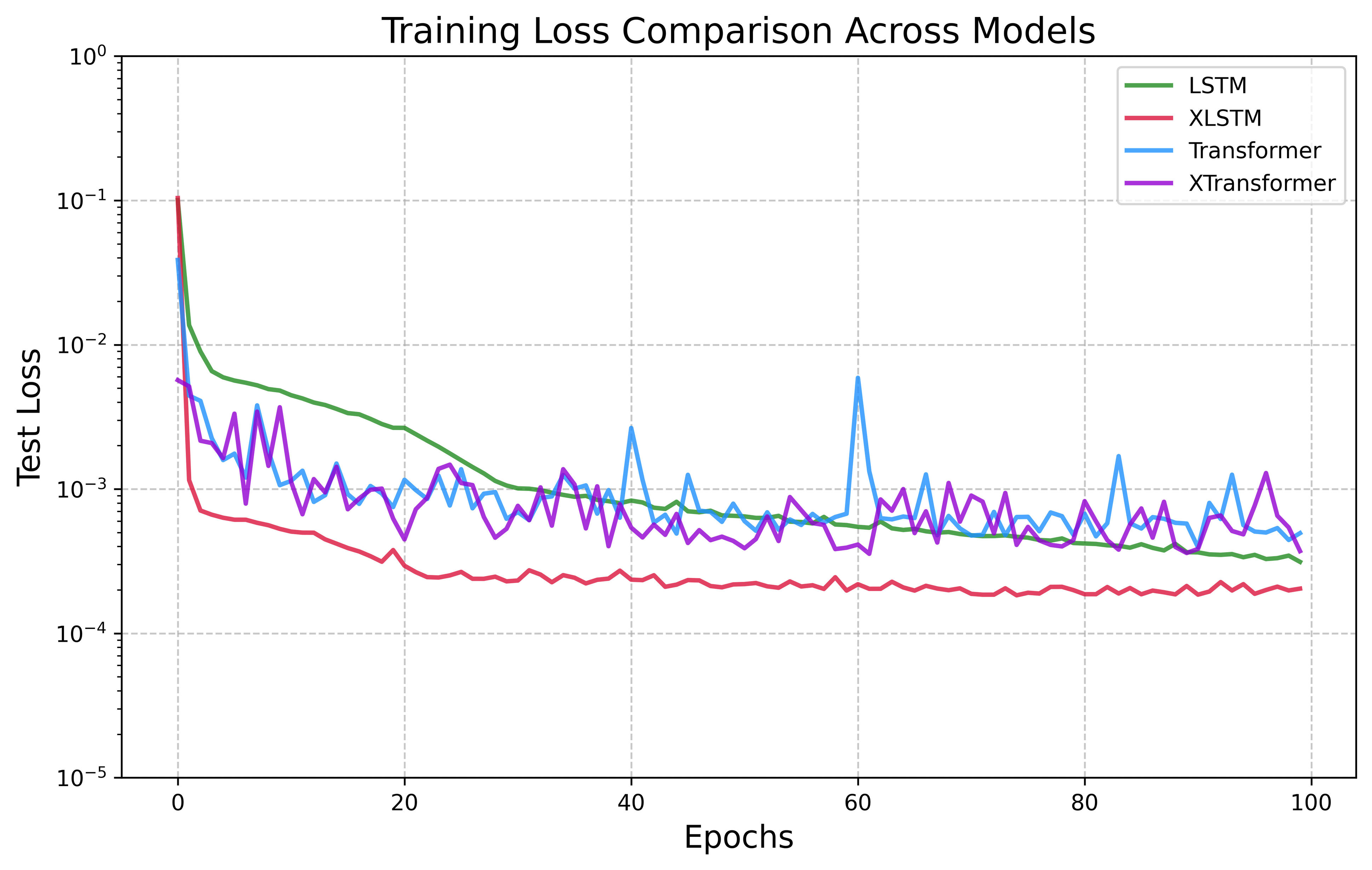}
      \caption{loss}
\end{figure}

\begin{table}[!ht]
    \centering
    \begin{tabular}{|l|l|l|l|l|}
    \hline
        ~ & LSTM & XLSTM & Transformer & XTransformer  \\ \hline
        MSE (Val) & 2.3937E+05 & 1.1505E+05 & 3.7482E+05 & 2.7868E+05  \\ \hline
        RMSE (Val) & 4.8925E+02 & 3.3920E+02 & 6.1223E+02 & 5.2790E+02  \\ \hline
        MAE (Val) & 3.2422E+02 & 2.6051E+02 & 4.9423E+02 & 4.1865E+02  \\ \hline
        MSE (Train) & 3.2729E+02 & 2.4623E+02 & 3.8049E+02 & 3.7939E+02  \\ \hline
        Time(s) & 15 & 26 & 127 & 90  \\ \hline
    \end{tabular}
\end{table}


\begin{thebibliography}{28}
\providecommand{\natexlab}[1]{#1}
\providecommand{\url}[1]{\texttt{#1}}
\expandafter\ifx\csname urlstyle\endcsname\relax
  \providecommand{\doi}[1]{doi: #1}\else
  \providecommand{\doi}{doi: \begingroup \urlstyle{rm}\Url}\fi

\bibitem[Braun \& Griebel(2009)Braun and Griebel]{Braun09}
J\"{u}rgen Braun and Michael Griebel.
\newblock On a constructive proof of kolmogorov's superposition theorem.
\newblock \emph{Constructive approximation}, 30:\penalty0 653--675, 2009.

\bibitem[Cunningham et~al.(2023)Cunningham, Ewart, Riggs, Huben, and
  Sharkey]{Cunningham23}
Hoagy Cunningham, Aidan Ewart, Logan Riggs, Robert Huben, and Lee Sharkey.
\newblock Sparse autoencoders find highly interpretable features in language
  models.
\newblock \emph{arXiv preprint arXiv:2309.08600}, 2023.

\bibitem[Cybenko(1989)]{Cybenko89}
George Cybenko.
\newblock Approximation by superpositions of a sigmoidal function.
\newblock \emph{Mathematics of control, signals and systems}, 2\penalty0
  (4):\penalty0 303--314, 1989.

\bibitem[Fakhoury et~al.(2022)Fakhoury, Fakhoury, and Speleers]{Fakhoury22}
Daniele Fakhoury, Emanuele Fakhoury, and Hendrik Speleers.
\newblock Exsplinet: An interpretable and expressive spline-based neural
  network.
\newblock \emph{Neural Networks}, 152:\penalty0 332--346, 2022.

\bibitem[Haykin(1994)]{Haykin94}
Simon Haykin.
\newblock \emph{Neural networks: a comprehensive foundation}.
\newblock Prentice Hall PTR, 1994.

\bibitem[Hochreiter \& Schmidhuber(1997)Hochreiter and
  Schmidhuber]{Hochreiter1997LongSM}
Sepp Hochreiter and J{\"u}rgen Schmidhuber.
\newblock Long short-term memory.
\newblock \emph{Neural Computation}, 9:\penalty0 1735--1780, 1997.
\newblock URL \url{https://api.semanticscholar.org/CorpusID:1915014}.

\bibitem[Hornik et~al.(1989)Hornik, Stinchcombe, and White]{Hornik89}
Kurt Hornik, Maxwell Stinchcombe, and Halbert White.
\newblock Multilayer feedforward networks are universal approximators.
\newblock \emph{Neural networks}, 2\penalty0 (5):\penalty0 359--366, 1989.

\bibitem[Jin et~al.(2021)Jin, Lu, Pang, Zhang, and Karniadakis]{Jin2021}
P.~Jin, L.~Lu, G.~Pang, Z.~Zhang, and G.~E. Karniadakis.
\newblock Learning nonlinear operators via deeponet based on the universal
  approximation theorem of operators.
\newblock \emph{Nature Machine Intelligence}, 3\penalty0 (3):\penalty0
  218--229, 2021.
\newblock \doi{10.1038/s42256-021-00302-5}.
\newblock URL \url{https://www.nature.com/articles/s42256-021-00302-5}.

\bibitem[Kolmogorov(1956)]{Kolmogorov56}
A.N. Kolmogorov.
\newblock On the representation of continuous functions of several variables as
  superpositions of continuous functions of a smaller number of variables.
\newblock \emph{Dokl. Akad. Nauk}, 108\penalty0 (2), 1956.

\bibitem[K\"oppen(2002)]{Koppen02}
Mario K\"oppen.
\newblock On the training of a kolmogorov network.
\newblock In \emph{Artificial Neural Networks—ICANN 2002: International
  Conference Madrid, Spain, August 28–30, 2002 Proceedings}, volume~12, pp.\
  474--479. Springer, 2002.

\bibitem[Lai \& Shen(2021)Lai and Shen]{Lai21}
Ming-Jun Lai and Zhaiming Shen.
\newblock The kolmogorov superposition theorem can break the curse of
  dimensionality when approximating high dimensional functions.
\newblock \emph{arXiv preprint arXiv:2112.09963}, 2021.

\bibitem[Leni et~al.(2013)Leni, Fougerolle, and Truchetet]{Leni13}
Pierre-Emmanuel Leni, Yohan~D Fougerolle, and Frédéric Truchetet.
\newblock The kolmogorov spline network for image processing.
\newblock In \emph{Image Processing: Concepts, Methodologies, Tools, and
  Applications}, pp.\  54--78. IGI Global, 2013.

\bibitem[Li et~al.(2024)Li, Xia, and Zhang]{LXZ24}
Xin Li, Zhihong Xia, and Hongkun Zhang.
\newblock Cauchy activation function and xnet.
\newblock \emph{arXiv preprint arXiv:2409.19221}, 2024.

\bibitem[Lin \& Unbehauen(1993)Lin and Unbehauen]{Lin93}
Ji-Nan Lin and Rolf Unbehauen.
\newblock On the realization of a kolmogorov network.
\newblock \emph{Neural Computation}, 5\penalty0 (1):\penalty0 18--20, 1993.

\bibitem[Liu et~al.(2024)Liu, Wang, Vaidya, Ruehle, Halverson, Soljacic, Hou,
  and Tegmark]{liu2024kan}
Ziming Liu, Yixuan Wang, Sachin Vaidya, Fabian Ruehle, James Halverson, Marin
  Soljacic, Thomas~Y. Hou, and Max Tegmark.
\newblock Kan: Kolmogorov-arnold networks, 2024.
\newblock URL \url{https://arxiv.org/abs/2404.19756}.

\bibitem[Nair \& Hinton(2010)Nair and Hinton]{Hinton2010}
Vinod Nair and Geoffrey~E Hinton.
\newblock Rectified linear units improve restricted boltzmann machines.
\newblock In \emph{Proceedings of the 27th international conference on machine
  learning (ICML-10)}, pp.\  807--814, 2010.

\bibitem[Raissi et~al.(2019)Raissi, Perdikaris, and Karniadakis]{Raissi2019}
M.~Raissi, P.~Perdikaris, and G.~E. Karniadakis.
\newblock Physics-informed neural networks: A deep learning framework for
  solving forward and inverse problems involving nonlinear partial differential
  equations.
\newblock \emph{Journal of Computational Physics}, 378:\penalty0 686--707,
  2019.
\newblock \doi{10.1016/j.jcp.2018.10.045}.
\newblock URL
  \url{https://www.sciencedirect.com/science/article/pii/S0021999118307125}.

\bibitem[Sirignano \& Spiliopoulos(2018)Sirignano and
  Spiliopoulos]{Sirignano2018}
J.~Sirignano and K.~Spiliopoulos.
\newblock Dgm: A deep learning algorithm for solving partial differential
  equations.
\newblock \emph{Journal of Computational Physics}, 375:\penalty0 1339--1364,
  2018.
\newblock \doi{10.1016/j.jcp.2018.08.029}.
\newblock URL
  \url{https://www.sciencedirect.com/science/article/pii/S0021999118305525}.

\bibitem[Sprecher \& Draghici(2002)Sprecher and Draghici]{Sprecher02}
David~A Sprecher and Sorin Draghici.
\newblock Space-filling curves and kolmogorov superposition-based neural
  networks.
\newblock \emph{Neural Networks}, 15\penalty0 (1):\penalty0 57--67, 2002.

\bibitem[Staudemeyer \& Morris(2019)Staudemeyer and
  Morris]{staudemeyer2019understanding}
Ralf~C Staudemeyer and Eric~Rothstein Morris.
\newblock Understanding lstm--a tutorial into long short-term memory recurrent
  neural networks.
\newblock \emph{arXiv preprint arXiv:1909.09586}, 2019.

\bibitem[Vaswani et~al.(2017)Vaswani, Shazeer, Parmar, Uszkoreit, Jones, Gomez,
  Kaiser, and Polosukhin]{Vaswani17}
Ashish Vaswani, Noam Shazeer, Niki Parmar, Jakob Uszkoreit, Llion Jones,
  Aidan~N Gomez, Lukasz Kaiser, and Illia Polosukhin.
\newblock Attention is all you need.
\newblock In \emph{Advances in neural information processing systems},
  volume~30, 2017.

\bibitem[Wen et~al.(2023)Wen, Zhou, Zhang, Chen, Ma, Yan, and
  Sun]{wen2023transformers}
Qingsong Wen, Tian Zhou, Chaoli Zhang, Weiqi Chen, Ziqing Ma, Junchi Yan, and
  Liang Sun.
\newblock Transformers in time series: a survey.
\newblock In \emph{Proceedings of the Thirty-Second International Joint
  Conference on Artificial Intelligence}, pp.\  6778--6786, 2023.

\bibitem[Wu et~al.(2024)Wu, Luo, Wang, Wang, and Long]{Transolver2024}
Haixu Wu, Huakun Luo, Haowen Wang, Jianmin Wang, and Mingsheng Long.
\newblock Transolver: A fast transformer solver for pdes on general geometries.
\newblock \emph{arXiv preprint arXiv:2402.02366}, 2024.
\newblock URL \url{https://arxiv.org/abs/2402.02366}.

\bibitem[Xu et~al.(2024)Xu, Chen, and Wang]{xu2024kolmogorov}
Kunpeng Xu, Lifei Chen, and Shengrui Wang.
\newblock Kolmogorov-arnold networks for time series: Bridging predictive power
  and interpretability.
\newblock \emph{arXiv preprint arXiv:2406.02496}, 2024.

\bibitem[Yu et~al.(2019)Yu, Si, Hu, and Zhang]{Yu2019}
Yong Yu, Xiaosheng Si, Changhua Hu, and Jianxun Zhang.
\newblock A review of recurrent neural networks: Lstm cells and network
  architectures.
\newblock \emph{Neural Computation}, 31, 2019.

\bibitem[Zhang et~al.(2024)Zhang, Li, and Xia]{ZLX24}
Hongkun Zhang, Xin Li, and Zhihong Xia.
\newblock Cauchynet: Utilizing complex activation functions for enhanced
  time-series forecasting and data imputation.
\newblock Submitted for publication, 2024.

\bibitem[Zhao et~al.(2017)Zhao, Chen, Wu, Chen, and Liu]{Zhao2017}
Zheng Zhao, Weihai Chen, Xingming Wu, Peter C.~Y. Chen, and Jingmeng Liu.
\newblock Lstm network: a deep learning approach for short-term traffic
  forecast.
\newblock \emph{IET Intelligent Transport Systems}, 11, 2017.

\bibitem[Zhao et~al.(2023)Zhao, Ding, and Aditya~Prakash]{PINNsFormer2023}
Zhiyuan Zhao, Xueying Ding, and B~Aditya~Prakash.
\newblock Pinnsformer: A transformer-based framework for physics-informed
  neural networks.
\newblock \emph{arXiv preprint arXiv:2307.11833}, 2023.
\newblock URL \url{https://arxiv.org/abs/2307.11833}.

\end{thebibliography}
\end{document}